\documentclass[10pt,twocolumn,letterpaper]{article}

\usepackage[pagenumbers]{cvpr} 

\definecolor{cvprblue}{rgb}{0.21,0.49,0.74}
\usepackage[pagebackref,breaklinks,colorlinks,allcolors=cvprblue]{hyperref}
\usepackage{afterpage}
\usepackage{algpseudocode}
\usepackage{algorithm}
\usepackage{amsthm}
\usepackage{bm}
\usepackage{tabularx}
\usepackage{array}
\usepackage{multirow}
\usepackage{subcaption}
\usepackage{listings}
\usepackage{xcolor}
\usepackage{multicol}
\usepackage{pdflscape}
\usepackage{colortbl}

\def\paperID{10114} 
\def\confName{CVPR}
\def\confYear{2025}

\title{
CLIP is Strong Enough to Fight Back:
Test-time Counterattacks towards Zero-shot Adversarial Robustness of CLIP
}

\author{Songlong Xing\textsuperscript{1} \quad Zhengyu Zhao\textsuperscript{2}\thanks{Corresponding author} \quad
Nicu Sebe\textsuperscript{1} \\
\textsuperscript{1}University of Trento, Italy \quad  \textsuperscript{2}Xi'an Jiaotong University, China \\
{\tt\small \{songlong.xing, niculae.sebe\}@unitn.it \quad  zhengyu.zhao@xjtu.edu.cn}
}

\begin{document}
\newcommand{\std}[1]{$\textcolor{gray}{\pm#1}$}
\newcommand{\blue}[1]{\textcolor{blue}{#1}}
\newcommand{\redd}[1]{\textcolor{red}{#1}}
\newcommand{\green}[1]{\textcolor{green}{#1}}

\maketitle
\begin{abstract}
Despite its prevalent use in image-text matching tasks in a zero-shot manner, CLIP has been shown to be highly vulnerable to adversarial perturbations added onto images.
Recent studies propose to finetune the vision encoder of CLIP with adversarial samples generated on the fly, 
and show improved robustness against adversarial attacks on a spectrum of downstream datasets, a property termed as zero-shot robustness.
In this paper, we show that malicious perturbations that seek to maximise the classification loss lead to `falsely stable' images, and propose to leverage the pre-trained vision encoder of CLIP to counterattack such adversarial images during inference to achieve robustness. 
Our paradigm is simple and training-free, providing the first method to defend CLIP from adversarial attacks at test time, which is orthogonal to existing methods aiming to boost zero-shot adversarial robustness of CLIP.
We conduct experiments across 16 classification datasets, and demonstrate stable and consistent gains compared to test-time defence methods adapted from existing adversarial robustness studies that do not rely on external networks, without noticeably impairing performance on clean images. 
We also show that our paradigm can be employed on CLIP models that have been adversarially finetuned to further enhance their robustness at test time.
Our code is available \href{https://github.com/Sxing2/CLIP-Test-time-Counterattacks}{here}.

\end{abstract}    
\section{Introduction}
\label{sec:intro}

\begin{figure}
    \centering
    \includegraphics[width=1.0\linewidth]{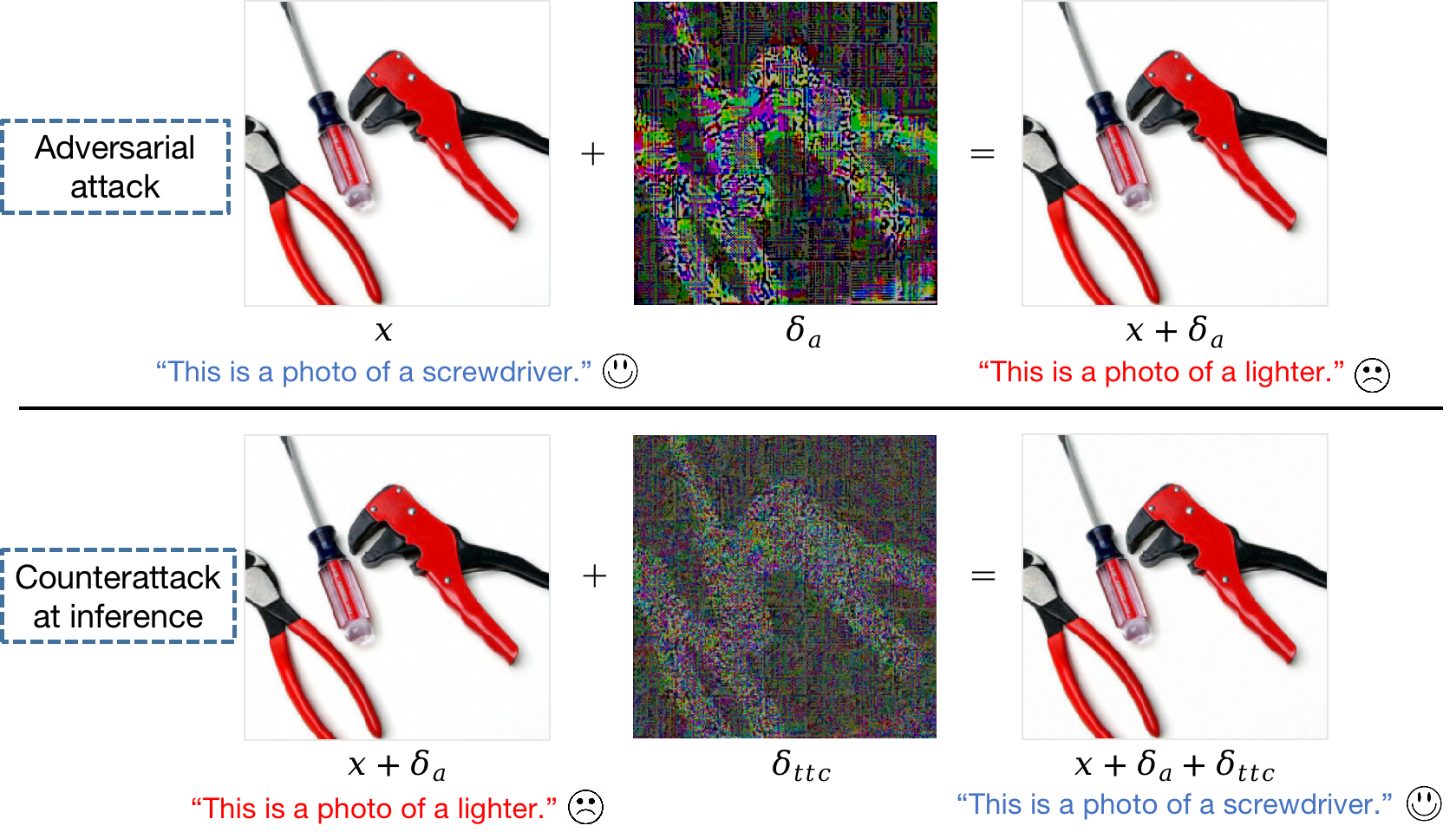}
    \caption{Test-time counterattacks harness the expressive power of CLIP to generate a counterattack to defend CLIP against adversaries without finetuning the vision encoder.
    }
    \label{fig:enter-label}
    \vspace{-0.5cm}
\end{figure}

With the increasing availability of image-text data and the advancement of self-supervised learning techniques \cite{chen2020simple,grill2020bootstrap,caron2021emerging}, vision-language models (VLM) have continued to spark research interests in both academia and industry \cite{radford2021learning,jia2021scaling,yu2022coca,ramesh2021zero,saharia2022photorealistic,NEURIPS2023_6dcf277e}. 
As a representative VLM, CLIP \cite{radford2021learning} has shown impressive abilities to match an image with its descriptive text in a zero-shot manner. 
However, recent studies have shown that adding small imperceptible perturbations to an image can cause CLIP to misclassify it \cite{mao2023understanding,zhou2024few,li2024language,li2024one,wang2024pre,pmlr-v235-schlarmann24a,zhang2023adversarial}, a common problem plaguing nearly all neural networks \cite{szegedy2013intriguing,madry2018towards,athalye2018obfuscated,carlini2017towards,kurakin2018adversarial,papernot2016limitations,moosavi2016deepfool,zhang2019theoretically,yucel2020deep}.
As foundational models are deployed in real-world applications, their safety and reliability have become a pressing concern. In this paper, we focus on the  robustness of CLIP against adversarial perturbations.

Unlike conventional models for which adversarial robustness has been extensively studied
\cite{croce2020reliable,athalye2018obfuscated,carlini2017towards,su2018robustness,moosavi2016deepfool,szegedy2013intriguing}, 
CLIP is a pre-trained foundation model that has learned massive amounts of real-world knowledge, and should be dealt with carefully to minimise damage to its generalisation abilities. 
Adversarial robustness of CLIP has just started to garner research attention \cite{mao2023understanding,li2024one,wang2024pre,pmlr-v235-schlarmann24a} in recent years. Existing efforts fall into two categories. 
The first type is based on adversarial training \cite{zhang2019theoretically,ijcai2021p0591}, which alternately generates adversarial images on one dataset and uses them to finetune the vision encoder of CLIP \cite{mao2023understanding,wang2024pre}. This type of methods, known as \textit{adversarial finetuning} (AFT), dynamically mimics a min-max game between CLIP and the threat model in the finetuning phase, and deploys the finetuned model in a wide variety of downstream classification tasks without further training. This method shows transferable robustness to downstream datasets, a property termed as zero-shot robustness \cite{mao2023understanding,wang2024pre}.
The other type of methods resorts to prompt tuning \cite{zhou2022learning,zhou2022conditional}, which inserts learnable text tokens in the embedding space, aligns the corresponding text prompts with adversarial images,
and tunes the learnable tokens by propagating gradients to the text embeddings.
This approach is known as \textit{adversarial prompt tuning} (APT) \cite{li2024one,zhang2023adversarial}.
Although these methods have shown improved robustness over the original CLIP, there are apparent limitations. 
\textit{Firstly}, they require time-consuming training, especially adversarial finetuning which involves generating adversarial images on the fly.
\textit{Secondly}, the model overfits to the training data, which compromises generalisation on clean and adversarial images from other data distributions. 
In the case of adversarial finetuning, the adversarially finetuned CLIP outperforms the original CLIP on clean images from the dataset used for finetuning, indicating that the model has overfitted to the distribution of the dataset (See \cref{tab:white_box_eps1} in \cref{sec:exp}).
\textit{Thirdly}, for adversarial finetuning methods \cite{mao2023understanding,wang2024pre}, robustness to adversarial samples comes at the cost of a significant decline of classification accuracy on clean images.

To address these limitations, we draw inspiration from existing adversarial robustness studies that robustify non-foundational target models at test time \cite{alfarra2022combating,wu2021attacking,mao2021adversarial,guo2018countering}, and propose a test-time paradigm that utilises the expressive power of CLIP to defend itself from adversarial attacks (\cref{fig:enter-label}).
Following previous studies, we focus on adversaries that aim to maximise the classification loss of CLIP given a test image.
We observe that such adversaries cause images to be more stable when a small random noise is added, compared to clean images.
We term this behaviour of adversarial images `false stability', which can be interpreted as the images being trapped in a toxic surrounding in the latent space by an adversary.
Intuitively, the pre-trained vision encoder of CLIP is highly expressive, and can be leveraged to push the adversarial image away from its toxic embedding.
Therefore, we propose to employ the vision encoder of CLIP to counteract the false stability of adversarial images, 
thereby achieving robustness to these attacks.
Since no label information is available at test time, we formulate the test image as an anchor, and iteratively update the counterattack perturbation such that it maximises the $L_2$ distance with the anchor in the embedding space \cite{pmlr-v235-schlarmann24a}. 
However, pushing the test image away from its embedding risks hurting performance on clean images. 
To address this, we propose \textbf{$\tau$-thresholded weighted counterattacks}, which employ a threshold to prevent further counterattacking if the test image does not exhibit false stability, thus preserving performance on clean images.
To our best knowledge, our paradigm is the first work to utilise the expressive power of CLIP to defend itself from adversarial attacks, and can be categorised as a test-time input purification method for VLMs.
We conduct extensive experiments and analyses on 16 classification datasets, establishing our method as an effective test-time defence for CLIP.
We summarise the main contributions of this paper as follows:
\begin{itemize}
    \item We propose the first method that harnesses the power of CLIP to defend itself from adversarial attacks at inference time without relying on any auxiliary networks.
    Our method is simple and training-free, and can be easily employed in other VLMs.
    \item We propose a test-time counterattack method based on Projected Gradient Descent (PGD). 
    We show that our method can defend CLIP with a small number of counterattack steps without significantly impacting performance on clean images.
    \item 
    We conduct experiments across 16 classification datasets, and demonstrate superior performance compared to test-time defences adapted from existing adversarial robustness literature. Our paradigm can be employed on adversarially finetuned CLIP to further enhance robustness performance at test time.
\end{itemize}


\section{Related Work}
\label{sec:related_work}

\noindent\textbf{Adversarial Robustness.} Since the early development of deep neural networks \cite{krizhevsky2012imagenet,he2016deep,szegedy2015going}, it has been found that they are vulnerable to adversarial attacks. Specifically, a small adversarial perturbation bounded by a $L_p$-radius ball, usually imperceptible by humans, can cause the network to misclassify the sample entirely \cite{szegedy2013intriguing,carlini2017towards}. To address this vulnerability, adversarial training (AT) \cite{madry2018towards,zhang2019theoretically,rice2020overfitting} alternately generates adversarial samples with the target model on the fly, and trains the network with these adversarial samples.
This practice has shown significantly improved robustness to adversarial attacks, and has become a \textit{de facto} standard in adversarial machine learning, despite the presence of limitations such as expensive training \cite{shafahi2019adversarial,wongfast}.
Other types of methods are also proposed, of which the most related to this work is test-time defence.
This can be achieved by employing a generative model to purify the test image with an auxiliary generative model \cite{nie2022diffusion,yoon2021adversarial,samangouei2018defense}, or by adjusting the test image based on an objective \cite{alfarra2022combating,wu2021attacking,mao2021adversarial,hwang2023aid}. 
However, subsequent studies show that test-time defence can be circumvented by adaptive attacks specially designed for the defence \cite{croce2022evaluating}.
Amongst these methods, \textit{Hedge Defense} (HD) \cite{wu2021attacking} is the most closely related to our work. They attack the test image by maximising the cross entropy loss with respect to all classes, based on the finding that the loss surface is smoother around the ground-truth label. 
An important difference between HD and this work is that they apply the defence method on an adversarially trained model.
In contrast, we focus on CLIP and show that foundation models like CLIP possess the inherent ability to defend themselves against attacks that seek to maximise the classification loss, by producing a counterattack perturbation that leads the adversarial image away from its original embedding in the latent space,
with no need for an adversarially trained model.

\noindent\textbf{VLMs and Their Adversarial Robustness.}
Recently, adversarial robustness of foundation models have garnered increasing research attention \cite{zhao2023evaluating,shayegani2023jailbreak}. This paper focuses on enhancing the adversarial robustness of CLIP \cite{radford2021learning} since it is a representative foundation model that aligns images and text.
Existing methods can be divided into two types: (1) \textit{Adversarial finetuning}. \citet{mao2023understanding} propose TeCoA, which finetunes the vision encoder of CLIP using adversarial samples generated on the fly on one dataset, and transfers the learned robustness to downstream classification datasets. Based on this pipeline, \citet{wang2024pre} further propose to employ the original CLIP to guide adversarial finetuning by imposing two regularisation terms, showing improved generalization on clean and adversarial images across downstream datasets.
(2) \textit{Adversarial prompt tuning.} This line of research is built on prompt tuning of CLIP \cite{zhou2022learning,zhou2022conditional}, where model weights are kept frozen. \citet{li2024one} show that textual prompts play an important role in the effectiveness of both adversarial attacks and robustness. They propose to insert learnable tokens and tune the tokens by aligning the textual prompt with adversarial images. \citet{zhang2023adversarial} propose a similar pipeline by training robust text tokens, assuming that the attacker has access to the model but not to the text prompts employed by the end user.
In this work, we discard any training and show that CLIP possesses the ability to defend itself from adversarial attacks by counterattacking adversarial images. 
Our method is the first test-time defence method for CLIP.

\section{Methodology}
\label{sec:methodology}

\begin{figure}[t]
    \centering
    \includegraphics[width=0.8\linewidth]{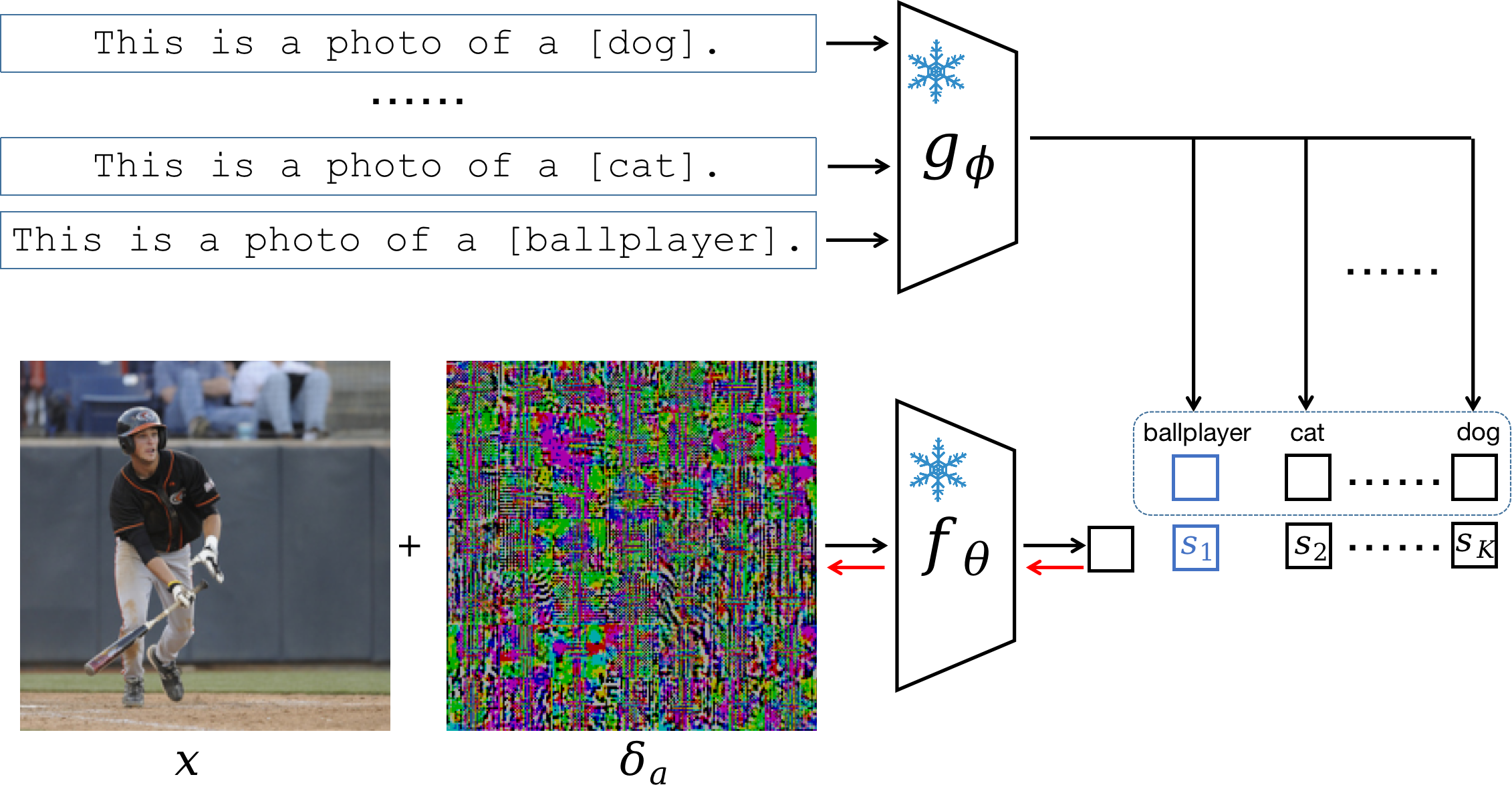}
    \caption{Pipeline to generate an adversarial perturbation $\delta$ given an image $x$ and its ground-truth label based on CLIP. Black and red arrows denote the forward and backward pass, respectively.}
    \label{fig:generate_adv}
    \vspace{-0.5cm}
\end{figure}

\begin{figure}[t]
    \centering
    \includegraphics[width=0.8\linewidth]{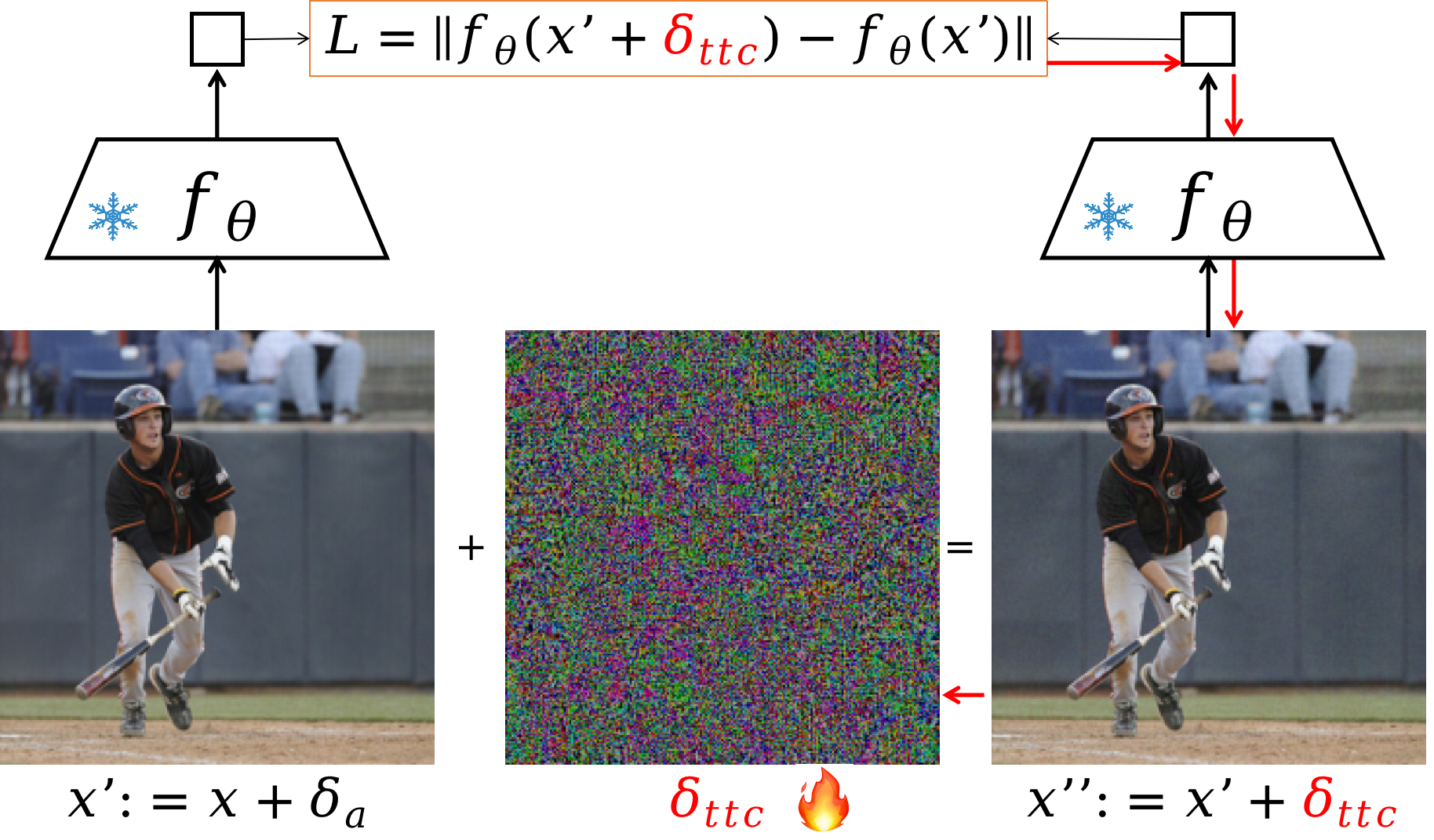}
    \caption{Our test-time counterattack paradigm. We craft a counterattack perturbation $\delta_{ttc}$ to lead an adversarial image away from its original embedding at test time without finetuning.}
    \label{fig:ttc}
    \vspace{-0.5cm}
\end{figure}

In this section, we first provide preliminaries regarding CLIP and adversarial robustness for CLIP in an image classification context. Then we proceed to introduce our test-time counterattack paradigm.

\subsection{Preliminaries and Setup}
\label{sec:preliminary}
\noindent\textbf{Zero-shot classification of CLIP.} 
CLIP \cite{radford2021learning} is a vision-language model that matches images with their descriptive text. It has been contrastively pre-trained on 400 million image-text pairs.
Specifically, it aligns an image $x$ with its corresponding text $t$ through the cosine similarity between their representations produced by a vision encoder $f_\theta(\cdot)$ and a text encoder $g_\phi(\cdot)$, respectively, where $\theta$ and $\phi$ are model weights.
At inference time, CLIP performs classification in a zero-shot manner.
Given a set of classes defined in their textual names $c_1,\dots,c_K$, CLIP matches a test image $x$ against the textual prompts corresponding to candidate class names wrapped by a template $T$ (usually `a photo of [CLASS]'):
\begin{equation}\label{eq:cossim}
    s_i = \frac{f_\theta(x)^T g_\phi(T(c_i))}{\parallel f_\theta(x)\parallel\cdot\parallel g_\phi(T(c_i))\parallel}
\end{equation}
The probability of $x$ belonging to class $c_i$ is calculated as the normalized similarity $p_i=\frac{\exp(s_i)}{\sum_{j}\exp(s_j)}$, and the candidate class with the highest probability is the predicted class.

\noindent\textbf{Adversarial attacks for CLIP.} CLIP is highly vulnerable to adversarial perturbations \cite{mao2023understanding}. In a setting where the attacker has full knowledge of the model weights and gradients of CLIP, a small perturbation $\delta$ bounded by a $L_p$-radius can be maliciously designed to cause CLIP to misclassify:
\begin{equation}\label{eq:delta}
    \delta_a = \arg\max\limits_{\delta} L(x+\delta,t_c), \;\;s.t. \parallel\delta\parallel_p\leq\epsilon_a
\end{equation}
where $t_c$ is the ground-truth label of $x$ and $L$ is a loss function which is usually a cross-entropy loss, and $\epsilon_a$ is the attack budget, which ensures that the manipulation is subtle and imperceptible to humans. \cref{eq:delta} can be approximated by Projected Gradient Descent (PGD) \cite{carlini2017towards}. The adversarial image is the addition of the image and the perturbation $x':=x+\delta_a$. This process is illustrated in \cref{fig:generate_adv}.

\noindent\textbf{Adversarial finetuning} strengthens the adversarial robustness of CLIP \cite{mao2023understanding,wang2024pre} by alternately generating adversarial images $x'$ following \cref{eq:delta} and using them to finetune $f_\theta$.
TeCoA \cite{mao2023understanding} performs finetuning by aligning $x'$ with the ground-truth text $T(t_c)$ on one dataset. PMG-AFT \cite{wang2024pre} imposes two CLIP-guided regularization terms on top of TeCoA to improve the generalization on  clean and adversarial images.
After the finetuning phase, CLIP has learned robustness to adversarial attacks, which transfers to downstream classification datasets without further training \cite{mao2023understanding}.

\subsection{Test-time Counterattacks}
\label{sec:ttc}
Although adversarial finetuning has been shown to significantly improve CLIP's adversarial robustness, limitations are apparent, such as cumbersome training involving the generation of adversarial samples and finetuning of the vision encoder weights $\theta$. 
In this paper, we investigate the ability of CLIP to defend itself at test time, with no need for any training, providing the first test-time defence method for CLIP. Our proposed paradigm, termed as \textbf{Test-time Counterattacks} (TTC), is illustrated in \cref{fig:ttc}. Following previous studies \cite{mao2023understanding,wang2024pre,li2024one,zhang2023adversarial}, we focus on attacks that aim to maximise the classification loss of CLIP.

Intuitively, a pre-trained vision encoder $f_\theta$ is highly expressive in capturing the nuanced pixel pattern in an image. 
In this sense, we speculate that an adversarial image that successfully fools CLIP is trapped in a toxic surrounding induced by the adversary, 
and that the pre-trained $f_\theta$ is able to lead the image away from this surrounding by utilising its expressiveness.
Recently, \citet{pmlr-v235-schlarmann24a} propose an unsupervised attack method for CLIP,
where a perturbation is updated such that it maximises the $L_2$ distance between the image and its original embedding.
Inspired by this label-free attack method, we employ the same loss function in our paradigm.
Specifically, we employ the original image embedding $f_\theta(x)$ as the anchor, and craft a \underline{t}est-\underline{t}ime \underline{c}ounterattack perturbation $\delta_{ttc}$ such that the $L_2$ distance between the embedding of the counterattacked image $f_\theta(x+\delta_{ttc})$ and the anchor $f_\theta(x)$ is maximised:
\begin{equation}\label{eq:ttc}
    \delta_{ttc} = \arg\max\limits_{\delta} \parallel f_\theta(x+\delta) - f_\theta(x) \parallel,
    \; s.t. \parallel \delta \parallel_p \leq \epsilon_{ttc}
\end{equation}
This counterattack can also be approximated by PGD \cite{carlini2017towards}.
Since this counterattack is performed by the end user at test time, the counterattack does not need to be imperceptible, hence a large user-defined counterattack budget $\epsilon_{ttc}$.
However, we hope to maintain a consistent attack style with existing studies, and still keep the counterattack budget low, bounded by a $L_p$-radius. In the experiments (\cref{sec:exp}), we show that a budget at $\epsilon_{ttc}=4/255$ is able to improve CLIP's adversarial robustness significantly.
Note that the vision encoder weights $\theta$ are kept frozen throughout.
Among existing methods for non-foundational models, the most closely related to ours is \textit{hedge defense} (HD) \cite{wu2021attacking}. 
An important difference is that they employ HD on adversarially-trained models, whilst we show that CLIP without adversarial finetuning can harness the expressiveness of its vision encoder to defend itself.

\begin{figure}[t]
    \centering
    \begin{subfigure}[t]{0.22\textwidth}
        \centering
        \includegraphics[width=\linewidth]{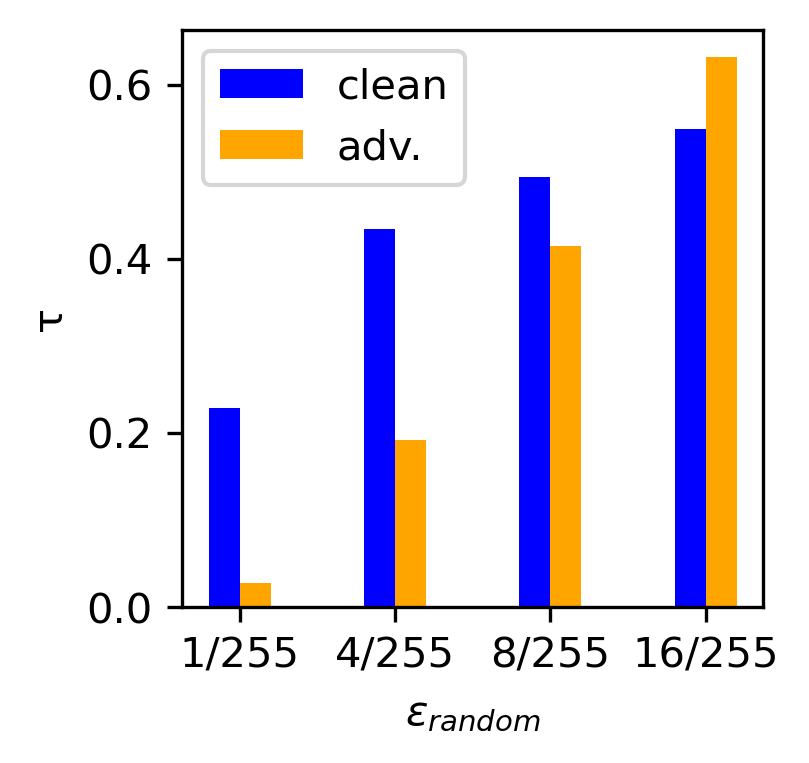}
        \caption{CIFAR10}
        \label{fig:cifar10}
    \end{subfigure}
    \hfill
    \begin{subfigure}[t]{0.22\textwidth}
        \centering
        \includegraphics[width=\linewidth]{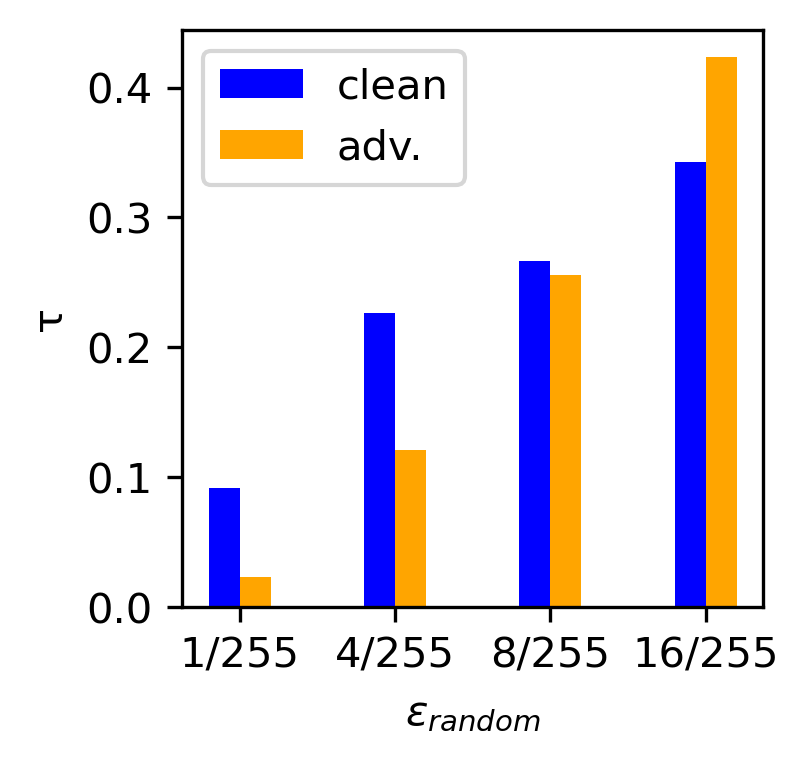}
        \caption{ImageNet}
        \label{fig:imagenet}
    \end{subfigure}
    \caption{Ratio of $L_2$ drift due to a random noise. The value of $\tau$ is the average $\tau$ across 100 randomly selected samples.}
    \label{fig:tau_sensitivity}
    \vspace{-0.5cm}
\end{figure}

\subsection{$\tau$-thresholded Weighted Counterattacks}
\cref{sec:ttc} has discussed the idea of defending CLIP with its pre-trained vision encoder. An undesirable risk is that the counterattacks can hurt natural images as well.
Based on the idea of TTC, we further propose \textbf{$\tau$-thresholded weighted counterattacks} to counterattack adversaries effectively while
reducing the impact on clean images.

\citet{wu2021attacking} show that adversarial images are more vulnerable to a small noise than clean images. 
In this study, we find that adversarial images are actually more robust to small random noises, and are only vulnerable to sufficiently large noises, based on our analysis of adversarial images obtained by iterative attack methods (PGD \cite{carlini2017towards} in our case).
Specifically, we define a stochastic variable $\tau$ induced by a random noise $n\in\mathcal{R}^{C\times W\times H}\thicksim U(-\epsilon_{random}, \epsilon_{random})$, conditioned on an image $x\in\mathcal{R}^{C\times W\times H}$:
\begin{equation}\label{eq:tau}
    \tau = \frac{\parallel f_\theta(x+n) - f_\theta(x)\parallel}{\parallel f_\theta(x)\parallel}
\end{equation}
which can be interpreted as the ratio of the $L_2$ drift in the latent space when a random noise $n$ is applied on an image. 
The values of $\tau$ are reported in \cref{fig:tau_sensitivity} for ImageNet and CIFAR10. We report more results and analysis on $\tau$ for other datasets in Appendix (\cref{sec:tau_analysis}). 
As can be seen from \cref{fig:tau_sensitivity}, when a small random noise ($\epsilon_{random}=1/255,\,4/255$) is imposed, the ratio of $L_2$ drift in the latent space is unusually small, showing that they are trapped in a toxic surrounding and rendered `falsely stable' by an adversary. 
Adversarial images only become vulnerable when the strength of random noise is increased, as evidenced by the disproportionately rising values of $\tau$. 
We term this behaviour of adversarial images obtained by maximising CLIP's classification loss
`false stability', and provide more theoretical analysis in Appendix (\cref{sec:tau_analysis}).

Built upon the analysis above, we propose $\tau$-thresholded weighted counterattacks based on PGD \cite{carlini2017towards}. Specifically, we follow a standard pipeline of PGD iterations, with the attack objective being \cref{eq:ttc}. At the \textit{zero}-th iteration, a random perturbation without any update $\delta_{ttc}^0$ is applied, where we compute the $\tau$ value based on \cref{eq:tau} as an indicator. If it is higher than a user-defined threshold $\tau_{thres}$, meaning that it is not `falsely stable', we halt the counterattack and return the random noise $\delta_{ttc}^0$. Otherwise, the counterattack is resumed. 
Note that the selection of $\tau_{thres}$ is dependent only on the $\tau$ value of clean images and the strength of random noise, irrespective of types and strengths of attacks.
Since employing only one $\delta_{ttc}$ may be suboptimal, we weight the counterattack perturbation vectors across all steps:
\begin{equation}\label{eq:weight}
    w_j = \frac{\exp(\beta\cdot j)}{\sum_{j=0}^{N}\exp(\beta\cdot j)}
\end{equation}
\begin{equation}\label{eq:weighted_ttc}
    \delta_{ttc}=\sum_{j=0}^{N}{w_j\delta_{ttc}^{j}}
\end{equation}
where $\beta>0$ is a hyperparameter controlling the ascending rate of weights, $N$ is the number of steps for performing the counterattack, and $\delta_{ttc}^{j}$ is the counterattack perturbation obtained after $j$ steps. We summarise our $\tau$-thresholded weighted counterattacks in \cref{alg:ttc}.

\begin{algorithm}[t]
    \caption{$\tau$-thresholded weighted counterattacks.}\label{alg:ttc}
    \begin{algorithmic}[1]
    \Require Test image $x$, pre-trained CLIP vision encoder $f_\theta$, 
    counterattack budget $\epsilon_{ttc}$,  stepsize $\alpha$, number of steps $N$, user-defined parameters $\tau_{thres}$ and $\beta$.
    \Procedure{Test-time Counterattacks}{}
        \State $\delta_{ttc}^{0}\thicksim U(-\epsilon_{ttc}, \epsilon_{ttc})$.
        \State Compute $\tau$ based on \cref{eq:tau} using $\delta_{ttc}^{0}$.
        \If{$\tau\geq\tau_{thres}$}
            \State $w_0=1$
            \State \Return $\delta_{ttc}=\delta_{ttc}^0$
        \ElsIf{$\tau<\tau_{thres}$}
            \State $\textbf{w},\bm{\delta}_{ttc}:=\{\},\{\}$
            \For{$i=1,2,\dots,N$}
                \State $\delta_{ttc}^{i}$$=$$\Pi(\delta_{ttc}^{i-1}+\alpha\nabla_{\delta}\Vert f_\theta(x+\delta_{ttc}^{i-1})-f_\theta(x)\Vert)$  
                \State $w_i=\exp(\beta\cdot i)/\sum_{j=0}^{N}\exp(\beta\cdot j)$ (\cref{eq:weight})
                \State $\textbf{w}\leftarrow w_i,\;\bm{\delta}_{ttc}\leftarrow \delta_{ttc}^i$
            \EndFor
            \State \Return $\delta_{ttc}=\sum_{i=0}^{N}w_i\cdot\delta_{ttc}^{i}$ (\cref{eq:weighted_ttc})
        \EndIf
    \EndProcedure
    \end{algorithmic}
\end{algorithm}

\section{Experiments}
\label{sec:exp}

In this section, we conduct extensive experiments to verify the effectiveness of our test-time counterattack paradigm. 

\subsection{Experiment setup}
\noindent\textbf{Datasets.} 
Following previous work that studies adversarial robustness of CLIP \cite{mao2023understanding,wang2024pre}, we conduct our experiments on 16 datasets, which include general object recognition datasets 
CIFAR10 \cite{krizhevsky2009learning}, 
CIFAR100 \cite{krizhevsky2009learning}, 
STL10 \cite{coates2011analysis}, 
ImageNet \cite{deng2009imagenet}, 
Caltech101 \cite{fei2006one} and 
Caltech256 \cite{griffin2007caltech};
fine-grained recognition datasets OxfordPets \cite{parkhi2012cats}, 
Flowers102 \cite{nilsback2008automated}, 
Food101 \cite{bossard2014food},
StanfordCars \cite{krause20133d}; 
scene recognition datasets SUN397 \cite{xiao2010sun},
Country211 \cite{radford2021learning};
domain-specific datasets FGVCAircraft \cite{maji2013fine},
EuroSAT \cite{helber2019eurosat},
DTD \cite{cimpoi2014describing},
PCAM \cite{bejnordi2017diagnostic}. 
All datasets are preprocessed by the same pre-processing pipeline of CLIP \cite{radford2021learning}.

\noindent\textbf{Implementation Details.} We use a counterattack budget of $\epsilon_{ttc}=4/255$ and a threshold $\tau_{thres}=0.2$, which is selected based on clean images. We set the number of steps for counterattacks as $N=2$, unless otherwise stated. $\beta$ is set to $2.0$. All attacks and counterattacks in experiments are bounded by a $L_\infty$ radius. 

\noindent\textbf{Baselines.} Since there are no test-time defence methods for CLIP, we implement several test-time methods from existing adversarial robustness studies that do not rely on auxiliary networks. Specifically, we implement \textit{Anti-adversary} \cite{alfarra2022combating} and \textit{Hedge Defense} (HD) \cite{wu2021attacking}, which are the most closely related to our method. 
\textit{Anti-adversary} \cite{alfarra2022combating} generates a perturbation to reinforce the confidence of the classifier given a test image. We adapt it in CLIP by increasing the cosine similarity between the image and the text with the highest cosine similarity. 
HD \cite{wu2021attacking} performs a counterattack on the test image by increasing the cross-entropy w.r.t. all candidate classes, based on their finding that the loss function surface is smoother around the ground-truth class. We also adapt their method in our experiments with CLIP. 
For these two methods, we employ a test-time perturbation budget of $4/255$, which is equal to the counterattack budget of our TTC. Following the original papers, the number of steps are 2 and 20 for \textit{Anti-adversary} and HD, respectively.
Considering that previous studies establish image transformations as a simple and effective defence method \cite{xie2018mitigating,perez2021enhancing,guo2018countering},
we also implement \textit{test-time transformation ensembling} (TTE) \cite{perez2021enhancing},
which ensembles image transformation as a defence. We implement TTE with image flip, 4 crops, and image flip of all crops, totaling 9 augmentative views, which we find to provide the best performance. 
As a simplest baseline, we also include \textit{random noise} (RN) which adds a random perturbation noise with the same strength as our $\epsilon_{ttc}$, i.e., $n\sim U(-\epsilon_{ttc}, \epsilon_{ttc})$.
Although our paradigm can be categorised as test-time defence, we also implement adversarially finetuning methods as a useful reference. We implement TeCoA \cite{mao2023understanding}, PMG-AFT \cite{wang2024pre} and FARE \cite{pmlr-v235-schlarmann24a} by finetuning the CLIP vision encoder with adversarial images on TinyImageNet, based on the objective functions proposed in their papers\footnote{Unlike original implementations, we randomly hold out 10\% of the training set of TinyImageNet for evaluation in our implementation, without consulting downstream datasets. We also find 
preprocessing significantly affects the performance of finetuned models on 
CIFAR10, CIFAR100, and STL10. We follow the preprocessing pipeline recommended by CLIP for all datasets (\cref{tab:white_box_eps1}).}. 
We also finetune CLIP with clean images (CLIP-FT) on TinyImageNet.
In the phase of finetuning, we use a 2-step PGD attack, with the stepsize $\alpha=1/255$ and attack budget $\epsilon=1/255$, following \cite{mao2023understanding,wang2024pre}. The learning rate for finetuning is $5e-5$.
After the finetuning phase, the finetuned models are deployed on 16 downstream datasets.

\begin{table*}[t]
    \centering
    \resizebox{\textwidth}{!}{
    \begin{tabular}{cc|c|cccc||ccccc|c}
         \toprule
         
         \multicolumn{2}{c|}{\multirow{2}{*}{(\%)}} & \multirow{2}{*}{CLIP} & \multicolumn{4}{c||}{Adversarial Finetuning} & \multicolumn{5}{c|}{Test-time Defence} & \multirow{2}{*}{$\Delta$} \\
         \cline{4-7}\cline{8-12}
         & & &  CLIP-FT & TeCoA & PMG-AFT & FARE & RN & TTE & Anti-adv & HD & TTC (ours) & \\
         \midrule
         \color{lightgray}\multirow{2}{*}{TinyImageNet} & \color{lightgray}Rob. & \color{lightgray}0.19 & \color{lightgray}2.19 & \color{lightgray}48.64 & \color{lightgray}46.12 & \color{lightgray}25.47 &\color{lightgray}0.28$\pm0.02$ &\color{lightgray}19.52$\pm4.21$ &\color{lightgray}4.46$\pm0.23$ &\color{lightgray}3.11$\pm0.05$ &\color{lightgray}20.64$\pm0.17$ &\color{lightgray}+20.45 \\
         & \color{lightgray}Acc.& \color{lightgray}57.64 & \color{lightgray}77.06 & \color{lightgray}70.86 &\color{lightgray}66.85 &\color{lightgray}73.63 &\color{lightgray}51.83$\pm0.16$ &\color{lightgray}56.74$\pm0.22$ &\color{lightgray}52.55$\pm0.06$ &\color{lightgray}51.37$\pm0.15$ &\color{lightgray}51.84$\pm0.17$ &\color{lightgray}-5.80\\
         \midrule
         \multirow{2}{*}{CIFAR10} & Rob. & 0.74 & 3.34 & 33.61 & 40.66 & 19.65 & 2.01\std{0.08} &\textbf{41.35}\std{6.14} &12.39\std{0.07} &17.22\std{0.45} &28.75\std{0.18} &\redd{+28.01} \\
         & Acc.& 85.12 & 84.90 & 64.61 & 70.69 & 74.44 &81.18\std{0.07} &\textbf{84.74}\std{0.40} &83.52\std{0.09} &78.23\std{0.16} &81.18\std{0.07} &\blue{-3.94} \\
         \midrule
         \multirow{2}{*}{CIFAR100} & Rob. & 0.26 & 0.90 & 18.95 & 22.52 & 11.40 & 0.67\std{0.05} &\textbf{20.06}\std{4.03} &5.73\std{0.04} &3.86\std{0.10} &14.31\std{0.25} &\redd{+14.05} \\
         & Acc.& 57.14 & 59.51 & 35.96 & 40.32 & 46.67 & 56.34\std{0.20} &\textbf{58.61}\std{0.25} &53.95\std{0.15} &52.86\std{0.16} &56.34\std{0.20} &\blue{-0.80} \\
         \midrule
         \multirow{2}{*}{STL10} & Rob. & 11.0 & 12.73 & 70.08 & 73.08 & 59.06 & 16.23\std{0.08} &\textbf{78.48}\std{3.83} &37.42\std{0.40} &39.02\std{0.30} &76.70\std{0.23} &\redd{+65.70} \\
         & Acc.& 96.40 & 94.49 & 87.40 & 88.56 & 91.72 & 95.85\std{0.04} &\textbf{96.26}\std{0.04} &95.45\std{0.08} &89.50\std{0.07} &95.85\std{0.04}&\blue{-0.55} \\
         \midrule
         \multirow{2}{*}{ImageNet} & Rob. & 1.15 & 0.93 & 18.89 & 21.43 & 14.00 &1.77\std{0.03} &31.01\std{4.40} &8.67\std{0.05} &6.63\std{0.05} &\textbf{38.41}\std{0.07}&\redd{+37.26} \\
         & Acc.& 59.69 & 54.24 & 34.89 & 36.12 & 48.79 &59.34\std{0.06} &\textbf{60.02}\std{0.12} &54.27\std{0.14} &54.54\std{0.05} &49.39\std{0.00} &\blue{-10.30} \\
         \midrule
         \multirow{2}{*}{Caltech101} & Rob. & 14.67 & 14.21 & 55.51 & 61.08 & 50.74 &18.90\std{0.14} &\textbf{67.56}\std{3.88} &34.81\std{0.16} &31.53\std{0.22} &65.78\std{0.07} &\redd{+51.11} \\
         & Acc.& 85.66 & 83.63 & 71.68 & 75.45 & 80.95 &\textbf{86.61}\std{0.10} &85.84\std{0.09} &84.02\std{0.10} &82.33\std{0.04} &86.53\std{0.07} &\redd{+0.87} \\
         \midrule
         \multirow{2}{*}{Caltech256} & Rob. & 8.47 & 6.76 & 43.19  & 45.91 & 38.79 &11.33\std{0.04} &60.09\std{4.03} &25.36\std{0.17} &23.48\std{0.10} &\textbf{60.11}\std{0.04} &\redd{+51.64} \\
         & Acc.& 81.72 & 78.53 & 61.14 & 62.24 & 73.32 &81.25\std{0.03} &\textbf{82.49}\std{0.08} &79.38\std{0.12} &79.12\std{0.01} &79.66\std{0.04} &\blue{-2.06} \\
         \midrule
         \multirow{2}{*}{OxfordPets} & Rob. & 1.04 & 2.10 & 38.35 & 41.18 & 31.07 &1.86\std{0.01} &50.33\std{7.30} &20.42\std{0.22} &12.04\std{0.16} &\textbf{57.87}\std{0.15} &\redd{+56.83} \\
         & Acc.& 87.44 & 84.14 & 62.12 & 65.88 & 79.37 &87.41\std{0.12} &\textbf{88.13}\std{0.13} &80.62\std{0.35} &80.91\std{0.05} &83.35\std{0.21} &\blue{-4.09} \\
         \midrule
         \multirow{2}{*}{Flowers102} & Rob. & 1.14 & 0.54 & 21.94 & 23.43 & 17.14 &1.52\std{0.01} &35.88\std{4.72} &7.16\std{0.41} &7.29\std{0.06} &\textbf{39.14}\std{0.28} &\redd{+38.00} \\
         & Acc.& 65.46 & 53.37 & 36.80 & 37.00 & 47.98 &64.62\std{0.19} &\textbf{65.18}\std{0.22} &62.66\std{0.14} &58.22\std{0.12} &64.16\std{0.19} &\blue{-1.30} \\
         \midrule
         \multirow{2}{*}{FGVCAircraft} & Rob. & 0.00 & 0.00 & 2.49 & 2.22 & 1.35 &0.00\std{0.00} &6.23\std{1.37} &1.27\std{0.07} &1.26\std{0.07} &\textbf 
         {13.77}\std{0.38} &\redd{+13.77} \\
         & Acc.& 20.10 & 14.04 & 5.31 & 5.55 & 10.86 &19.25\std{0.18} &\textbf{20.19}\std{0.36} &15.88\std{0.23} &16.36\std{0.03} &18.00\std{0.16} &\blue{-2.10} \\
         \midrule
         \multirow{2}{*}{StanfordCars} & Rob. & 0.02 & 0.06 & 8.76 & 11.65 & 6.75 &0.16\std{0.02} &22.36\std{4.17} &4.40\std{0.30} &2.71\std{0.09} &\textbf{33.01}\std{0.07} &\redd{+32.99} \\
         & Acc.& 52.02 & 42.11 & 20.91 & 25.44 & 38.68 &52.14\std{0.09} &\textbf{52.73}\std{0.31} &36.21\std{0.27} &44.28\std{0.02} &48.16\std{0.16} &\blue{-3.86} \\
         \midrule
         \multirow{2}{*}{SUN397} & Rob. & 1.14 & 0.94 & 19.39 & 22.58 & 14.91 &1.72\std{0.01} &30.79\std{4.43} &8.05\std{0.04} &6.40\std{0.06} &\textbf{41.52}\std{0.04} &\redd{+40.38} \\
         & Acc.& 58.50 & 55.73 & 36.69 & 37.98 & 52.42 & \textbf{59.69}\std{0.06} &59.12\std{0.08} &56.00\std{0.04} &53.17\std{0.02} &55.13\std{0.06} &\blue{-3.37} \\
         \midrule
         \multirow{2}{*}{Country211} & Rob. & 0.04 & 0.03 & 1.78 & 2.12 & 0.85 & 0.06\std{0.00} &3.05\std{0.89} &0.67\std{0.05} &0.47\std{0.02} &\textbf{7.09}\std{0.04} &\redd{+7.05} \\
         & Acc.& 15.25 & 12.07 & 4.75 & 4.64 & 9.26 & \textbf{14.80}\std{0.02} &14.66\std{0.16} &11.58\std{0.12} &11.72\std{0.07} &13.08\std{0.05} &\blue{-2.17} \\
         \midrule
         \multirow{2}{*}{Food101} & Rob. & 0.70 & 0.42 & 13.90 & 18.57 & 11.65 &1.20\std{0.01} &43.94\std{6.97} &13.12\std{0.16} &8.03\std{0.11} &\textbf{57.84}\std{0.15} &\redd{+57.14} \\
         & Acc.& 83.88 & 64.86 & 29.98 & 36.61 & 55.31 &83.44\std{0.04} &\textbf{83.96}\std{0.02} &75.81\std{0.22} &80.30\std{0.05} &82.18\std{0.02}&\blue{-1.70} \\
         \midrule
         \multirow{2}{*}{EuroSAT} & Rob. & 0.03 & 0.04 & 11.96 & 12.60 & 10.67 &0.15\std{0.01} &6.91\std{2.13} &2.15\std{0.04} &4.57\std{0.09} &\textbf{12.19}\std{0.24}&\redd{+12.16} \\
         & Acc.& 42.59 & 27.64 & 16.58 & 18.53 & 21.88 &\textbf{53.24}\std{0.09} &44.38\std{1.60} &36.78\std{0.18} &39.08\std{0.06} &\textbf{53.24}\std{0.09}&\redd{+10.65} \\
         \midrule
         \multirow{2}{*}{DTD} & Rob. & 2.98 & 2.39 & 17.61 & 14.95 & 15.64 &3.71\std{0.09} &23.90\std{2.34} &5.62\std{0.07} &11.63\std{0.17} &\textbf{27.32}\std{0.25} &\redd{+24.34} \\
         & Acc.& 40.64 & 36.49 & 25.16 & 21.76 & 32.07 &37.96\std{0.13} &\textbf{41.33}\std{0.32} &38.92\std{0.22} &34.89\std{0.35} &36.98\std{0.21}&\blue{-3.66} \\
         \midrule
         \multirow{2}{*}{PCAM} & Rob. & 0.08 & 1.11 & 48.24 & 46.18 & 16.23 &0.41\std{0.01} &10.62\std{3.22} &4.97\std{0.12} &44.74\std{0.17} &\textbf{52.85}\std{0.20}&\redd{+52.77} \\
         & Acc.& 52.02 & 47.21 & 49.96 & 50.03 & 52.54 & \textbf{52.73}\std{0.07} &51.01\std{0.08} &52.49\std{0.02} &50.38\std{0.04} &\textbf{52.73}\std{0.07}&\redd{+0.71} \\
         \midrule
         \midrule
         \multirow{2}{*}{\textbf{Avg.}} & Rob. & 2.70 & 2.91 & 26.54 & 28.76 & 20.00 &3.86\std{0.02} &33.28\std{3.98} &12.01\std{0.04} &13.81\std{0.06} &\textbf{39.17}\std{0.02}&\redd{+36.47} \\
         & Acc. & 61.51 & 55.80 & 40.25 & 42.30 & 51.02 &61.61\std{0.03} &\textbf{61.79}\std{0.13} &57.35\std{0.03} &56.62\std{0.02} &59.75\std{0.06}&\blue{-1.76} \\
         \bottomrule
    \end{tabular}}
    \caption{\label{tab:white_box_eps1}Classification accuracy (\%) on both adversarial images (Rob.) under 10-step PGD attack at $\epsilon_a=1/255$ and clean images (Acc.) across 16 datasets. We include the results on TinyImageNet because it is used to finetune the model for CLIP-FT, TeCoA \cite{mao2023understanding}, PMG-AFT \cite{wang2024pre}, and FARE \cite{pmlr-v235-schlarmann24a}. 
    Weights and gradients of the deployed model are assumed to be known to the threat model. Comparison is made among our paradigm and test-time defences adapted from existing adversarial studies, with finetuning-based models implemented as a reference. We report the mean and standard deviation for test-time methods over 3 runs. The last column reports the gains w.r.t. original CLIP without any finetuning or test-time operations.}
    \vspace{-0.6cm}
\end{table*}

\begin{table}[h]
    \centering
    \resizebox{0.65\linewidth}{!}{
    \begin{tabular}{c|cc}
    \toprule
    (\%) & Rob. & Acc. \\
    \midrule
    CLIP & 0.09  & 61.51 \\
    \midrule
    CLIP-FT & 0.96 & 55.80 \\
    $\textrm{TeCoA}^1$ \cite{mao2023understanding} & 6.51 & 40.25 \\
    $\textrm{TeCoA}^4$ \cite{mao2023understanding} & 10.03 & 35.57 \\
    $\textrm{PMG-AFT}^1$ \cite{wang2024pre} & 7.03 & 42.30 \\
    $\textrm{PMG-AFT}^4$ \cite{wang2024pre}& 10.70 & 37.58 \\
    $\textrm{FARE}^1$ \cite{pmlr-v235-schlarmann24a}& 1.50 & 51.02 \\
    $\textrm{FARE}^4$ \cite{pmlr-v235-schlarmann24a}& 3.67 & 46.17 \\
    \midrule
    RN & 0.06\std{0.00} &61.61\std{0.03} \\
    TTE \cite{perez2021enhancing}& 7.79\std{3.23} &\textbf{61.79}\std{0.13} \\
    Anti-adv \cite{alfarra2022combating}&0.53\std{0.00} &57.32\std{0.03} \\
    HD \cite{wu2021attacking}&1.19\std{0.01} &56.62\std{0.02} \\
    TTC (ours) &\textbf{20.63}\std{0.05} &55.99\std{0.06} \\
    \midrule
    $\Delta$ &\redd{+20.54} &\blue{-5.52} \\
    \bottomrule
    \end{tabular}}
    \caption{Classification accuracy (\%) on adversarial images (Rob.) under 10-step PGD at $\epsilon_a=4/255$ and clean images (Acc.) averaged on 16 datasets. The finetuning-based methods are implemented as a reference, with the superscript indicating the attack budget used in the finetuning phase.
    The last row reports the gains compared to the original CLIP.}
    \label{tab:white_box_eps4}
    \vspace{-0.5cm}
\end{table}

\begin{table*}[t!]
    \centering
    \resizebox{\textwidth}{!}{
    \begin{tabular}{c|cccccccccccccccc|cc}
    \toprule
    (\%) & \rotatebox{60}{CIFAR10} & \rotatebox{60}{CIFAR100} & \rotatebox{60}{STL10} & \rotatebox{60}{ImageNet} & \rotatebox{60}{Caltech101} & \rotatebox{60}{Caltech256} & \rotatebox{60}{OxfordPets} & \rotatebox{60}{Flower102} & \rotatebox{60}{FGVCAircraft} & \rotatebox{60}{StanfordCars} & \rotatebox{60}{SUN397} & \rotatebox{60}{Country211} & \rotatebox{60}{Food101} & \rotatebox{60}{EuroSAT} & \rotatebox{60}{DTD} & \rotatebox{60}{PCAM} & \rotatebox{60}{Avg. Rob.} & \rotatebox{60}{Avg. Acc.}\\
    \midrule
    TeCoA & 33.61 & 18.95 & 70.08 & 18.89 & 55.51 & 43.19 & 38.35 & 21.94 & 2.49 & 8.76 & 19.39 & 1.78 & 13.90 & 11.96 & 17.61 & 48.24 & 26.54 & 40.25\\
    TeCoA+TTC & 34.81 & 20.09 & 71.40 & 23.22 & 58.90 & 48.52 & 42.25 & 25.04 & 3.00 & 11.95 & 23.86 & 2.51 & 17.67 & 12.73 & 19.36 & 48.39 & 29.06 &  39.81 \\
    \cline{2-19}
    $\Delta$ & $\redd{1.20\uparrow}$ & $\redd{1.14\uparrow}$ & $\redd{1.32\uparrow}$ & $\redd{4.33\uparrow}$ & $\redd{3.39\uparrow}$ & $\redd{5.33\uparrow}$ & $\redd{3.90\uparrow}$ & $\redd{3.10\uparrow}$ & $\redd{0.51\uparrow}$ & $\redd{3.19\uparrow}$ & $\redd{4.47\uparrow}$ & $\redd{0.73\uparrow}$ & $\redd{3.77\uparrow}$ & $\redd{0.77\uparrow}$ & $\redd{1.75\uparrow}$ & $\redd{0.15\uparrow}$ & $\redd{2.52\uparrow}$ & $\blue{-0.44\downarrow}$ \\
    \midrule
    PMG-AFT & 40.66 & 22.52 & 73.08 & 21.43 & 61.08 & 45.91 & 41.18 & 23.43 & 2.22 & 11.65 & 22.58 & 2.12 & 18.57 & 12.60 & 14.95 & 46.18 & 28.76 & 42.30 \\
    PMG-AFT+TTC & 42.30 & 24.23 & 73.19 & 24.32 & 63.67 & 49.99 & 44.73 & 25.63 & 3.12 & 15.09 & 25.74 & 2.57 & 22.55 & 13.89 & 15.53 & 46.35 & 30.81 & 41.89 \\
    \cline{2-19}
    $\Delta$ & $\redd{1.64\uparrow}$ & $\redd{1.77\uparrow}$ & $\redd{0.11\uparrow}$ & $\redd{2.89\uparrow}$ & $\redd{2.59\uparrow}$ & $\redd{4.08\uparrow}$ & $\redd{3.55\uparrow}$ & $\redd{2.20\uparrow}$ & $\redd{0.90\uparrow}$ & $\redd{3.44\uparrow}$ & $\redd{3.16\uparrow}$ & $\redd{0.45\uparrow}$ & $\redd{3.98\uparrow}$ & $\redd{1.29\uparrow}$ & $\redd{0.58\uparrow}$ & $\redd{0.17\uparrow}$ & $\redd{2.05\uparrow}$ & $\blue{-0.41\downarrow}$ \\
    \midrule
    FARE & 19.65 & 11.40 & 59.06 & 14.00 & 50.74 & 38.79 & 31.07 & 17.14 & 1.35 & 6.85 & 14.90 & 0.85 & 11.65 & 10.67 & 15.64 & 16.23 & 20.00 & 51.02 \\
    FARE+TTC & 35.84 & 21.88 & 76.51 & 30.59 & 67.97 & 59.10 & 51.46 & 29.40 & 5.28 & 20.64 & 33.48 & 3.98 & 32.20 & 15.46 & 22.55 & 35.35 & 33.85 & 49.92\\ 
    \cline{2-19}
    $\Delta$ & $\redd{16.19\uparrow}$ & $\redd{10.48\uparrow}$ & $\redd{17.45\uparrow}$ & $\redd{16.59\uparrow}$ & $\redd{17.23\uparrow}$ & $\redd{20.31\uparrow}$ & $\redd{20.39\uparrow}$ & $\redd{12.26\uparrow}$ & $\redd{3.93\uparrow}$ & $\redd{13.79\uparrow}$ & $\redd{18.58\uparrow}$ & $\redd{3.13\uparrow}$ & $\redd{20.55\uparrow}$ & $\redd{4.79\uparrow}$ & $\redd{6.91\uparrow}$ & $\redd{19.12\uparrow}$ & $\redd{13.85\uparrow}$ & $\blue{-1.1\downarrow}$\\

    \bottomrule
    \end{tabular}}
    \caption{TTC employed on adversarially finetuned models at test time. We report the robust accuracy at $\epsilon_a=1/255$ and the robustness gain by employing TTC for each dataset.}
    \label{tab:ttc_robust_models}
\end{table*}

\subsection{TTC on Original CLIP}

\noindent \textbf{Robustness under $\epsilon_a=1/255$.} 
We first test the robustness of all methods
under the attack budget of $\epsilon_{a}=1/255$.
Following previous studies on CLIP's adversarial robustness \cite{mao2023understanding,wang2024pre}, we test all baselines under 10-step PGD attacks across 16 datasets, assuming that the attacker has full access to the weights and gradients of the deployed model, but not to the test-time operations made by the end user. We report the accuracy on both adversarial images and clean images in \cref{tab:white_box_eps1}. 
It can be seen that all finetuning-based methods overfit to the dataset used for adversarial finetuning to varying extents, as evidenced by the higher accuracy of clean images than the original CLIP on TinyImageNet. The improved robustness on downstream datasets comes at a cost of a noticeable clean accuracy drop.
Among test-time methods, both Anti-adversary and HD, which generate an additive perturbation based on an objective, lead to limited improvement of robust accuracy. Our TTC, which utilises the pre-trained vision encoder of CLIP to produce counterattacks, shows the best robust accuracy on most downstream datasets, usually with a large gain. We also retain the best clean accuracy compared to these two perturbation update methods.
Adding random noise (RN) brings little robustness,  even though the added noise is four times larger than the attack budget, i.e., $\epsilon_{ttc}\gg\epsilon_a$.
RN can be viewed as a special case of TTC with the number of $N$ being 0. 
By exploiting the pre-trained model $f_\theta$ to optimize the noise, TTC significantly improves robustness.
TTE ensembles a number of image transformations, which improves CLIP's robustness at test time to an average accuracy of 33.28\%. However, this gain is generally unstable across runs, as indicated by the high standard deviation of robust accuracy. 
Overall, our proposed TTC leads to consistent gains on robust accuracy (+36.47\%) averaged on downstream datasets with a slight loss (-1.76\%) on clean accuracy compared to the original CLIP,
serving as a stable defence at inference time. We test the robustness under CW attacks \cite{carlini2017towards} in Appendix (\cref{sec:cw_attacks}) for limited space.

\noindent\textbf{Robustness under $\epsilon_a=4/255$.} We further test the robustness of all methods under a high attack budget $\epsilon_a=4/255$. For this setting, we increase the number of steps $N$ to 5 for more effective counterattacks, while other hyperparameters are unchanged. We also implement finetuning-based methods with attack budget $\epsilon=4/255$ during finetuning, in this setting. 
We report the average accuracy across 16 datasets in \cref{tab:white_box_eps4} and provide the full table in Appendix (\cref{tab:white_box_eps4_full}). It can be seen that a high attack budget at $\epsilon_a=4/255$ degrades the accuracy of all models to a very low level. 
Anti-Adversary \cite{alfarra2022combating} and HD \cite{wu2021attacking} provide little to no robustness under this setting. TTE defends the model to a limited extent, but still with low reliability as indicated by the high standard deviation. 
In comparison, our proposed TTC provides a stable robustness gain averaged on 16 datasets.

\subsection{TTC on Adversarially Finetuned CLIP}
Since our method performs counterattacks using the victim model at test time, it can also be employed on adversarially finetuned models in a plug-in manner. 
In this section, we apply TTC to finetuning-based models, assuming that the attacker has full access to the deployed model, but not to the operations by the end user.
Note that we still employ the original vision encoder $f_\theta$ of CLIP to compute $\tau$ (\cref{eq:tau}), because the sensitivity of adversarial finetuned vision encoders is largely reduced.
We report the results in \cref{tab:ttc_robust_models}. 
It can be seen that TTC can further boost adversarial robustness by exploiting the finetuned model to perform counterattacks at test time.
Specifically, TTC achieves a robustness accuracy of $29.06\%$ and $30.81\%$ when employed on TeCoA and PMG-AFT, surpassing the original finetuned models by $2.52$ and $2.05$ points, respectively. 
A significant gain of $13.85$ points is achieved when we employ TTC on top of FARE, an unsupervised adversarially finetuned model. 
Interestingly, we find that adversarial finetuning greatly reduces the sensitivity of CLIP to variations in the pixel space, thus hurting the expressive power of the pre-trained encoder. We provide more in-depth analyses of such loss in Appendix (\cref{sec:pitfalls_aft}). 
Since our counterattacks rely heavily on the expressiveness of the pre-trained vision encoder $f_\theta$, this also explains the smaller gains achieved on adversarially finetuned models, compared to the original CLIP. 
The larger increase of robust accuracy on FARE implies that adversarially finetuning CLIP in an unsupervised manner better retains the expressiveness of the model.

\subsection{Ablation studies}
We experimentally find that the number of steps $N$ of our TTC greatly affects performance on both adversarial and clean images. 
A general rule of thumb is that an attack with a higher budget $\epsilon_a$ would require more steps of counterattacks.
In this section, we investigate the effect of $N$ and keep the other hyperparameters unchanged. We provide analysis on other hyperparameters in Appendix (\cref{sec:other_hyperparams}). 
\cref{fig:N_effect} provides the performance of CLIP employing TTC on 12 datasets as $N$ varies.
As can be seen from \cref{fig:N_effect}, for smaller attacks at $\epsilon_a=1/255$, it takes fewer than three steps for CLIP to defend itself effectively on most datasets. 
Excessive counterattacks can impair the images, as evidenced by the decline after a certain number of steps.
In comparison, a strong attack $\epsilon_a=4/255$ requires a larger number of counterattack steps to reach a reasonable accuracy, showing that they are more resilient to counterattacks by the user side. 
TTC does not impact accuracy on clean images significantly on most datasets, except for SUN397 (\cref{fig:sun_N}), OxfordPets (\cref{fig:pets_N}), and ImageNet (\cref{fig:imagenet_N}), where clean images are found sensitive to the increase of $N$.

\begin{figure*}[t]
    \centering
    \begin{subfigure}{0.15\textwidth} 
        \centering
        \includegraphics[width=\textwidth]{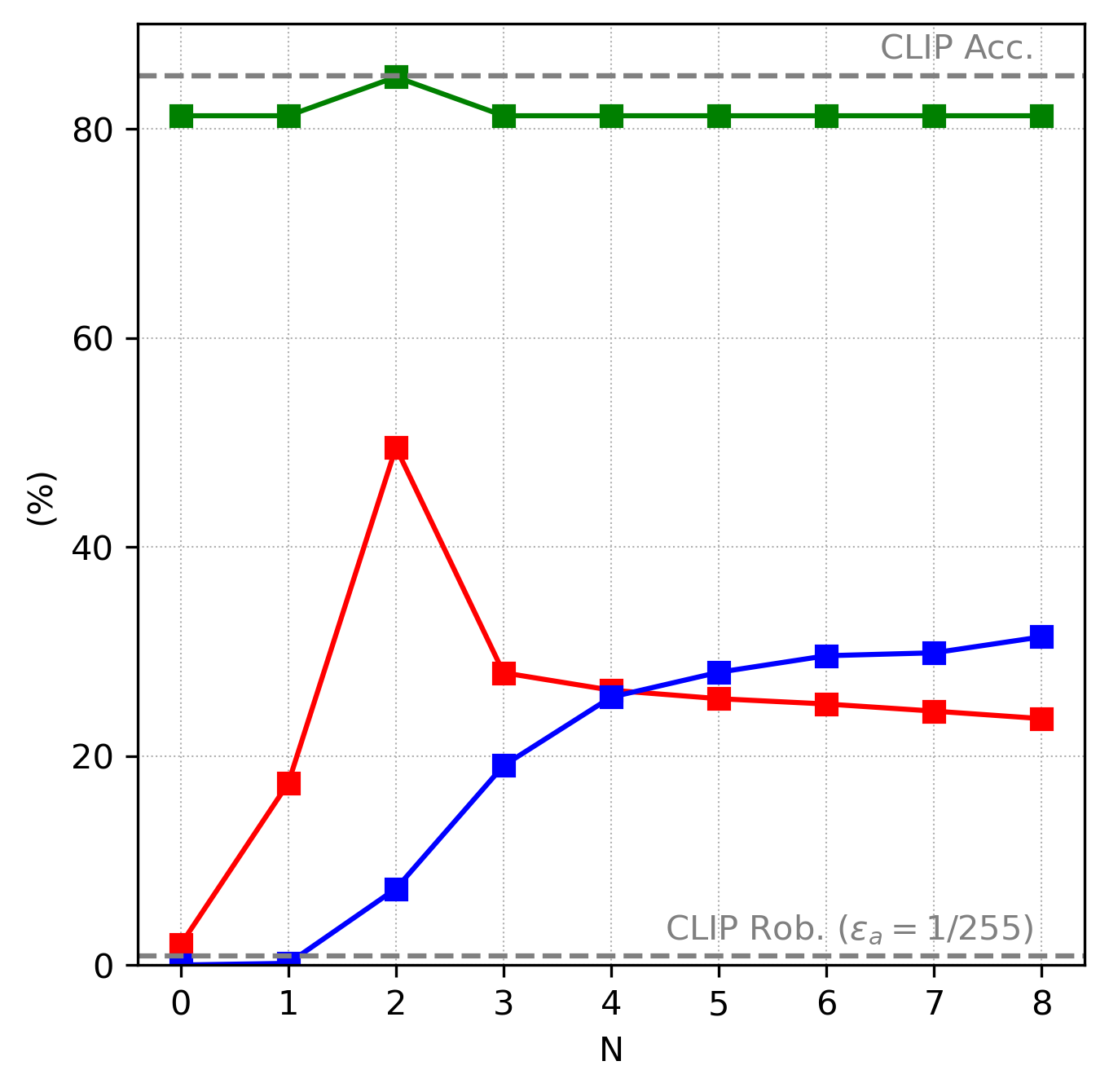}
        \caption{CIFAR10}
        \label{fig:cifar10_N}
    \end{subfigure}
    \hfill
    \begin{subfigure}{0.15\textwidth} 
        \centering
        \includegraphics[width=\textwidth]{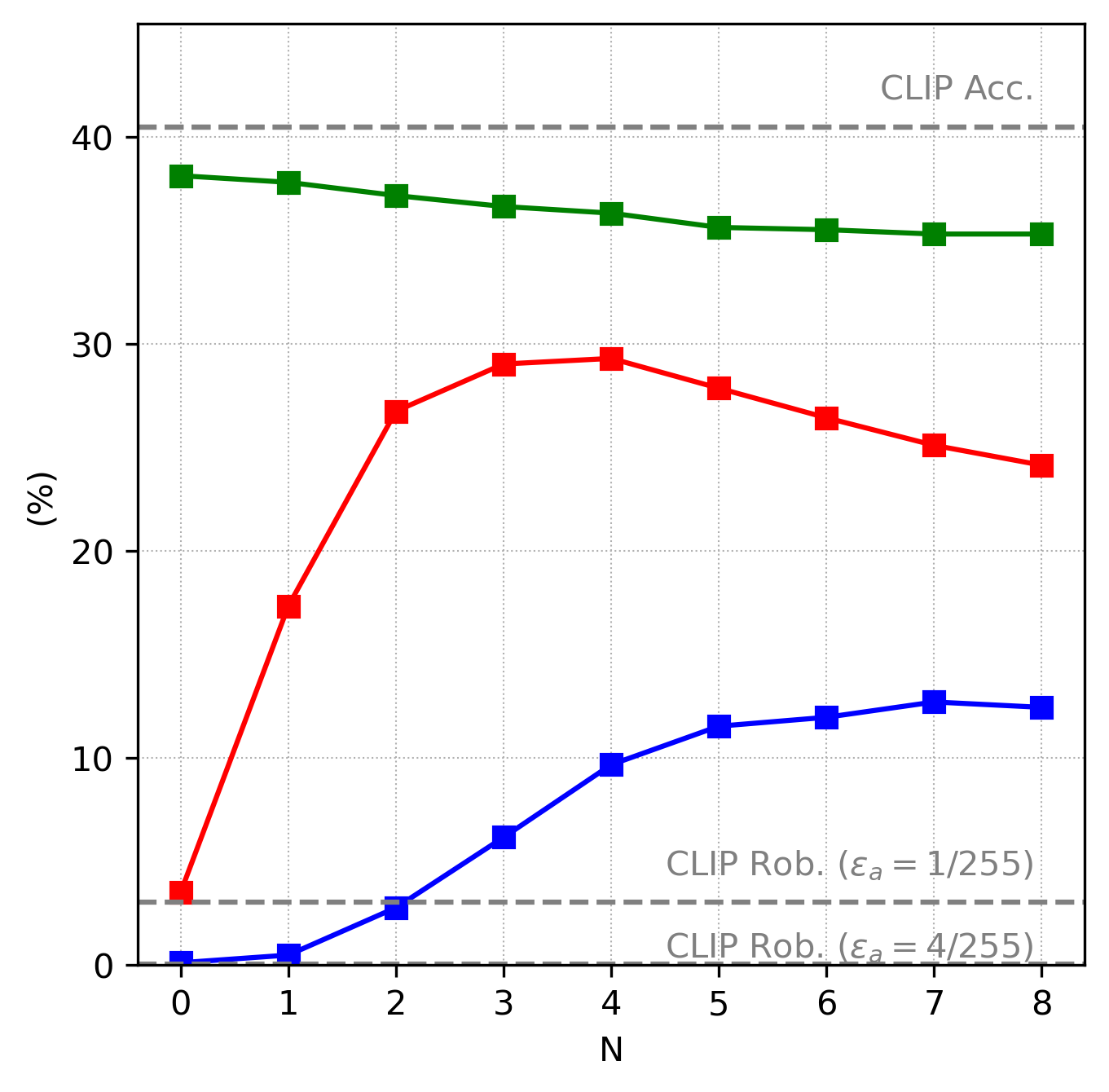}
        \caption{DTD}
        \label{fig:dtd_N}
    \end{subfigure}
    \hfill
    \begin{subfigure}{0.15\textwidth} 
        \centering
        \includegraphics[width=\textwidth]{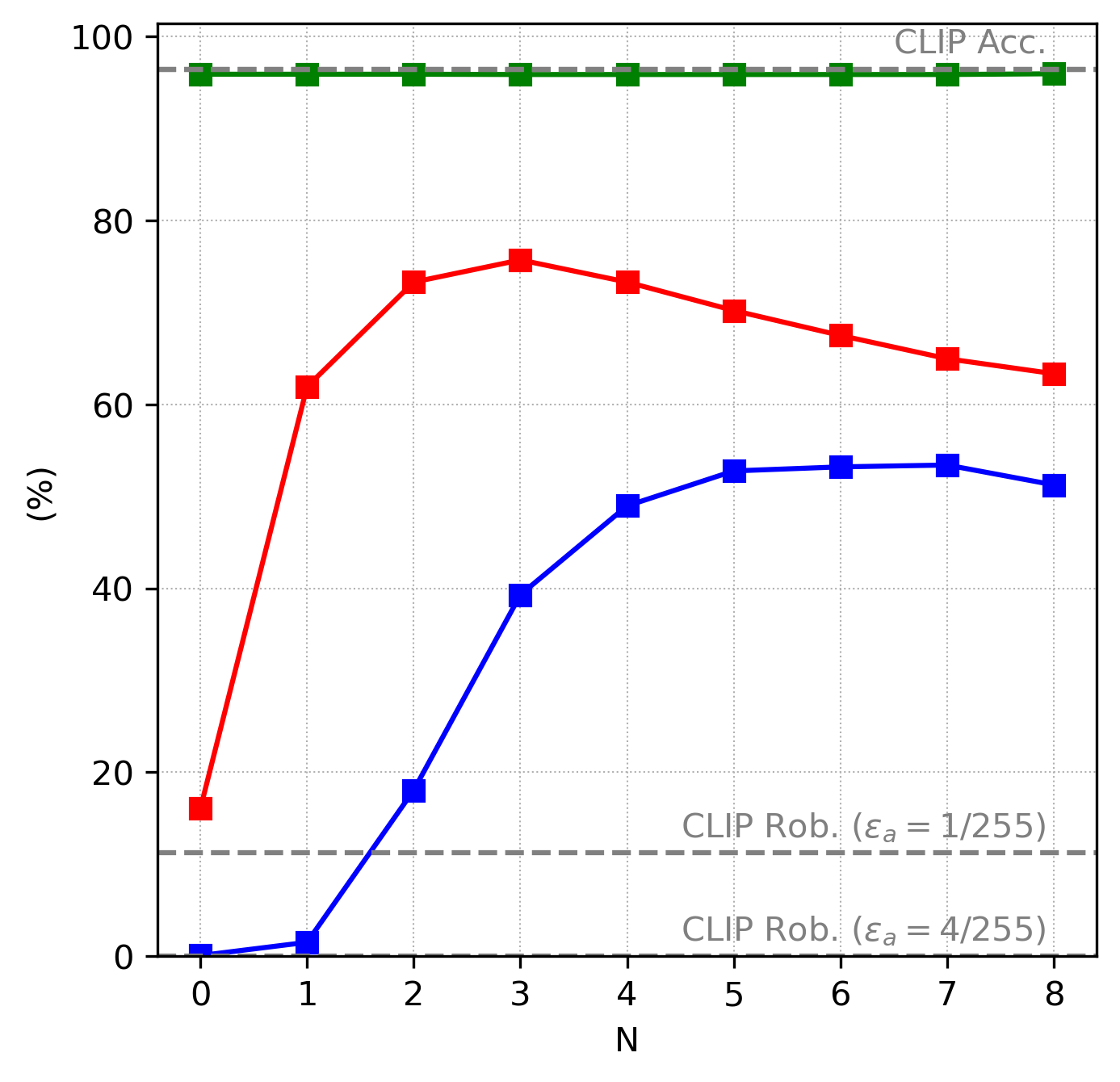}
        \caption{STL10}
        \label{fig:stl10_N}
    \end{subfigure}
    \hfill
    \begin{subfigure}{0.15\textwidth} 
        \centering
        \includegraphics[width=\textwidth]{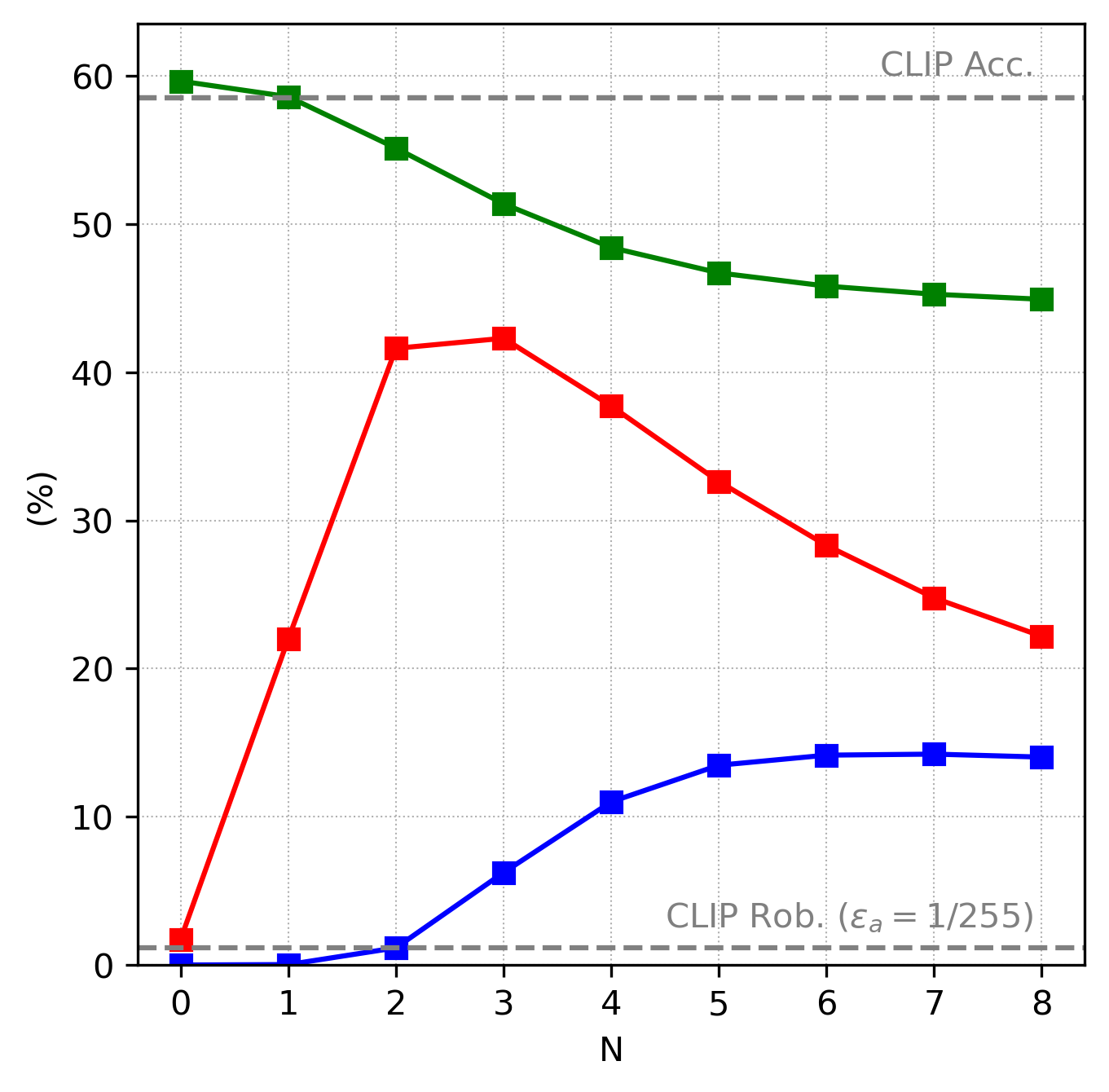}
        \caption{SUN397}
        \label{fig:sun_N}
    \end{subfigure}
    \hfill
    \begin{subfigure}{0.15\textwidth} 
        \centering
        \includegraphics[width=\textwidth]{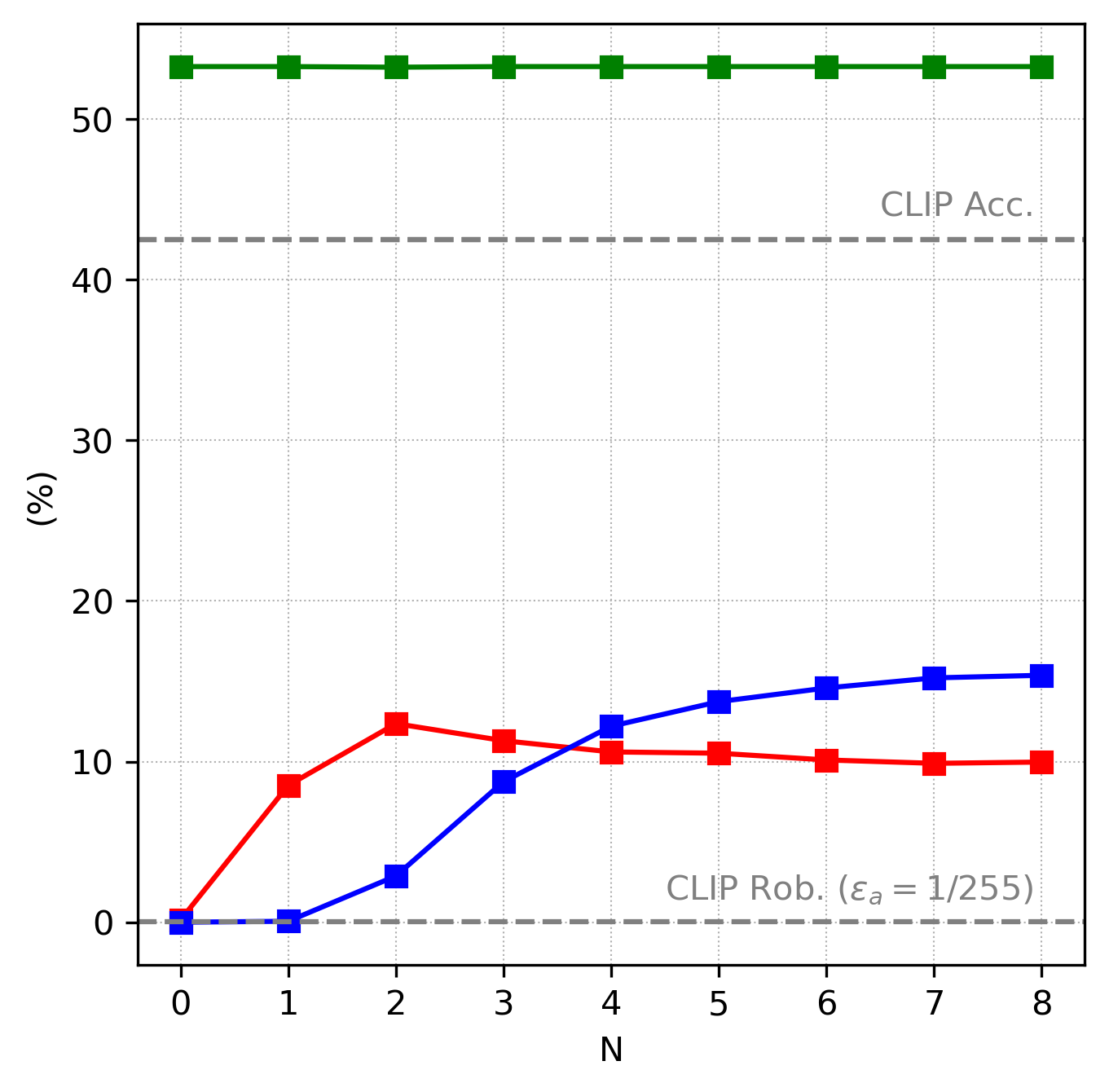}
        \caption{EuroSAT}
        \label{fig:eurosat_N}
    \end{subfigure}
    \hfill
    \begin{subfigure}{0.15\textwidth} 
        \centering
        \includegraphics[width=\textwidth]{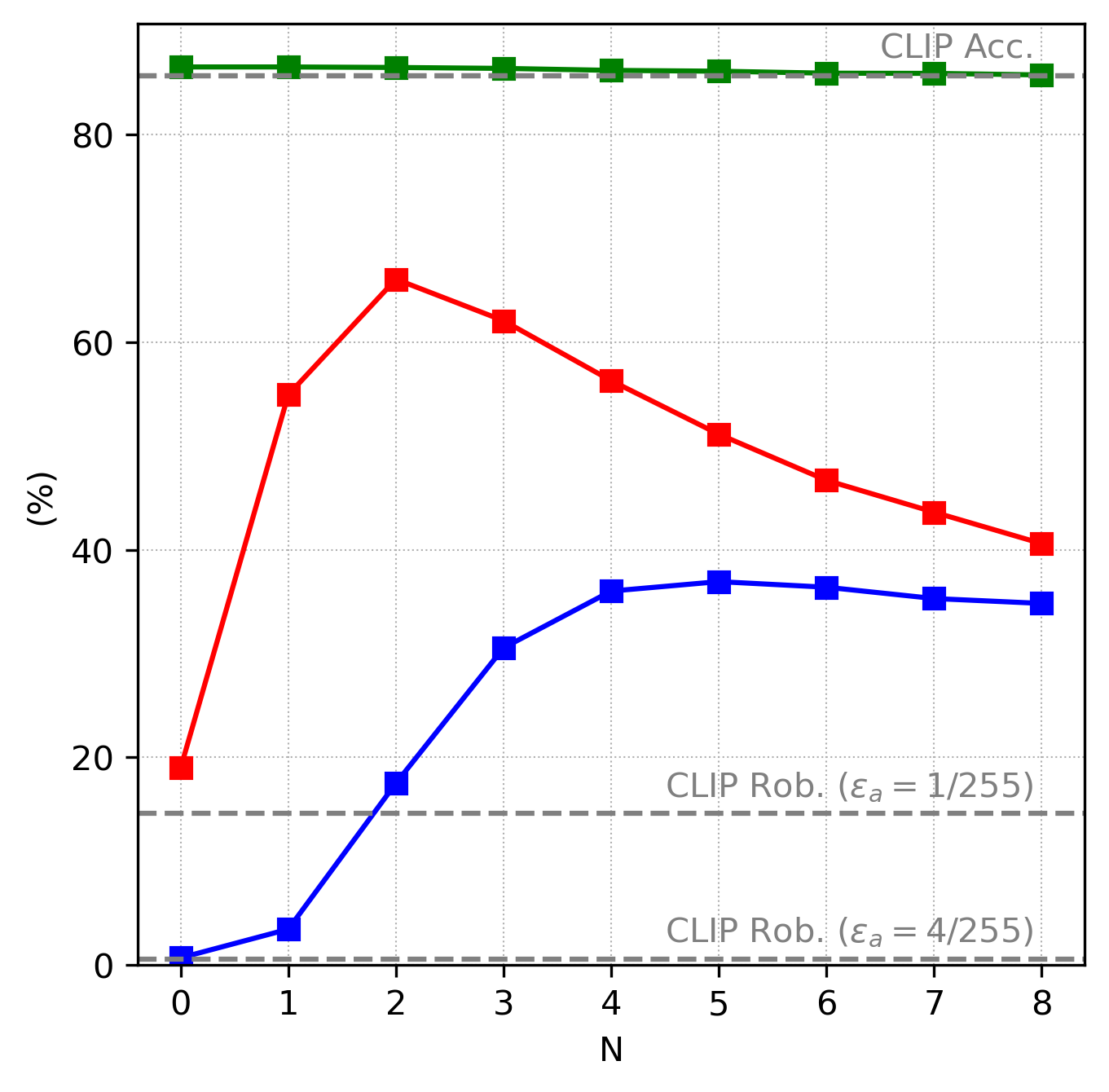}
        \caption{Caltech101}
        \label{fig:caltech101_N}
    \end{subfigure}
    
    \vskip\baselineskip

    \begin{subfigure}{0.15\textwidth} 
        \centering
        \includegraphics[width=\textwidth]{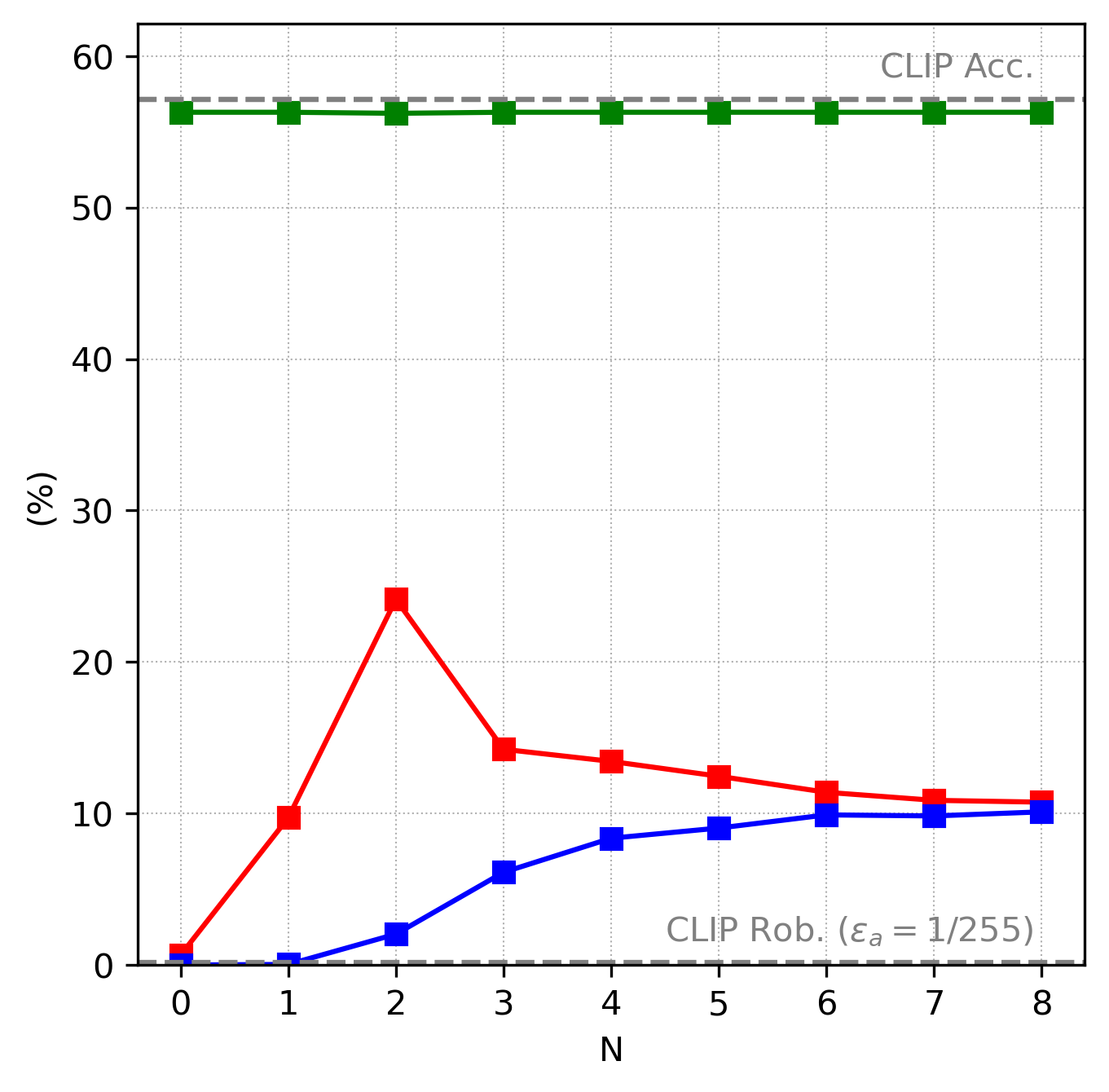}
        \caption{CIFAR100}
        \label{fig:cifar100_N}
    \end{subfigure}
    \hfill
    \begin{subfigure}{0.15\textwidth} 
        \centering
        \includegraphics[width=\textwidth]{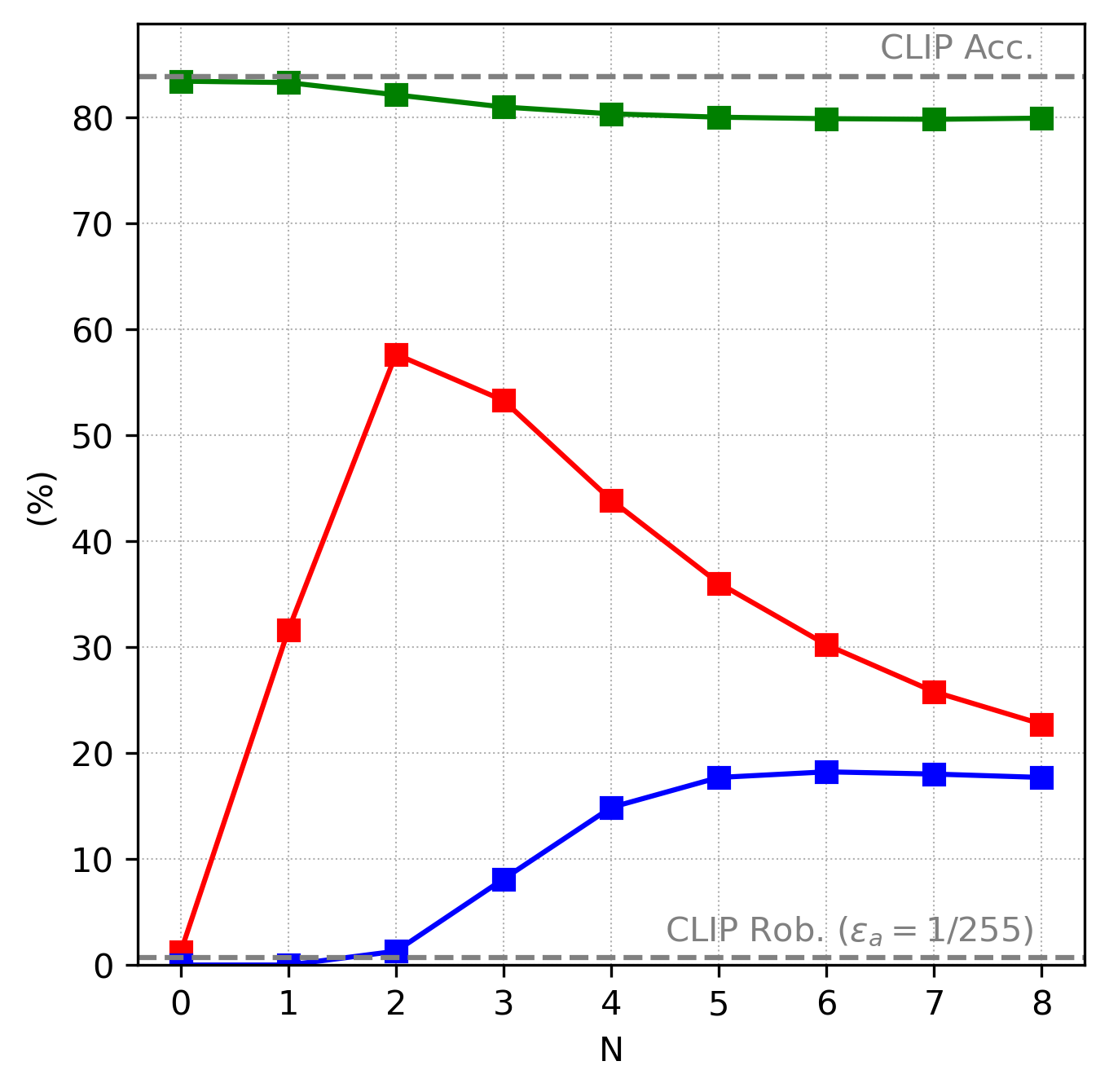}
        \caption{Food101}
        \label{fig:food_N}
    \end{subfigure}
    \hfill
    \begin{subfigure}{0.15\textwidth} 
        \centering
        \includegraphics[width=\textwidth]{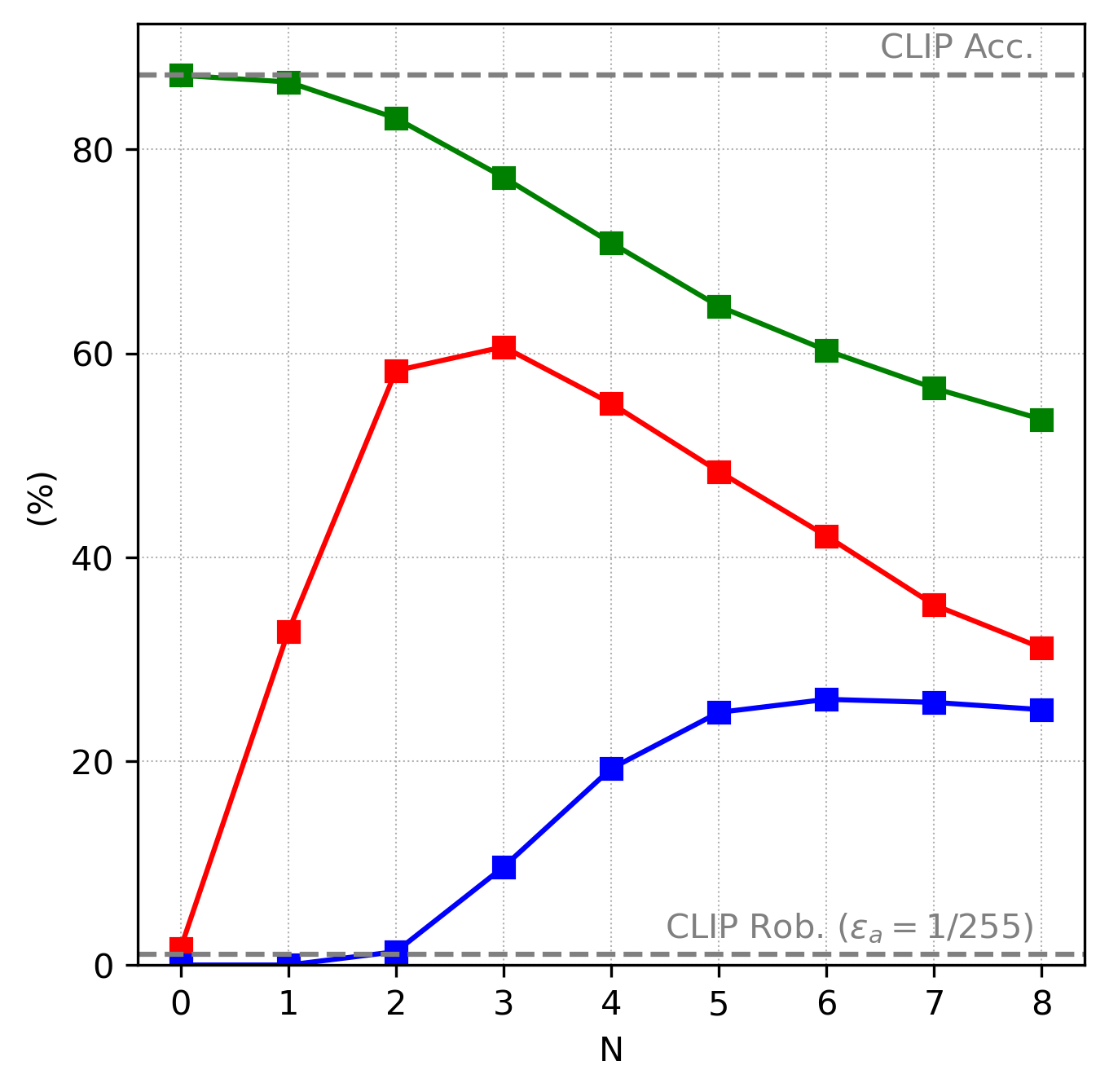}
        \caption{OxfordPets}
        \label{fig:pets_N}
    \end{subfigure}
    \hfill
    \begin{subfigure}{0.15\textwidth} 
        \centering
        \includegraphics[width=\textwidth]{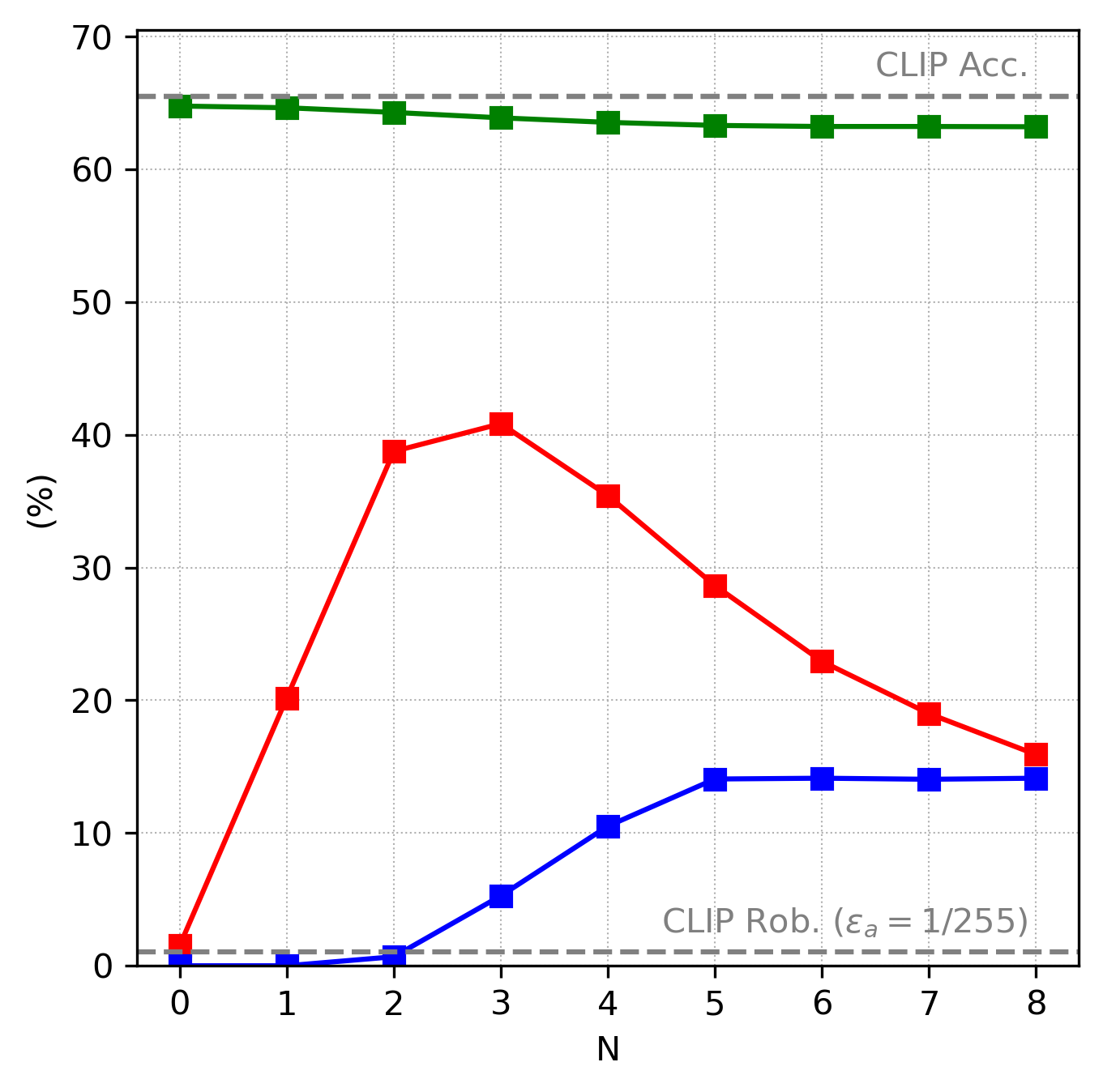}
        \caption{Flower102}
        \label{fig:flower_N}
    \end{subfigure}
    \hfill
    \begin{subfigure}{0.15\textwidth} 
        \centering
        \includegraphics[width=\textwidth]{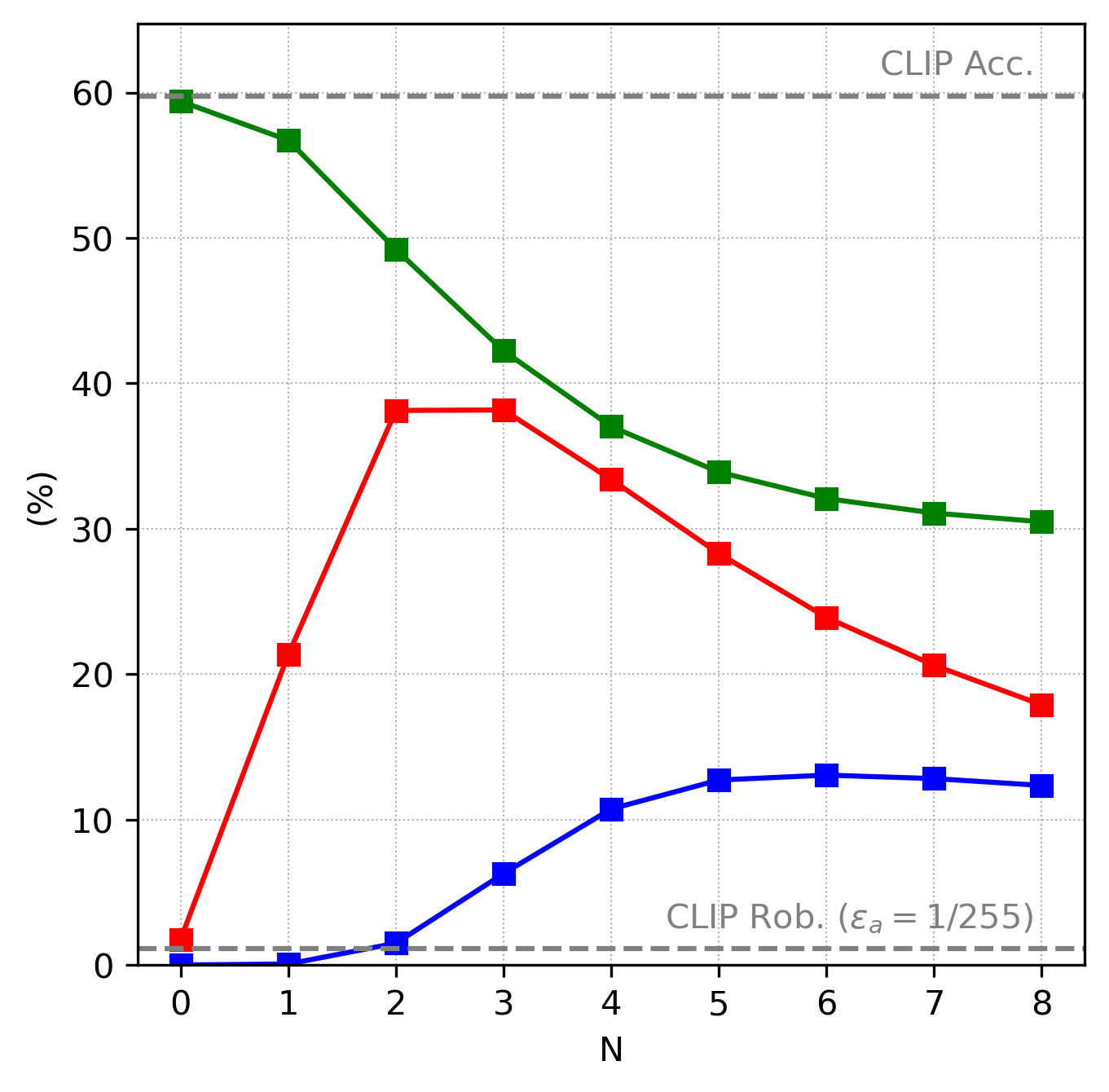}
        \caption{ImageNet}
        \label{fig:imagenet_N}
    \end{subfigure}
    \hfill
    \begin{subfigure}{0.15\textwidth} 
        \centering
        \includegraphics[width=\textwidth]{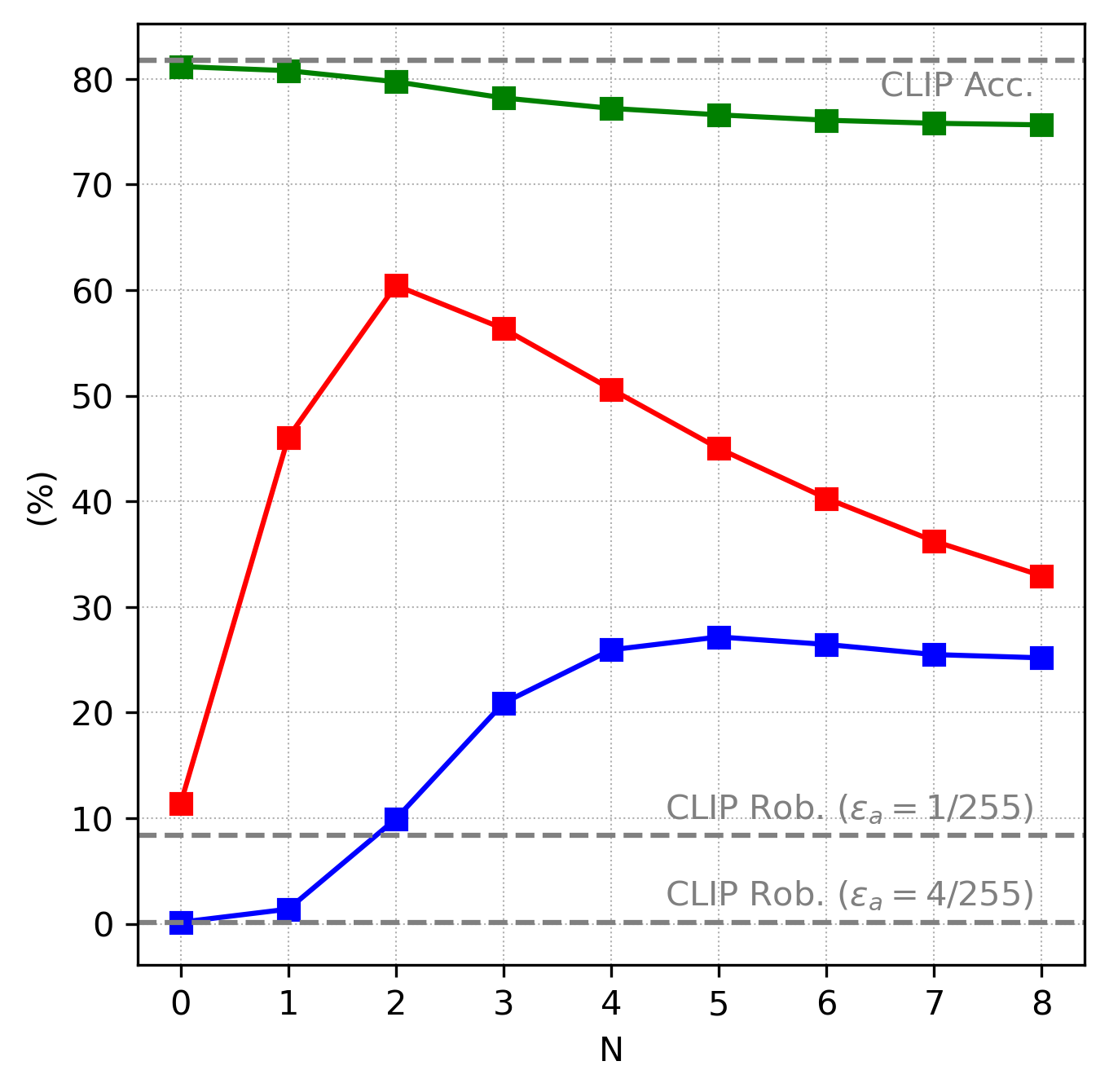}
        \caption{Caltech256}
        \label{fig:caltech256_N}
    \end{subfigure}
    \caption{Effects of the number of steps $N$ for counterattacks performed on CLIP. The \textcolor{ForestGreen}{green} lines represent accuracy on clean images, and \redd{red} and \blue{blue} lines accuracy on adversarial images at $\epsilon_a=1/255$ and $\epsilon_a=4/255$, respectively. 
    }
    \label{fig:N_effect}
    \vspace{-0.5cm}
\end{figure*}

\section{Limitations}
\label{sec:discussion}

Although we show that CLIP possesses the intriguing ability to defend itself from adversary 
that maximises the classification loss,
by performing counterattacks at test time without relying on any auxiliary networks, there are limitations as discussed below. 
\textbf{Firstly}, the robustness gain of applying TTC on TeCoA \cite{mao2023understanding} and PMG-AFT \cite{wang2024pre} is less obvious. 
This is due to the reduced expressiveness of CLIP caused by adversarial finetuning. 
We argue that for large pre-trained models like CLIP, adversarial finetuning should be employed sparingly, considering that a fundamental difference from adversarial studies on non-foundational models is that they have learned massive amounts of real-world knowledge. 
As part of future work, we intend to explore methods to co-ordinate adversarial training and our counterattack paradigm to achieve better robustness and reduce the use of adversarial finetuning.
\textbf{Secondly}, although TTC does not involve training on adversarial images, it incurs more computation expenses at inference time. 
Additionally, the number of counterattack steps affects robustness performance. 
It can be difficult to tune for the most suitable $N$, if the attack strength $\epsilon_a$ is not known \textit{a priori} (\cref{fig:N_effect}). 
We recommend fewer steps (no more than three) if the attack is unknown to avoid excessive counterattacks and unproductive computational overhead. In the future, we intend to explore methods to adjust the number of steps based on the test image.
\textbf{Thirdly}, according to adversarial robustness studies on conventional models, test-time defence can be circumvented by adaptive attacks \cite{croce2022evaluating}. We discuss in Appendix (\cref{sec:adaptive_attack}) possible adaptive attacks to break our counterattacks assuming the worst scenario where the attacker has access to the weights of the deployed CLIP model and TTC performed by the end user.
\section{Conclusion}
\label{sec:conclusion}

We show that CLIP can leverage its own pre-trained vision encoder to defend against adversary maliciously manipulated to maximise its loss by performing counterattacks at test time, 
without relying on any auxiliary networks. 
Based on the finding that adversarial images are `falsely stable', we propose $\tau$-thresholded counterattacks to 
guide the adversarial image away from its original embedding in the latent space.
Experiments on 16 datasets show that TTC employed on CLIP achieves stable and promising accuracy on adversarial images.
TTC is also shown to further enhance robustness of adversarially finetuned CLIP models. 
We also find that finetuning CLIP with adversarial images compromises its own expressiveness, and recommend cautious use of adversarial finetuning as the only approach to robustifying large pre-trained models. 
Our paradigm is the first test-time method to defend CLIP at inference time without any finetuning. We hope this study will encourage future research of robustifying approaches for CLIP alternative to adversarial finetuning.

\noindent
\textbf{Acknowledgement.}
This work was supported by the MUR PNRR project FAIR (PE00000013) funded by the NextGenerationEU and the EU Horizon projects ELIAS (No. 101120237) and AI4Trust (No. 101070190).
{
    \small
    \bibliographystyle{ieeenat_fullname}
    \bibliography{main}

\begin{thebibliography}{63}
\providecommand{\natexlab}[1]{#1}
\providecommand{\url}[1]{\texttt{#1}}
\expandafter\ifx\csname urlstyle\endcsname\relax
  \providecommand{\doi}[1]{doi: #1}\else
  \providecommand{\doi}{doi: \begingroup \urlstyle{rm}\Url}\fi

\bibitem[Alfarra et~al.(2022)Alfarra, P{\'e}rez, Thabet, Bibi, Torr, and Ghanem]{alfarra2022combating}
Motasem Alfarra, Juan~C P{\'e}rez, Ali Thabet, Adel Bibi, Philip~HS Torr, and Bernard Ghanem.
\newblock Combating adversaries with anti-adversaries.
\newblock In \emph{Proceedings of the AAAI Conference on Artificial Intelligence}, pages 5992--6000, 2022.

\bibitem[Athalye et~al.(2018)Athalye, Carlini, and Wagner]{athalye2018obfuscated}
Anish Athalye, Nicholas Carlini, and David Wagner.
\newblock Obfuscated gradients give a false sense of security: Circumventing defenses to adversarial examples.
\newblock In \emph{International conference on machine learning}, pages 274--283. PMLR, 2018.

\bibitem[Bai et~al.(2021)Bai, Luo, Zhao, Wen, and Wang]{ijcai2021p0591}
Tao Bai, Jinqi Luo, Jun Zhao, Bihan Wen, and Qian Wang.
\newblock Recent advances in adversarial training for adversarial robustness.
\newblock In \emph{Proceedings of the Thirtieth International Joint Conference on Artificial Intelligence}, pages 4312--4321. International Joint Conferences on Artificial Intelligence Organization, 2021.
\newblock Survey Track.

\bibitem[Bejnordi et~al.(2017)Bejnordi, Veta, Van~Diest, Van~Ginneken, Karssemeijer, Litjens, Van Der~Laak, Hermsen, Manson, Balkenhol, et~al.]{bejnordi2017diagnostic}
Babak~Ehteshami Bejnordi, Mitko Veta, Paul~Johannes Van~Diest, Bram Van~Ginneken, Nico Karssemeijer, Geert Litjens, Jeroen~AWM Van Der~Laak, Meyke Hermsen, Quirine~F Manson, Maschenka Balkenhol, et~al.
\newblock Diagnostic assessment of deep learning algorithms for detection of lymph node metastases in women with breast cancer.
\newblock \emph{Jama}, 318\penalty0 (22):\penalty0 2199--2210, 2017.

\bibitem[Bossard et~al.(2014)Bossard, Guillaumin, and Van~Gool]{bossard2014food}
Lukas Bossard, Matthieu Guillaumin, and Luc Van~Gool.
\newblock Food-101--mining discriminative components with random forests.
\newblock In \emph{Computer vision--ECCV 2014: 13th European conference, zurich, Switzerland, September 6-12, 2014, proceedings, part VI 13}, pages 446--461. Springer, 2014.

\bibitem[Carlini and Wagner(2017)]{carlini2017towards}
Nicholas Carlini and David Wagner.
\newblock Towards evaluating the robustness of neural networks.
\newblock In \emph{2017 ieee symposium on security and privacy (sp)}, pages 39--57. Ieee, 2017.

\bibitem[Caron et~al.(2021)Caron, Touvron, Misra, J{\'e}gou, Mairal, Bojanowski, and Joulin]{caron2021emerging}
Mathilde Caron, Hugo Touvron, Ishan Misra, Herv{\'e} J{\'e}gou, Julien Mairal, Piotr Bojanowski, and Armand Joulin.
\newblock Emerging properties in self-supervised vision transformers.
\newblock In \emph{Proceedings of the IEEE/CVF international conference on computer vision}, pages 9650--9660, 2021.

\bibitem[Chen et~al.(2020)Chen, Kornblith, Norouzi, and Hinton]{chen2020simple}
Ting Chen, Simon Kornblith, Mohammad Norouzi, and Geoffrey Hinton.
\newblock A simple framework for contrastive learning of visual representations.
\newblock In \emph{International conference on machine learning}, pages 1597--1607. PMLR, 2020.

\bibitem[Cimpoi et~al.(2014)Cimpoi, Maji, Kokkinos, Mohamed, and Vedaldi]{cimpoi2014describing}
Mircea Cimpoi, Subhransu Maji, Iasonas Kokkinos, Sammy Mohamed, and Andrea Vedaldi.
\newblock Describing textures in the wild.
\newblock In \emph{Proceedings of the IEEE conference on computer vision and pattern recognition}, pages 3606--3613, 2014.

\bibitem[Coates et~al.(2011)Coates, Ng, and Lee]{coates2011analysis}
Adam Coates, Andrew Ng, and Honglak Lee.
\newblock An analysis of single-layer networks in unsupervised feature learning.
\newblock In \emph{Proceedings of the fourteenth international conference on artificial intelligence and statistics}, pages 215--223. JMLR Workshop and Conference Proceedings, 2011.

\bibitem[Croce and Hein(2020)]{croce2020reliable}
Francesco Croce and Matthias Hein.
\newblock Reliable evaluation of adversarial robustness with an ensemble of diverse parameter-free attacks.
\newblock In \emph{International conference on machine learning}, pages 2206--2216. PMLR, 2020.

\bibitem[Croce et~al.(2022)Croce, Gowal, Brunner, Shelhamer, Hein, and Cemgil]{croce2022evaluating}
Francesco Croce, Sven Gowal, Thomas Brunner, Evan Shelhamer, Matthias Hein, and Taylan Cemgil.
\newblock Evaluating the adversarial robustness of adaptive test-time defenses.
\newblock In \emph{International Conference on Machine Learning}, pages 4421--4435. PMLR, 2022.

\bibitem[Deng et~al.(2009)Deng, Dong, Socher, Li, Li, and Fei-Fei]{deng2009imagenet}
Jia Deng, Wei Dong, Richard Socher, Li-Jia Li, Kai Li, and Li Fei-Fei.
\newblock Imagenet: A large-scale hierarchical image database.
\newblock In \emph{2009 IEEE conference on computer vision and pattern recognition}, pages 248--255. Ieee, 2009.

\bibitem[Fei-Fei et~al.(2006)Fei-Fei, Fergus, and Perona]{fei2006one}
Li Fei-Fei, Robert Fergus, and Pietro Perona.
\newblock One-shot learning of object categories.
\newblock \emph{IEEE transactions on pattern analysis and machine intelligence}, 28\penalty0 (4):\penalty0 594--611, 2006.

\bibitem[Griffin et~al.(2007)Griffin, Holub, Perona, et~al.]{griffin2007caltech}
Gregory Griffin, Alex Holub, Pietro Perona, et~al.
\newblock Caltech-256 object category dataset.
\newblock Technical report, Technical Report 7694, California Institute of Technology Pasadena, 2007.

\bibitem[Grill et~al.(2020)Grill, Strub, Altch{\'e}, Tallec, Richemond, Buchatskaya, Doersch, Avila~Pires, Guo, Gheshlaghi~Azar, et~al.]{grill2020bootstrap}
Jean-Bastien Grill, Florian Strub, Florent Altch{\'e}, Corentin Tallec, Pierre Richemond, Elena Buchatskaya, Carl Doersch, Bernardo Avila~Pires, Zhaohan Guo, Mohammad Gheshlaghi~Azar, et~al.
\newblock Bootstrap your own latent-a new approach to self-supervised learning.
\newblock \emph{Advances in neural information processing systems}, 33:\penalty0 21271--21284, 2020.

\bibitem[Guo et~al.(2018)Guo, Rana, Cisse, and van~der Maaten]{guo2018countering}
Chuan Guo, Mayank Rana, Moustapha Cisse, and Laurens van~der Maaten.
\newblock Countering adversarial images using input transformations.
\newblock In \emph{International Conference on Learning Representations}, 2018.

\bibitem[He et~al.(2016)He, Zhang, Ren, and Sun]{he2016deep}
Kaiming He, Xiangyu Zhang, Shaoqing Ren, and Jian Sun.
\newblock Deep residual learning for image recognition.
\newblock In \emph{Proceedings of the IEEE conference on computer vision and pattern recognition}, pages 770--778, 2016.

\bibitem[Helber et~al.(2019)Helber, Bischke, Dengel, and Borth]{helber2019eurosat}
Patrick Helber, Benjamin Bischke, Andreas Dengel, and Damian Borth.
\newblock Eurosat: A novel dataset and deep learning benchmark for land use and land cover classification.
\newblock \emph{IEEE Journal of Selected Topics in Applied Earth Observations and Remote Sensing}, 12\penalty0 (7):\penalty0 2217--2226, 2019.

\bibitem[Hwang et~al.(2023)Hwang, Lee, and Rhee]{hwang2023aid}
Duhun Hwang, Eunjung Lee, and Wonjong Rhee.
\newblock Aid-purifier: A light auxiliary network for boosting adversarial defense.
\newblock \emph{Neurocomputing}, 541:\penalty0 126251, 2023.

\bibitem[Jia et~al.(2021)Jia, Yang, Xia, Chen, Parekh, Pham, Le, Sung, Li, and Duerig]{jia2021scaling}
Chao Jia, Yinfei Yang, Ye Xia, Yi-Ting Chen, Zarana Parekh, Hieu Pham, Quoc Le, Yun-Hsuan Sung, Zhen Li, and Tom Duerig.
\newblock Scaling up visual and vision-language representation learning with noisy text supervision.
\newblock In \emph{International conference on machine learning}, pages 4904--4916. PMLR, 2021.

\bibitem[Krause et~al.(2013)Krause, Stark, Deng, and Fei-Fei]{krause20133d}
Jonathan Krause, Michael Stark, Jia Deng, and Li Fei-Fei.
\newblock 3d object representations for fine-grained categorization.
\newblock In \emph{Proceedings of the IEEE international conference on computer vision workshops}, pages 554--561, 2013.

\bibitem[Krizhevsky et~al.(2009)Krizhevsky, Hinton, et~al.]{krizhevsky2009learning}
Alex Krizhevsky, Geoffrey Hinton, et~al.
\newblock Learning multiple layers of features from tiny images.
\newblock 2009.

\bibitem[Krizhevsky et~al.(2012)Krizhevsky, Sutskever, and Hinton]{krizhevsky2012imagenet}
Alex Krizhevsky, Ilya Sutskever, and Geoffrey~E Hinton.
\newblock Imagenet classification with deep convolutional neural networks.
\newblock \emph{Advances in neural information processing systems}, 25, 2012.

\bibitem[Kurakin et~al.(2018)Kurakin, Goodfellow, and Bengio]{kurakin2018adversarial}
Alexey Kurakin, Ian~J Goodfellow, and Samy Bengio.
\newblock Adversarial examples in the physical world.
\newblock In \emph{Artificial intelligence safety and security}, pages 99--112. Chapman and Hall/CRC, 2018.

\bibitem[Li et~al.(2024{\natexlab{a}})Li, Guan, Qiu, and Spratling]{li2024one}
Lin Li, Haoyan Guan, Jianing Qiu, and Michael Spratling.
\newblock One prompt word is enough to boost adversarial robustness for pre-trained vision-language models.
\newblock In \emph{Proceedings of the IEEE/CVF Conference on Computer Vision and Pattern Recognition}, pages 24408--24419, 2024{\natexlab{a}}.

\bibitem[Li et~al.(2024{\natexlab{b}})Li, Zhang, Liu, Hu, Zhang, and Hu]{li2024language}
Xiao Li, Wei Zhang, Yining Liu, Zhanhao Hu, Bo Zhang, and Xiaolin Hu.
\newblock Language-driven anchors for zero-shot adversarial robustness.
\newblock In \emph{Proceedings of the IEEE/CVF Conference on Computer Vision and Pattern Recognition}, pages 24686--24695, 2024{\natexlab{b}}.

\bibitem[Liu et~al.(2023)Liu, Li, Wu, and Lee]{NEURIPS2023_6dcf277e}
Haotian Liu, Chunyuan Li, Qingyang Wu, and Yong~Jae Lee.
\newblock Visual instruction tuning.
\newblock In \emph{Advances in Neural Information Processing Systems}, pages 34892--34916. Curran Associates, Inc., 2023.

\bibitem[Madry et~al.(2018)Madry, Makelov, Schmidt, Tsipras, and Vladu]{madry2018towards}
Aleksander Madry, Aleksandar Makelov, Ludwig Schmidt, Dimitris Tsipras, and Adrian Vladu.
\newblock Towards deep learning models resistant to adversarial attacks.
\newblock In \emph{International Conference on Learning Representations}, 2018.

\bibitem[Maji et~al.(2013)Maji, Rahtu, Kannala, Blaschko, and Vedaldi]{maji2013fine}
Subhransu Maji, Esa Rahtu, Juho Kannala, Matthew Blaschko, and Andrea Vedaldi.
\newblock Fine-grained visual classification of aircraft.
\newblock \emph{arXiv preprint arXiv:1306.5151}, 2013.

\bibitem[Mao et~al.(2021)Mao, Chiquier, Wang, Yang, and Vondrick]{mao2021adversarial}
Chengzhi Mao, Mia Chiquier, Hao Wang, Junfeng Yang, and Carl Vondrick.
\newblock Adversarial attacks are reversible with natural supervision.
\newblock In \emph{Proceedings of the IEEE/CVF International Conference on Computer Vision}, pages 661--671, 2021.

\bibitem[Mao et~al.(2023)Mao, Geng, Yang, Wang, and Vondrick]{mao2023understanding}
Chengzhi Mao, Scott Geng, Junfeng Yang, Xin Wang, and Carl Vondrick.
\newblock Understanding zero-shot adversarial robustness for large-scale models.
\newblock In \emph{The Eleventh International Conference on Learning Representations}, 2023.

\bibitem[Moosavi-Dezfooli et~al.(2016)Moosavi-Dezfooli, Fawzi, and Frossard]{moosavi2016deepfool}
Seyed-Mohsen Moosavi-Dezfooli, Alhussein Fawzi, and Pascal Frossard.
\newblock Deepfool: a simple and accurate method to fool deep neural networks.
\newblock In \emph{Proceedings of the IEEE conference on computer vision and pattern recognition}, pages 2574--2582, 2016.

\bibitem[Nie et~al.(2022)Nie, Guo, Huang, Xiao, Vahdat, and Anandkumar]{nie2022diffusion}
Weili Nie, Brandon Guo, Yujia Huang, Chaowei Xiao, Arash Vahdat, and Animashree Anandkumar.
\newblock Diffusion models for adversarial purification.
\newblock In \emph{International Conference on Machine Learning}, pages 16805--16827. PMLR, 2022.

\bibitem[Nilsback and Zisserman(2008)]{nilsback2008automated}
Maria-Elena Nilsback and Andrew Zisserman.
\newblock Automated flower classification over a large number of classes.
\newblock In \emph{2008 Sixth Indian conference on computer vision, graphics \& image processing}, pages 722--729. IEEE, 2008.

\bibitem[Papernot et~al.(2016)Papernot, McDaniel, Jha, Fredrikson, Celik, and Swami]{papernot2016limitations}
Nicolas Papernot, Patrick McDaniel, Somesh Jha, Matt Fredrikson, Z~Berkay Celik, and Ananthram Swami.
\newblock The limitations of deep learning in adversarial settings.
\newblock In \emph{2016 IEEE European symposium on security and privacy (EuroS\&P)}, pages 372--387. IEEE, 2016.

\bibitem[Parkhi et~al.(2012)Parkhi, Vedaldi, Zisserman, and Jawahar]{parkhi2012cats}
Omkar~M Parkhi, Andrea Vedaldi, Andrew Zisserman, and CV Jawahar.
\newblock Cats and dogs.
\newblock In \emph{2012 IEEE conference on computer vision and pattern recognition}, pages 3498--3505. IEEE, 2012.

\bibitem[P{\'e}rez et~al.(2021)P{\'e}rez, Alfarra, Jeanneret, Rueda, Thabet, Ghanem, and Arbel{\'a}ez]{perez2021enhancing}
Juan~C P{\'e}rez, Motasem Alfarra, Guillaume Jeanneret, Laura Rueda, Ali Thabet, Bernard Ghanem, and Pablo Arbel{\'a}ez.
\newblock Enhancing adversarial robustness via test-time transformation ensembling.
\newblock In \emph{Proceedings of the IEEE/CVF International Conference on Computer Vision}, pages 81--91, 2021.

\bibitem[Radford et~al.(2021)Radford, Kim, Hallacy, Ramesh, Goh, Agarwal, Sastry, Askell, Mishkin, Clark, et~al.]{radford2021learning}
Alec Radford, Jong~Wook Kim, Chris Hallacy, Aditya Ramesh, Gabriel Goh, Sandhini Agarwal, Girish Sastry, Amanda Askell, Pamela Mishkin, Jack Clark, et~al.
\newblock Learning transferable visual models from natural language supervision.
\newblock In \emph{International conference on machine learning}, pages 8748--8763. PMLR, 2021.

\bibitem[Ramesh et~al.(2021)Ramesh, Pavlov, Goh, Gray, Voss, Radford, Chen, and Sutskever]{ramesh2021zero}
Aditya Ramesh, Mikhail Pavlov, Gabriel Goh, Scott Gray, Chelsea Voss, Alec Radford, Mark Chen, and Ilya Sutskever.
\newblock Zero-shot text-to-image generation.
\newblock In \emph{International conference on machine learning}, pages 8821--8831. Pmlr, 2021.

\bibitem[Rice et~al.(2020)Rice, Wong, and Kolter]{rice2020overfitting}
Leslie Rice, Eric Wong, and Zico Kolter.
\newblock Overfitting in adversarially robust deep learning.
\newblock In \emph{International conference on machine learning}, pages 8093--8104. PMLR, 2020.

\bibitem[Saharia et~al.(2022)Saharia, Chan, Saxena, Li, Whang, Denton, Ghasemipour, Gontijo~Lopes, Karagol~Ayan, Salimans, et~al.]{saharia2022photorealistic}
Chitwan Saharia, William Chan, Saurabh Saxena, Lala Li, Jay Whang, Emily~L Denton, Kamyar Ghasemipour, Raphael Gontijo~Lopes, Burcu Karagol~Ayan, Tim Salimans, et~al.
\newblock Photorealistic text-to-image diffusion models with deep language understanding.
\newblock \emph{Advances in neural information processing systems}, 35:\penalty0 36479--36494, 2022.

\bibitem[Samangouei et~al.(2018)Samangouei, Kabkab, and Chellappa]{samangouei2018defense}
Pouya Samangouei, Maya Kabkab, and Rama Chellappa.
\newblock Defense-gan: Protecting classifiers against adversarial attacks using generative models.
\newblock In \emph{International Conference on Learning Representations}, 2018.

\bibitem[Schlarmann et~al.(2024)Schlarmann, Singh, Croce, and Hein]{pmlr-v235-schlarmann24a}
Christian Schlarmann, Naman~Deep Singh, Francesco Croce, and Matthias Hein.
\newblock Robust {CLIP}: Unsupervised adversarial fine-tuning of vision embeddings for robust large vision-language models.
\newblock In \emph{Proceedings of the 41st International Conference on Machine Learning}, pages 43685--43704. PMLR, 2024.

\bibitem[Shafahi et~al.(2019)Shafahi, Najibi, Ghiasi, Xu, Dickerson, Studer, Davis, Taylor, and Goldstein]{shafahi2019adversarial}
Ali Shafahi, Mahyar Najibi, Mohammad~Amin Ghiasi, Zheng Xu, John Dickerson, Christoph Studer, Larry~S Davis, Gavin Taylor, and Tom Goldstein.
\newblock Adversarial training for free!
\newblock \emph{Advances in neural information processing systems}, 32, 2019.

\bibitem[Shayegani et~al.(2023)Shayegani, Dong, and Abu-Ghazaleh]{shayegani2023jailbreak}
Erfan Shayegani, Yue Dong, and Nael Abu-Ghazaleh.
\newblock Jailbreak in pieces: Compositional adversarial attacks on multi-modal language models.
\newblock In \emph{The Twelfth International Conference on Learning Representations}, 2023.

\bibitem[Su et~al.(2018)Su, Zhang, Chen, Yi, Chen, and Gao]{su2018robustness}
Dong Su, Huan Zhang, Hongge Chen, Jinfeng Yi, Pin-Yu Chen, and Yupeng Gao.
\newblock Is robustness the cost of accuracy?--a comprehensive study on the robustness of 18 deep image classification models.
\newblock In \emph{Proceedings of the European conference on computer vision (ECCV)}, pages 631--648, 2018.

\bibitem[Szegedy(2013)]{szegedy2013intriguing}
C Szegedy.
\newblock Intriguing properties of neural networks.
\newblock \emph{arXiv preprint arXiv:1312.6199}, 2013.

\bibitem[Szegedy et~al.(2015)Szegedy, Liu, Jia, Sermanet, Reed, Anguelov, Erhan, Vanhoucke, and Rabinovich]{szegedy2015going}
Christian Szegedy, Wei Liu, Yangqing Jia, Pierre Sermanet, Scott Reed, Dragomir Anguelov, Dumitru Erhan, Vincent Vanhoucke, and Andrew Rabinovich.
\newblock Going deeper with convolutions.
\newblock In \emph{Proceedings of the IEEE conference on computer vision and pattern recognition}, pages 1--9, 2015.

\bibitem[Wang et~al.(2024)Wang, Zhang, Yuan, and Shan]{wang2024pre}
Sibo Wang, Jie Zhang, Zheng Yuan, and Shiguang Shan.
\newblock Pre-trained model guided fine-tuning for zero-shot adversarial robustness.
\newblock In \emph{Proceedings of the IEEE/CVF Conference on Computer Vision and Pattern Recognition}, pages 24502--24511, 2024.

\bibitem[Wong et~al.(2020)Wong, Rice, and Kolter]{wongfast}
Eric Wong, Leslie Rice, and J~Zico Kolter.
\newblock Fast is better than free: Revisiting adversarial training.
\newblock In \emph{International Conference on Learning Representations}, 2020.

\bibitem[Wu et~al.(2021)Wu, Pan, Shen, Gu, Zhao, Li, Cai, He, and Liu]{wu2021attacking}
Boxi Wu, Heng Pan, Li Shen, Jindong Gu, Shuai Zhao, Zhifeng Li, Deng Cai, Xiaofei He, and Wei Liu.
\newblock Attacking adversarial attacks as a defense.
\newblock \emph{arXiv preprint arXiv:2106.04938}, 2021.

\bibitem[Xiao et~al.(2010)Xiao, Hays, Ehinger, Oliva, and Torralba]{xiao2010sun}
Jianxiong Xiao, James Hays, Krista~A Ehinger, Aude Oliva, and Antonio Torralba.
\newblock Sun database: Large-scale scene recognition from abbey to zoo.
\newblock In \emph{2010 IEEE computer society conference on computer vision and pattern recognition}, pages 3485--3492. IEEE, 2010.

\bibitem[Xie et~al.(2018)Xie, Wang, Zhang, Ren, and Yuille]{xie2018mitigating}
Cihang Xie, Jianyu Wang, Zhishuai Zhang, Zhou Ren, and Alan Yuille.
\newblock Mitigating adversarial effects through randomization.
\newblock In \emph{International Conference on Learning Representations}, 2018.

\bibitem[Yoon et~al.(2021)Yoon, Hwang, and Lee]{yoon2021adversarial}
Jongmin Yoon, Sung~Ju Hwang, and Juho Lee.
\newblock Adversarial purification with score-based generative models.
\newblock In \emph{International Conference on Machine Learning}, pages 12062--12072. PMLR, 2021.

\bibitem[Yu et~al.(2022)Yu, Wang, Vasudevan, Yeung, Seyedhosseini, and Wu]{yu2022coca}
Jiahui Yu, Zirui Wang, Vijay Vasudevan, Legg Yeung, Mojtaba Seyedhosseini, and Yonghui Wu.
\newblock Coca: Contrastive captioners are image-text foundation models.
\newblock \emph{Transactions on Machine Learning Research}, 2022.

\bibitem[Yucel et~al.(2020)Yucel, Cinbis, and Duygulu]{yucel2020deep}
Mehmet~Kerim Yucel, Ramazan~Gokberk Cinbis, and Pinar Duygulu.
\newblock A deep dive into adversarial robustness in zero-shot learning.
\newblock In \emph{Computer Vision--ECCV 2020 Workshops: Glasgow, UK, August 23--28, 2020, Proceedings, Part I 16}, pages 3--21. Springer, 2020.

\bibitem[Zhang et~al.(2019)Zhang, Yu, Jiao, Xing, El~Ghaoui, and Jordan]{zhang2019theoretically}
Hongyang Zhang, Yaodong Yu, Jiantao Jiao, Eric Xing, Laurent El~Ghaoui, and Michael Jordan.
\newblock Theoretically principled trade-off between robustness and accuracy.
\newblock In \emph{International conference on machine learning}, pages 7472--7482. PMLR, 2019.

\bibitem[Zhang et~al.(2024)Zhang, Ma, Wang, Qiu, Wang, Jiang, and Sang]{zhang2023adversarial}
Jiaming Zhang, Xingjun Ma, Xin Wang, Lingyu Qiu, Jiaqi Wang, Yu-Gang Jiang, and Jitao Sang.
\newblock Adversarial prompt tuning for vision-language models.
\newblock In \emph{Proceedings of the European conference on computer vision (ECCV)}, 2024.

\bibitem[Zhao et~al.(2023)Zhao, Pang, Du, Yang, Li, Cheung, and Lin]{zhao2023evaluating}
Yunqing Zhao, Tianyu Pang, Chao Du, Xiao Yang, Chongxuan Li, Ngai-Man Cheung, and Min Lin.
\newblock On evaluating adversarial robustness of large vision-language models.
\newblock In \emph{Proceedings of the 37th International Conference on Neural Information Processing Systems}, pages 54111--54138, 2023.

\bibitem[Zhou et~al.(2022{\natexlab{a}})Zhou, Yang, Loy, and Liu]{zhou2022conditional}
Kaiyang Zhou, Jingkang Yang, Chen~Change Loy, and Ziwei Liu.
\newblock Conditional prompt learning for vision-language models.
\newblock In \emph{Proceedings of the IEEE/CVF conference on computer vision and pattern recognition}, pages 16816--16825, 2022{\natexlab{a}}.

\bibitem[Zhou et~al.(2022{\natexlab{b}})Zhou, Yang, Loy, and Liu]{zhou2022learning}
Kaiyang Zhou, Jingkang Yang, Chen~Change Loy, and Ziwei Liu.
\newblock Learning to prompt for vision-language models.
\newblock \emph{International Journal of Computer Vision}, 130\penalty0 (9):\penalty0 2337--2348, 2022{\natexlab{b}}.

\bibitem[Zhou et~al.(2024)Zhou, Xia, Lin, Han, and Liu]{zhou2024few}
Yiwei Zhou, Xiaobo Xia, Zhiwei Lin, Bo Han, and Tongliang Liu.
\newblock Few-shot adversarial prompt learning on vision-language models.
\newblock \emph{arXiv preprint arXiv:2403.14774}, 2024.

\end{thebibliography}
}

\clearpage
\setcounter{page}{1}
\maketitlesupplementary

\begin{figure*}[t]
    \centering
    \begin{subfigure}{0.16\textwidth}  
        \centering
        \includegraphics[width=\textwidth]{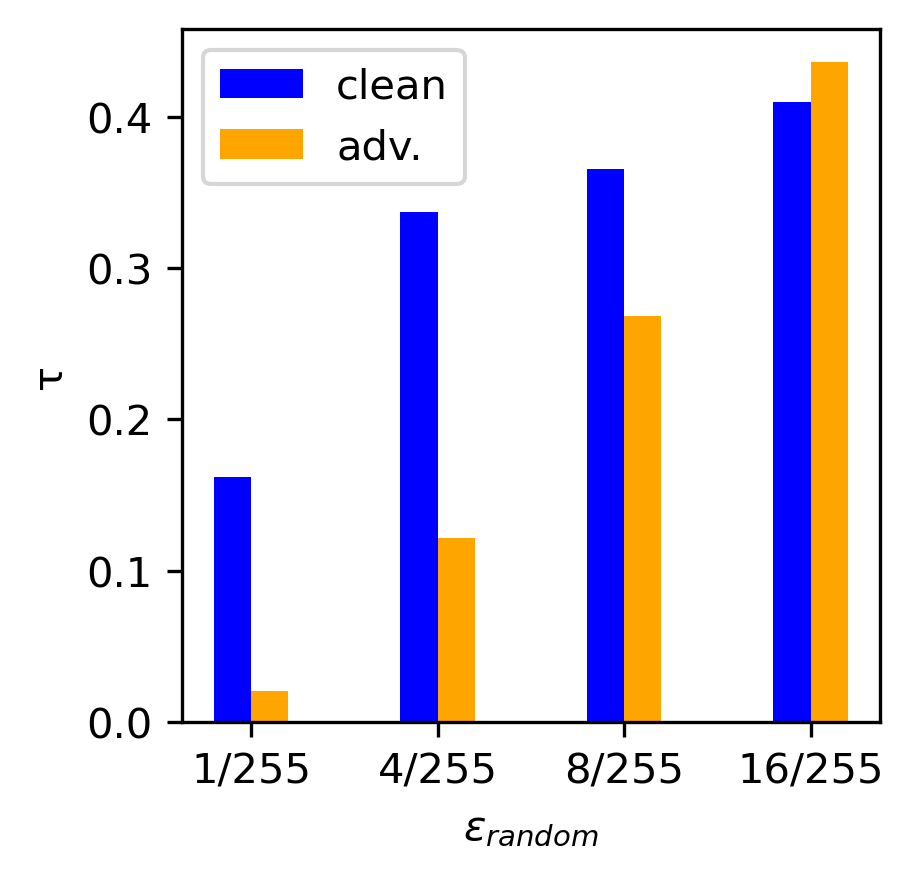}  
        \caption{Caltech101}  
        \label{fig:tau_caltech101}
    \end{subfigure}
    \hfill
    \begin{subfigure}{0.16\textwidth}
        \centering
        \includegraphics[width=\textwidth]{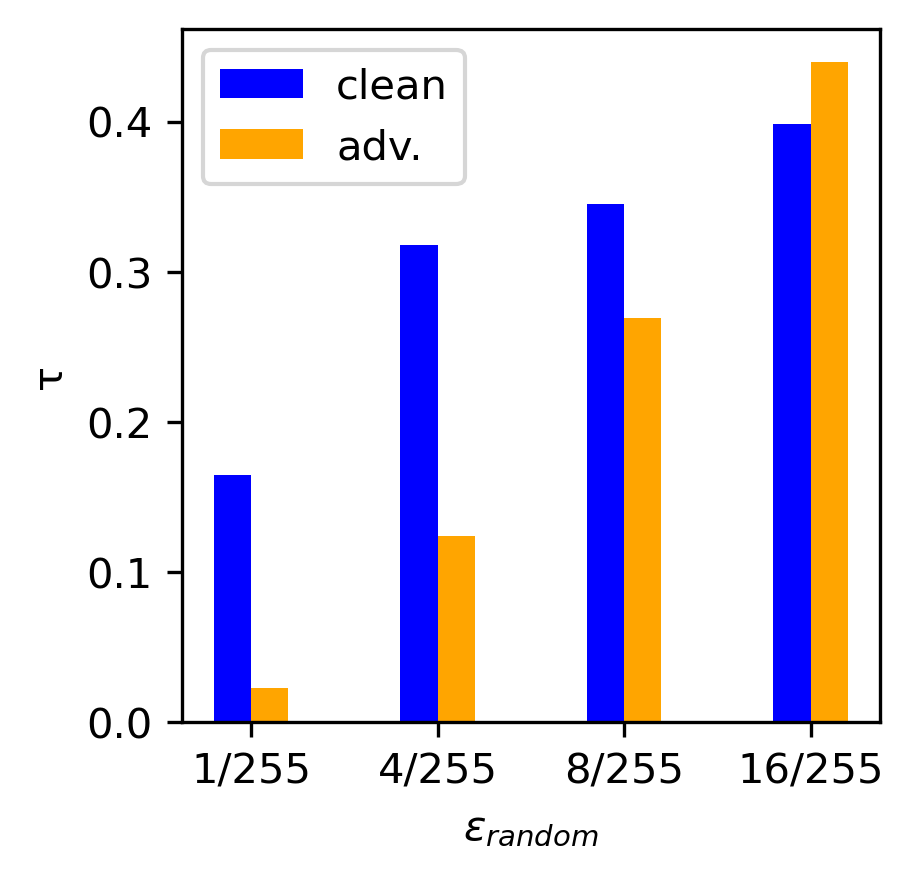}
        \caption{Caltech256}
        \label{fig:tau_caltech256}
    \end{subfigure}
    \hfill
    \begin{subfigure}{0.16\textwidth}
        \centering
        \includegraphics[width=\textwidth]{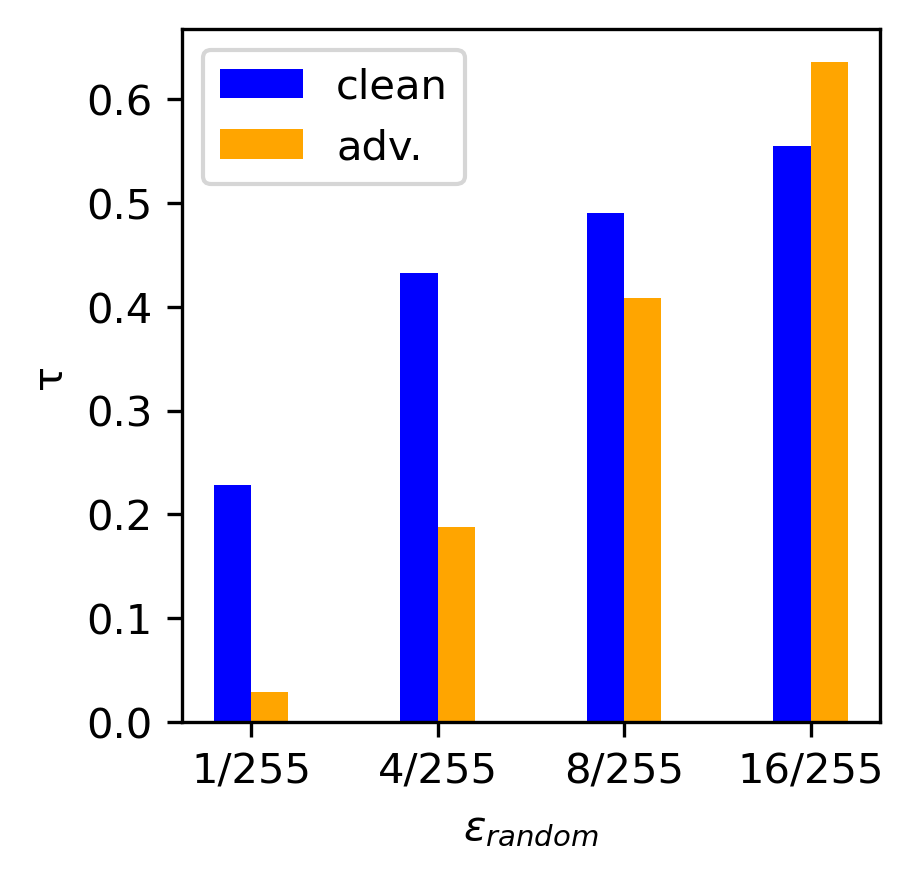}
        \caption{CIFAR10}
        \label{fig:tau_cifar10}
    \end{subfigure}
    \hfill
    \begin{subfigure}{0.16\textwidth}
        \centering
        \includegraphics[width=\textwidth]{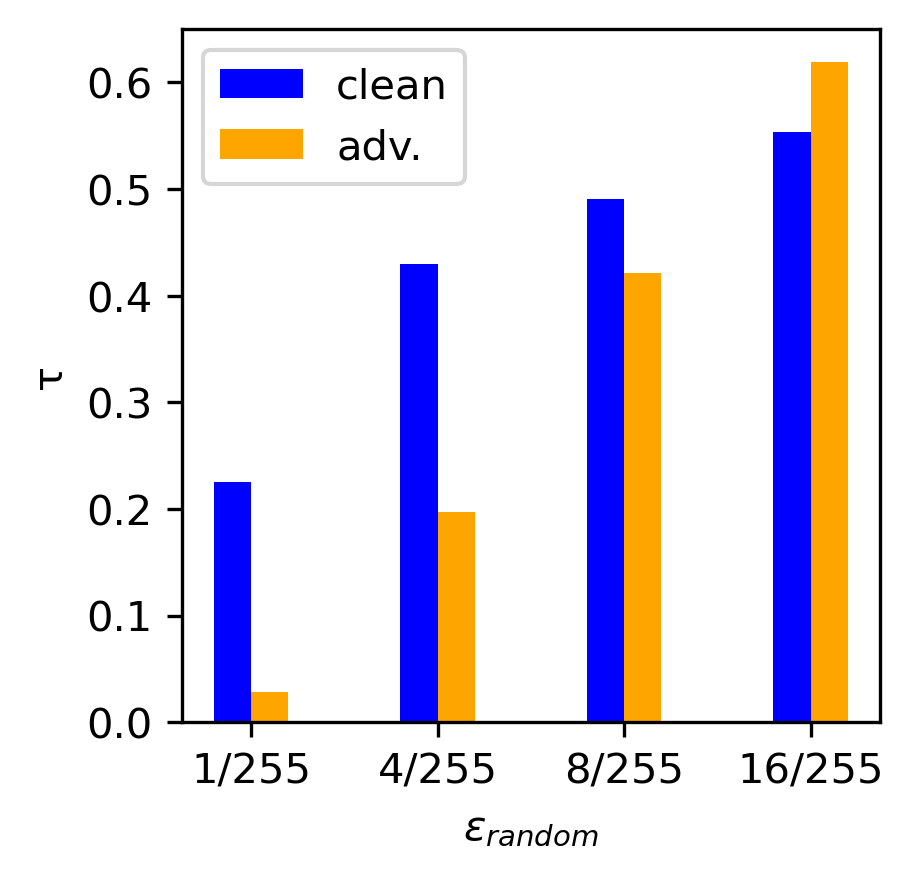}
        \caption{CIFAR100}
        \label{fig:tau_cifar100}
    \end{subfigure}
    \hfill
    \begin{subfigure}{0.16\textwidth}
        \centering
        \includegraphics[width=\textwidth]{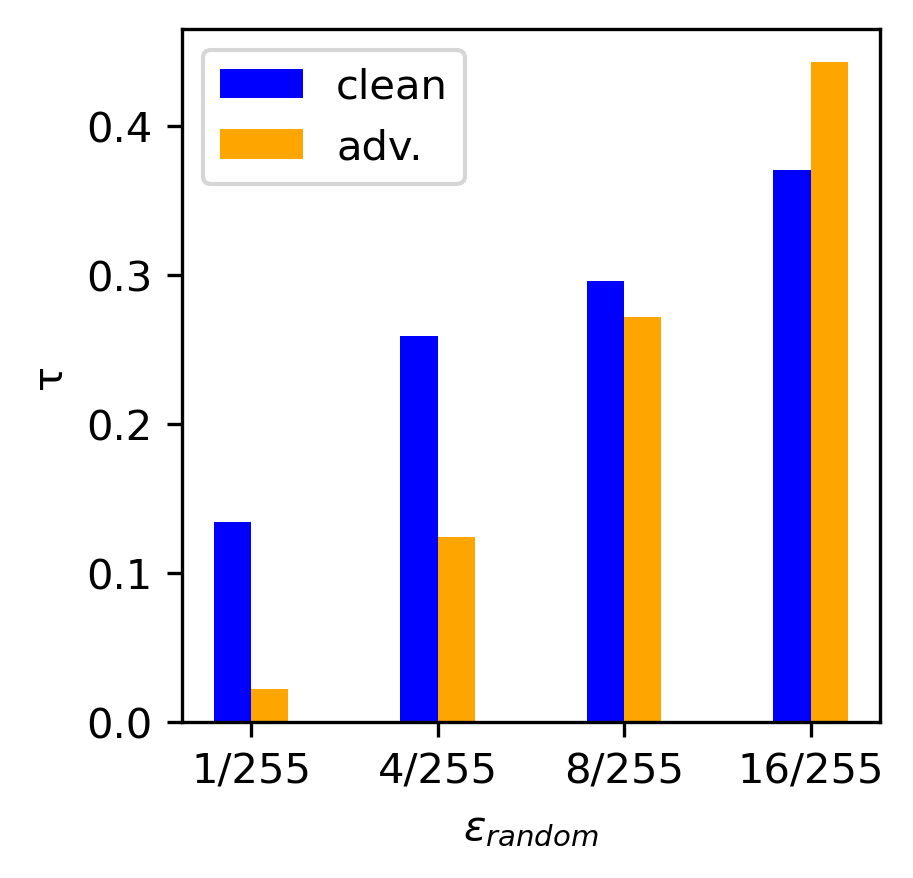}
        \caption{Country211}
        \label{fig:tau_country211}
    \end{subfigure}
    \hfill
    \begin{subfigure}{0.16\textwidth}
        \centering
        \includegraphics[width=\textwidth]{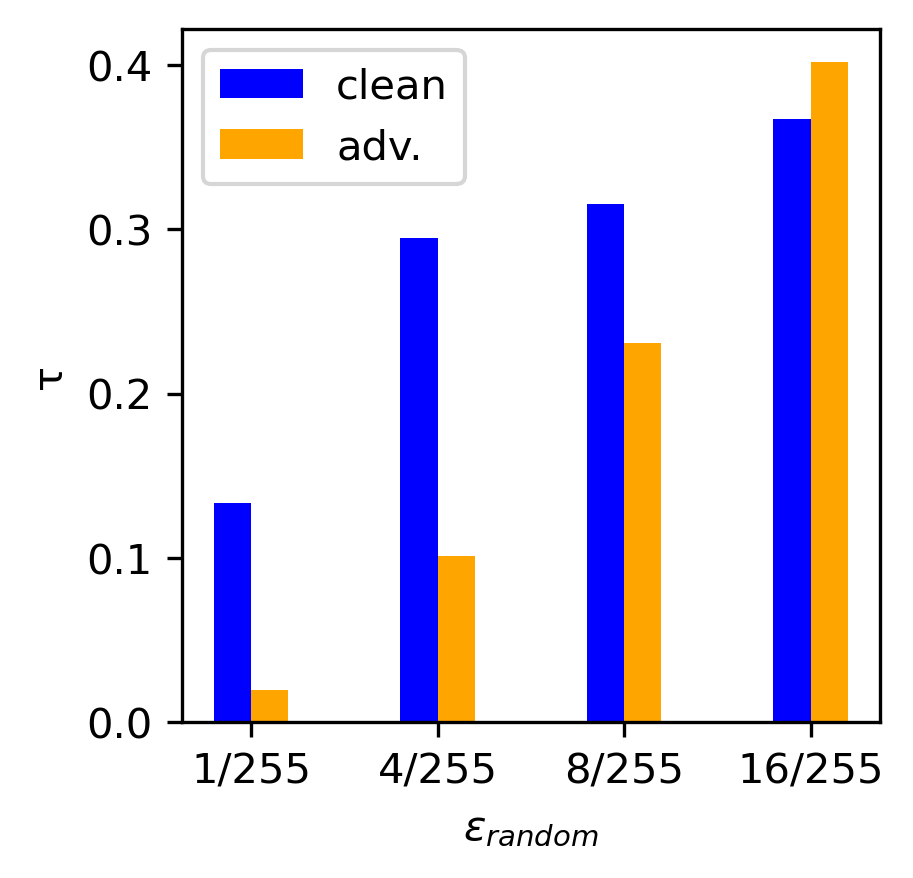}
        \caption{DTD}
        \label{fig:tau_dtd}
    \end{subfigure}

    \vskip\baselineskip  
    
    \begin{subfigure}{0.16\textwidth}
        \centering
        \includegraphics[width=\textwidth]{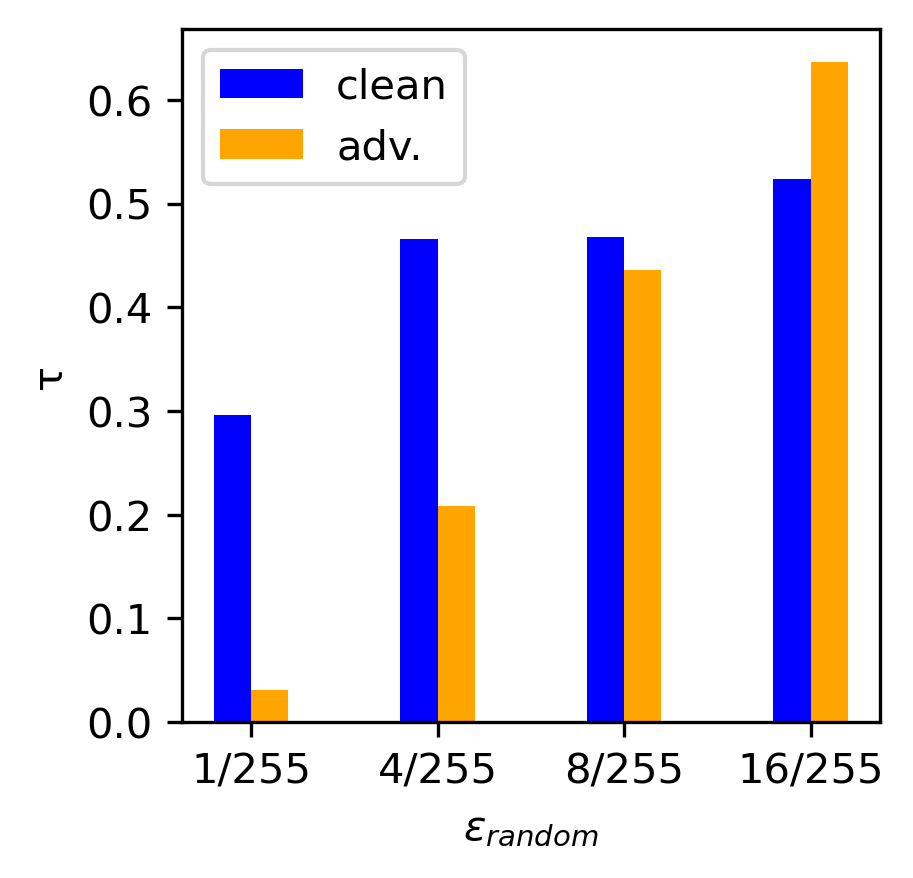}
        \caption{EuroSAT}
        \label{fig:tau_eurosat}
    \end{subfigure}
    \hfill
    \begin{subfigure}{0.16\textwidth}
        \centering
        \includegraphics[width=\textwidth]{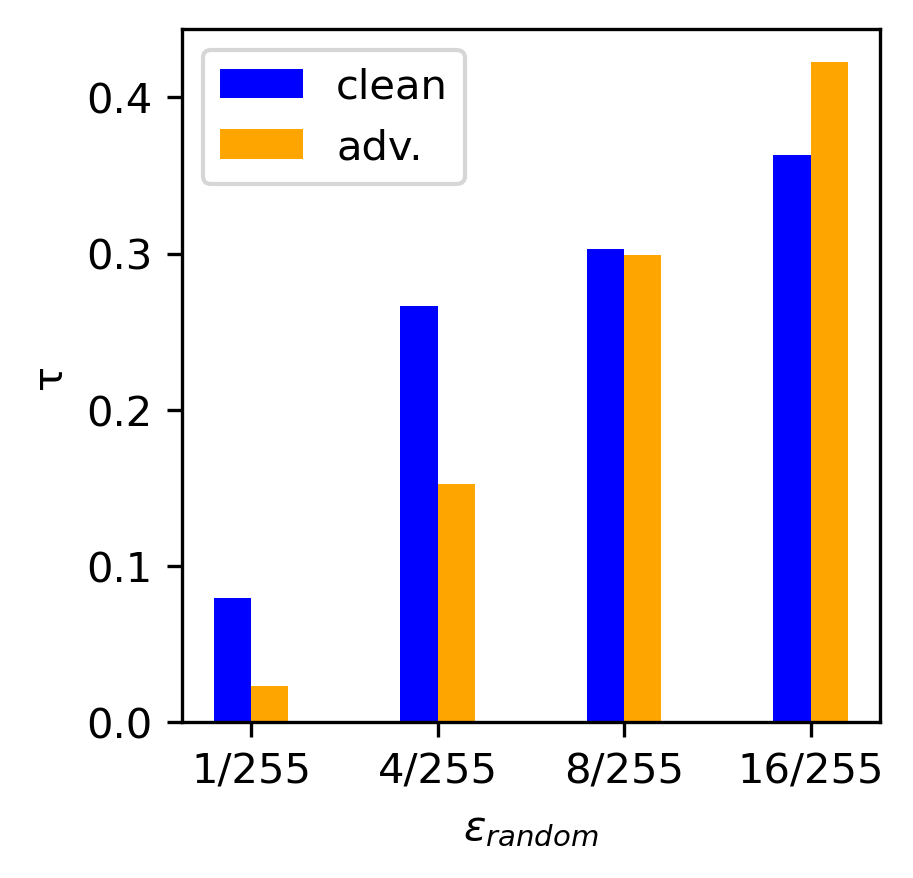}
        \caption{FGVCAircraft}
        \label{fig:tau_fgvc}
    \end{subfigure}
    \begin{subfigure}{0.16\textwidth}
        \centering
        \includegraphics[width=\textwidth]{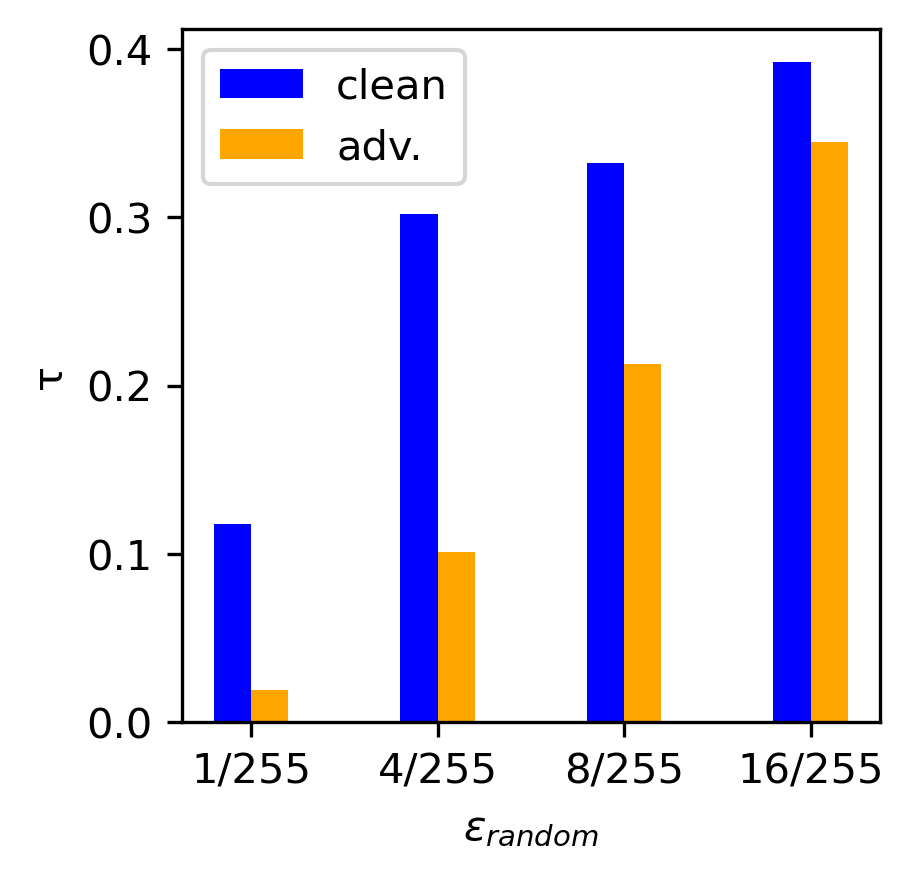}
        \caption{Flowers102}
        \label{fig:tau_flowers}
    \end{subfigure}
    \hfill
    \begin{subfigure}{0.16\textwidth}
        \centering
        \includegraphics[width=\textwidth]{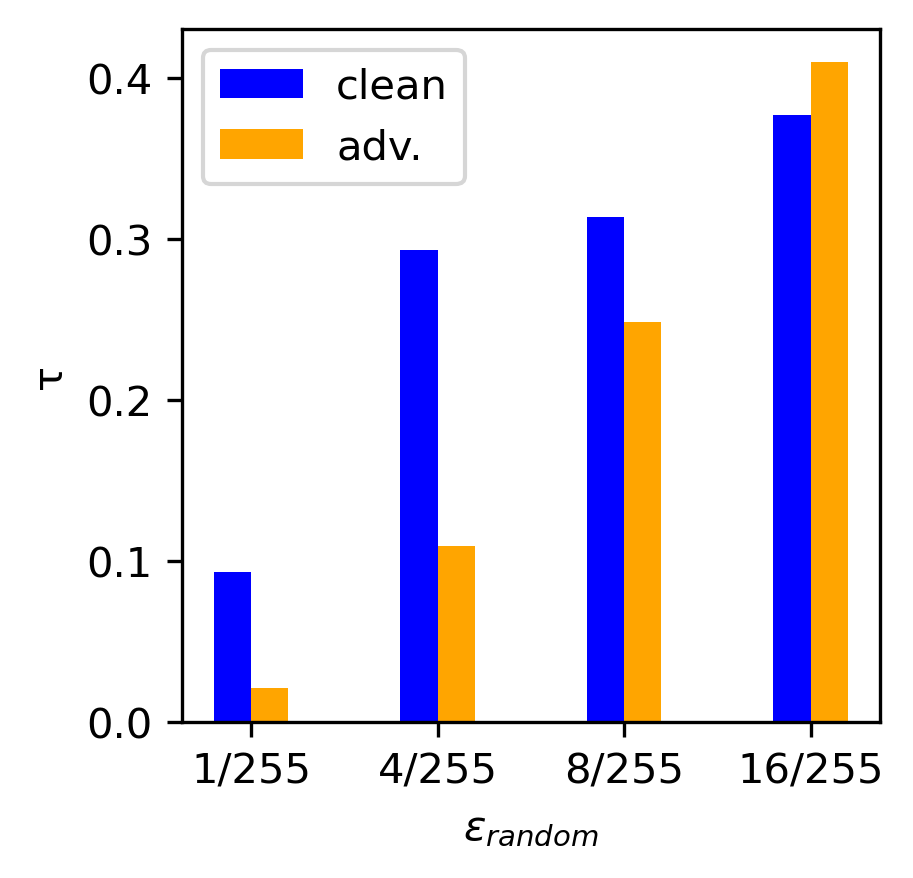}
        \caption{Food101}
        \label{fig:tau_food}
    \end{subfigure}
    \hfill
    \begin{subfigure}{0.16\textwidth}
        \centering
        \includegraphics[width=\textwidth]{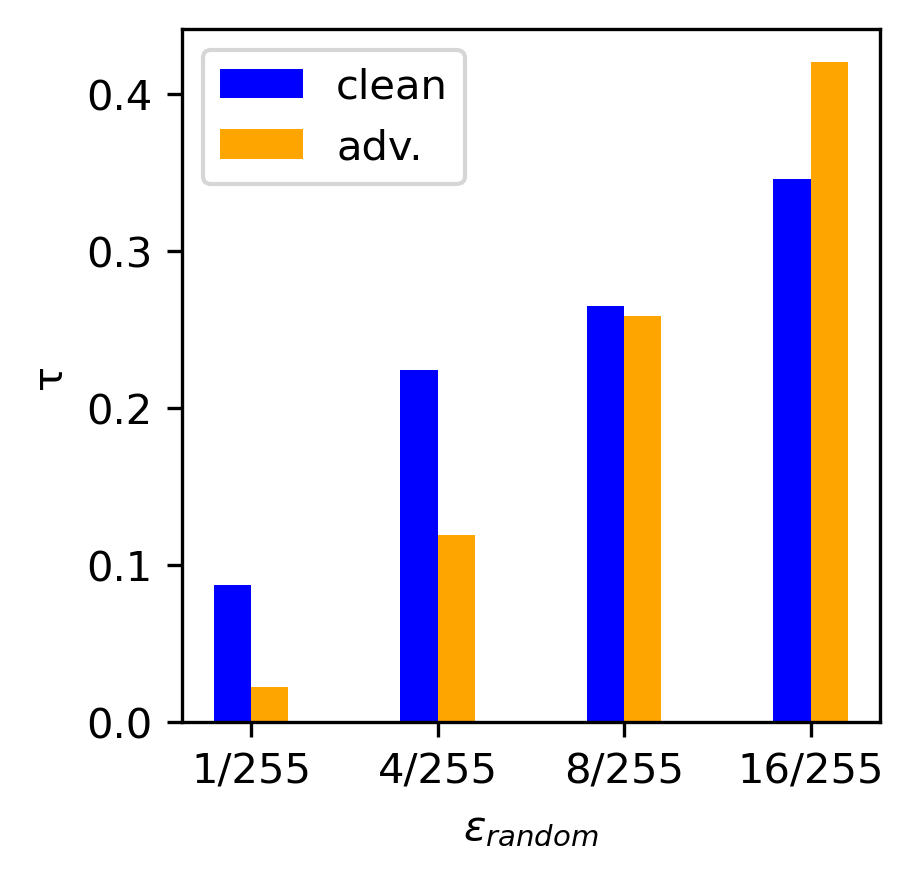}
        \caption{ImageNet}
        \label{fig:tau_imagenet}
    \end{subfigure}
    \hfill
    \begin{subfigure}{0.16\textwidth}
        \centering
        \includegraphics[width=\textwidth]{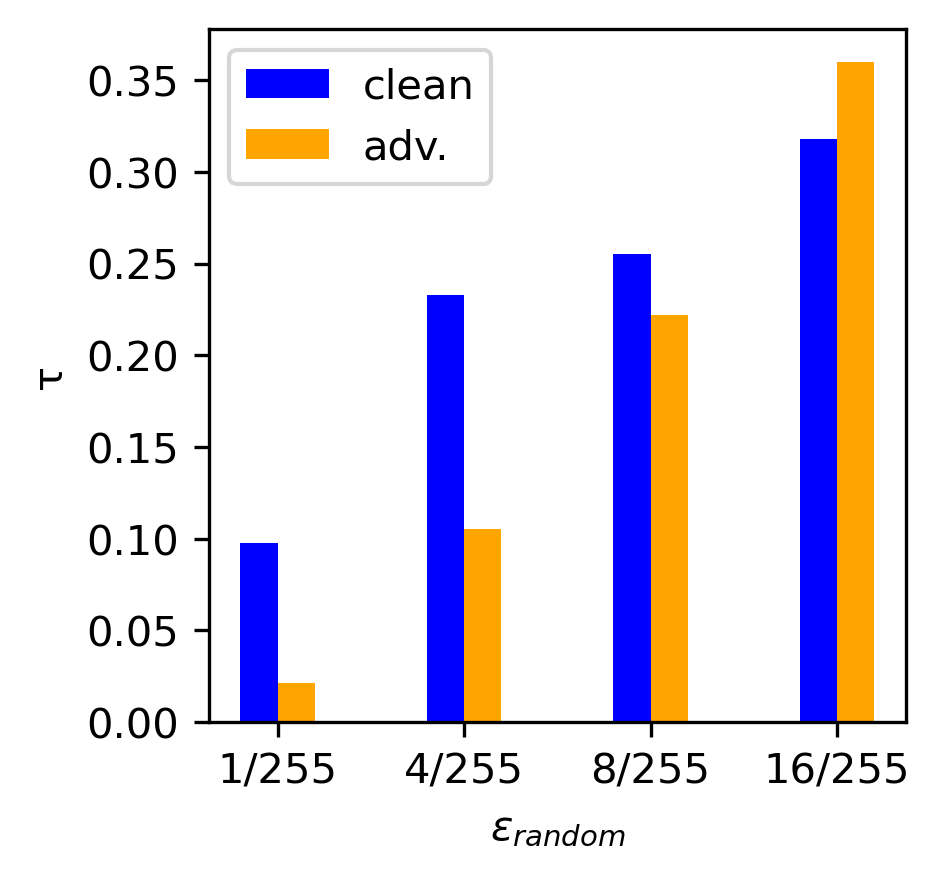}
        \caption{OxfordPets}
        \label{fig:tau_oxfordpets}
    \end{subfigure}

    \vskip\baselineskip  
    
    \begin{subfigure}{0.16\textwidth}
        \centering
        \includegraphics[width=\textwidth]{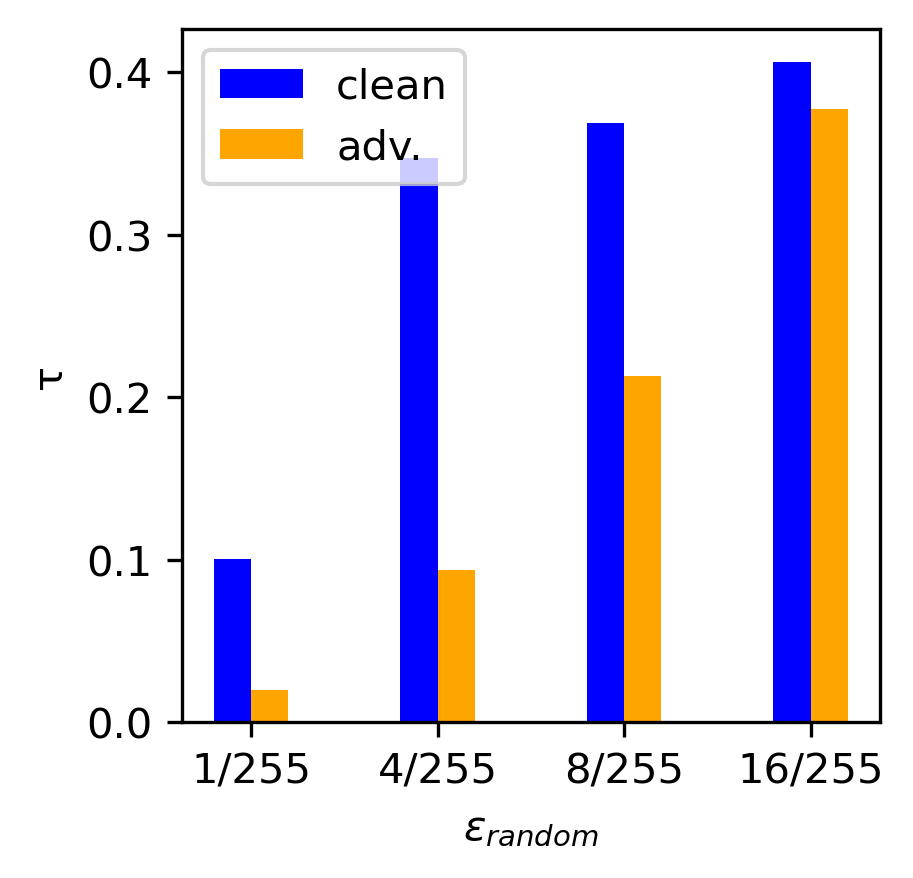}
        \caption{PCAM}
        \label{fig:tau_pcam}
    \end{subfigure}
    \hfill
    \begin{subfigure}{0.16\textwidth}
        \centering
        \includegraphics[width=\textwidth]{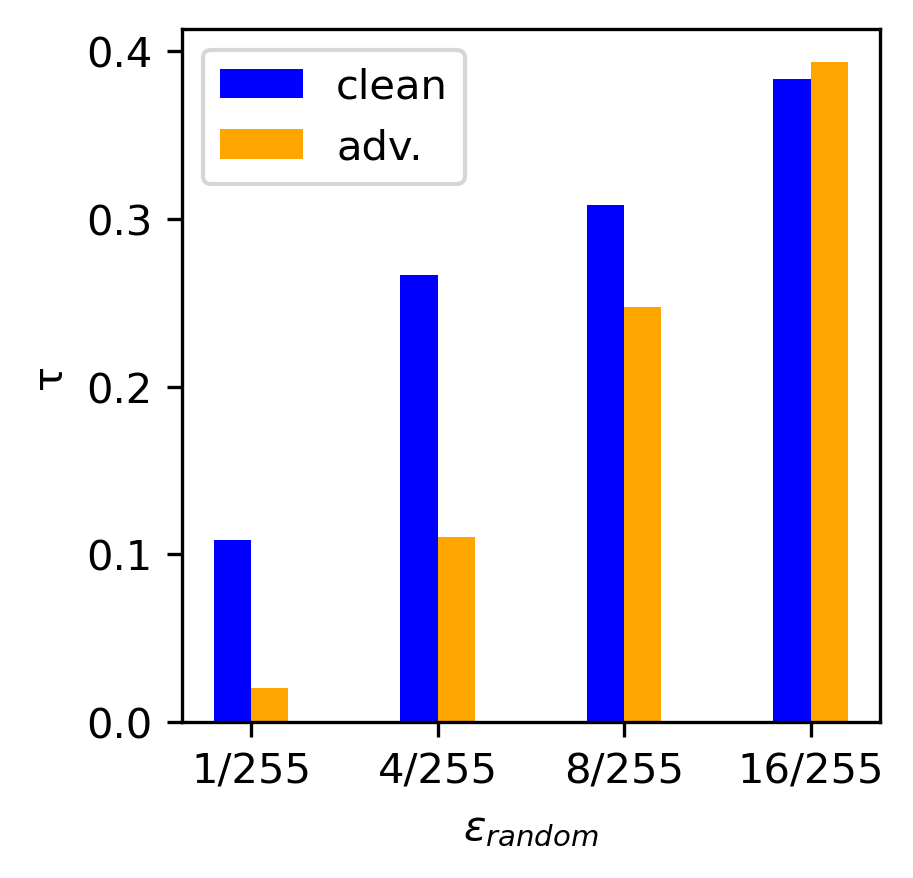}
        \caption{StanfordCars}
        \label{fig:tau_stanfordcars}
    \end{subfigure}
    \hfill
    \begin{subfigure}{0.16\textwidth}
        \centering
        \includegraphics[width=\textwidth]{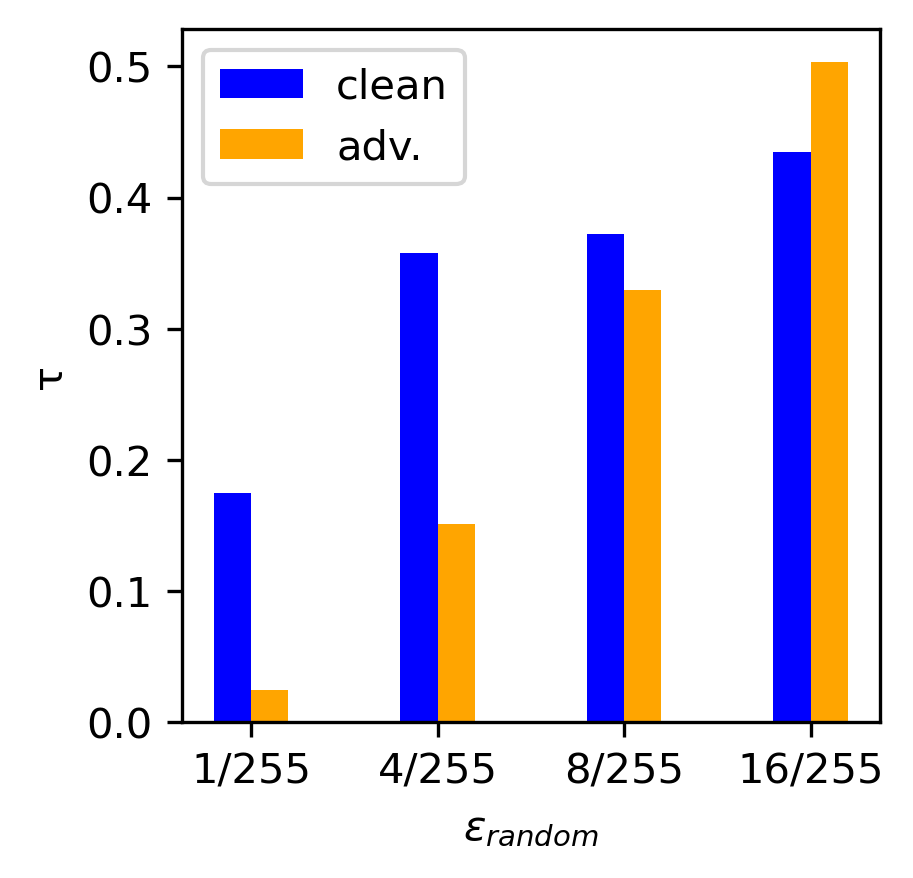}
        \caption{STL10}
        \label{fig:tau_stl10}
    \end{subfigure}
    \hfill
    \begin{subfigure}{0.16\textwidth}
        \centering
        \includegraphics[width=\textwidth]{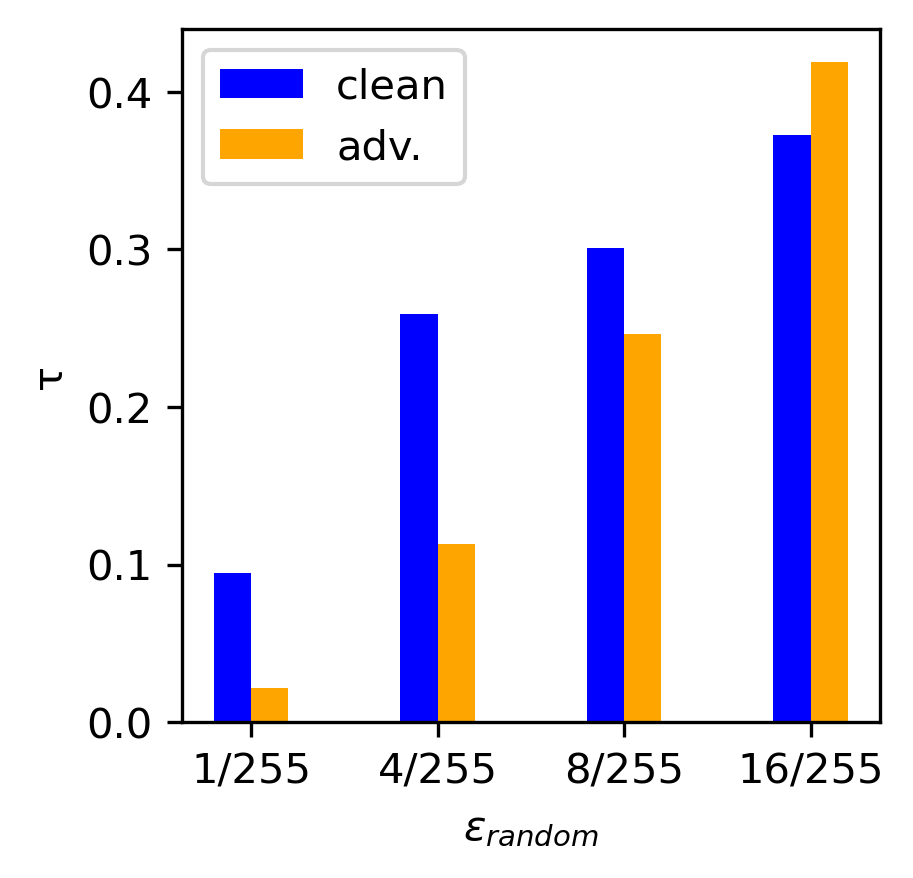}
        \caption{SUN397}
        \label{fig:tau_sun397}
    \end{subfigure}
    \hfill
    \begin{subfigure}{0.16\textwidth}
    \end{subfigure}
    \hfill
    \begin{subfigure}{0.16\textwidth}
    \end{subfigure}

    \caption{Values of $\tau$ on clean and adversarial images ($\epsilon_a=1/255$) across 16 datasets. For each dataset, we randomly sample 100 images and report the average values.}
    \label{fig:tau_full_analysis}
    \vspace{-0.55cm}
\end{figure*}

\section{Analysis of $\tau$}
\label{sec:tau_analysis}
In the paper, we define a stochastic variable $\tau$ (\cref{eq:tau}). In this section, we provide more results of $\tau$ and theoretical analyses. We report the values of $\tau$ for 16 datasets in \cref{fig:tau_full_analysis}. As can be seen from \cref{fig:tau_full_analysis}, when the random noise added onto the image is small, the resultant $L_2$ drift of the adversarial images in the embedding space is unusually small, indicating that they are trapped in their toxic local surroundings induced by the adversaries that seek to maximise the classification loss of CLIP.
This behaviour is termed as `false stability' in the main paper.
When the strength of the random noise is sufficiently large, the $L_2$ drift of adversarial images is disproportionately enlarged.
In contrast, the values of $\tau$ increase more steadily for clean images, as the noise strength $\epsilon_{random}$ increases, without showing disproportionate changes.
Below we theoretically analyse the behaviour of `false stability' of adversarial images.

\subsection{Theoretical Analysis}
Given a pre-trained vision encoder $f_\theta$, a natural (unattacked) image $x\in\mathcal{R}^{C\times W\times H}$, and an adversarial image $x'$ that is manipulated to maximise the classification loss of CLIP:
\begin{equation}\label{eq:advimg}
    x' = \arg\max\limits_{x_s} L(f_\theta(x_s), t_c),\;\;s.t.\,\| x_s - x\|_{\infty} \leq\epsilon
\end{equation}
the resultant embedding $f_\theta(x+n)$ when a small random noise $n\in\mathcal{R}^{C\times W\times H}\thicksim U(-\epsilon_{random}, \epsilon_{random})$ is imposed can be written as the Taylor expansion of $f$ at $x$:
\begin{equation}
    f_\theta(x+n) = f_\theta(x) + J_f(x)\cdot n + \frac{1}{2}n^T \cdot H_f(x)\cdot n + \cdots
\end{equation}
where $J_f(x)$ and $H_f(x)$ are the Jacobian matrix and Hessian tensor of $f$ at $x$, respectively,
assuming that $f$ is smooth around $x$.
Provided that the random noise $n$ is small, the above embedding can be approximated by the first-order expansion:
\begin{equation}
    f_\theta(x+n) \approx f_\theta(x) + J_f(x)\cdot n
\end{equation}
Therefore, the $L_2$ drift induced by $n$ can be written as:
\begin{equation}\label{eq:activation}
\begin{split}
    \parallel f_\theta(x+n)-f_\theta(x)\parallel & \approx \parallel J_f(x)\cdot n\parallel \\
    & = \left( \sum\limits_{j=1}^{d} \left(
    \sum\limits_{i=1}^{N}\frac{\partial f_j}{\partial x_i}n_i \right)^2\right)^{\frac{1}{2}}
\end{split}
\end{equation}
where $d$ is the latent space dimensionality of CLIP, and $N=C\times W\times H$ is the pixel space dimensionality.

When \cref{eq:advimg} is computed by gradient-based methods such as PGD \cite{carlini2017towards}, $x'$ is obtained through gradient ascent in the direction that increases the classification loss $L$:
\begin{equation}\label{eq:jacob}
    \frac{\partial }{\partial x} L(f_\theta(x), t_c) = \frac{\partial L}{\partial f}\cdot \frac{\partial f}{\partial x} = \frac{\partial L}{\partial f} \cdot J_f(x)
\end{equation}
As such, the approximation of \cref{eq:advimg} can be seen as constantly searching the pixel space for the trajectory starting from $x$ that causes the steepest ascent of $L$, i.e., the strongest activation of $J_f(x)$, within a limited number of steps.
Since $x'$ is the approximation result of \cref{eq:advimg}, it lies in the trajectory
where $J_f(x)$ is the most activated,
and is therefore insensitive to a random noise $n$, which is statistically isotropic in the pixel space with a tiny component that lies in the direction of $\frac{\partial L}{\partial f}\cdot J_f(x)|_{x=x'}$.
In contrast, a clean image $x$ without being manipulated based on $J_f(x)$ does not show unusually strong activations in any direction, and can be more activated by an isotropic noise $n$. Therefore,
$\| J_f(x)\cdot n\| > \| J_f(x')\cdot n \|$ holds when $n$ is a small random noise in the pixel space, rendering the adversarial image $x'$ `falsely stable'.


\section{More Results on Adversarial Robustness}
In this section, we provide more complete results on adversarial robustness.

\subsection{Robustness under CW attacks}\label{sec:cw_attacks}
Following previous studies \cite{mao2023understanding,wang2024pre}, we further test adversarial robustness of our test-time counterattack paradigm under CW attack \cite{carlini2017towards}, with the attack budget at $\epsilon_a=1/255$. \cref{tab:white_box_eps1_CW} reports the full table of results.
It can be seen that for CW attacks, 
our TTC paradigm can still achieve stable robustness gains across 16 datasets. 
RN and TTE do not degrade accuracy on clean images since they do not counter the potential adversary by perturbing test images.
Similarly to when tested under PGD attacks, TTE does not provide stable robustness.
Compared to \textit{Anti-adversary} and \textit{HD}, which optimise a perturbation based on some objective, our TTC retains the best clean accuracy while significantly improving robustness. 
This shows that our paradigm can also be employed in test time to defend CLIP against other attack methods that maximise the classification loss of CLIP.

\begin{table*}[t]
    \centering
    \resizebox{\textwidth}{!}{
    \begin{tabular}{cc|c|cccc||ccccc|c}
         \toprule
         
         \multicolumn{2}{c|}{\multirow{2}{*}{(\%)}} & \multirow{2}{*}{CLIP} & \multicolumn{4}{c||}{Adversarial Finetuning} & \multicolumn{5}{c|}{Test-time Defence} & \multirow{2}{*}{$\Delta$} \\
         \cline{4-7}\cline{8-12}
         & & &  CLIP-FT & TeCoA & PMG-AFT & FARE & RN & TTE & Anti-adv & HD & TTC (ours) & \\
         \midrule
         \color{lightgray}\multirow{2}{*}{TinyImageNet} & \color{lightgray}Rob. & \color{lightgray}0.36 & \color{lightgray}1.06 & \color{lightgray}48.00 & \color{lightgray}43.79 & \color{lightgray}27.71 &\color{lightgray}0.57$\pm0.02$ &\color{lightgray}19.40$\pm4.08$ &\color{lightgray}5.48$\pm0.05$ &\color{lightgray}3.70$\pm0.12$ &\color{lightgray}19.75$\pm0.38$ &\color{lightgray}+19.39 \\
         
         & \color{lightgray}Acc.& \color{lightgray}57.64 & \color{lightgray}77.06 & \color{lightgray}70.86 &\color{lightgray}66.85 &\color{lightgray}73.63 &\color{lightgray}51.85$\pm0.04$ &\color{lightgray}56.73$\pm0.22$ &\color{lightgray}52.76$\pm0.16$ &\color{lightgray}52.49$\pm0.12$ &\color{lightgray}51.85$\pm0.04$ &\color{lightgray}-5.79
         \\
         \midrule
         
         \multirow{2}{*}{CIFAR10} & Rob. &0.87 &0.94 &33.27 &39.50 &20.6 &2.05\std{0.05} &\textbf{40.01}\std{6.25} &12.53\std{0.01} &14.79\std{0.10} &29.04\std{0.02} &\redd{+28.17} \\
         & Acc.& 85.12 & 84.90 & 64.61 & 70.69 & 74.44 &81.18\std{0.07} &\textbf{84.74}\std{0.40} &83.52\std{0.09} &78.64\std{0.02} &81.18\std{0.07} &\blue{-3.94} \\
         
         \midrule
         
         \multirow{2}{*}{CIFAR100} & Rob. &0.29 &0.39 &18.27 &20.83 &11.67 &0.63\std{0.06} &\textbf{18.73}\std{3.87} &6.56\std{0.23} &3.04\std{0.04} &14.38\std{0.23} &\redd{+14.09} \\
         & Acc.& 57.14 & 59.51 & 35.96 & 40.32 & 46.67 &56.34\std{0.20} &\textbf{58.61}\std{0.25} &53.95\std{0.15} &53.50\std{0.02} &56.34\std{0.20} &\blue{-0.80} \\
         
         \midrule
         
         \multirow{2}{*}{STL10} & Rob. &12.23 &9.95 &69.73 &72.39 &59.60 &17.20\std{0.15} &\textbf{78.64}\std{3.91} &38.66\std{0.17} &37.73\std{0.22} &76.40\std{0.16} &\redd{+64.17} \\
         & Acc.& 96.40 & 94.49 & 87.40 & 88.56 & 91.72 &95.85\std{0.04} &\textbf{96.26}\std{0.04} &95.45\std{0.08} &89.54\std{0.05} &95.85\std{0.04} &\blue{-0.55} \\
         
         \midrule
         
         \multirow{2}{*}{ImageNet} & Rob. &1.46 &1.27 &18.28 &19.42 &27.71 &2.21\std{0.00} &29.77\std{4.19} &9.37\std{0.05} &7.46\std{0.05} &\textbf{36.01}\std{0.15} &\redd{+34.55} \\
         & Acc.& 59.69 & 54.24 & 34.89 & 36.12 & 48.79 &59.34\std{0.06} &\textbf{60.02}\std{0.13} &54.27\std{0.14} &55.06\std{0.05} &49.39\std{0.00} &\blue{-10.30} \\
         
         \midrule
         
         \multirow{2}{*}{Caltech101} & Rob. &20.88 &15.95 &56.23 &61.58 &54.86 &25.89\std{0.11} &\textbf{69.44}\std{3.09} &41.47\std{0.02} &36.26\std{0.08} &66.17\std{0.31} &\redd{+45.29} \\
         & Acc.& 85.66 & 83.63 & 71.68 & 75.45 & 80.95 &\textbf{86.61}\std{0.10} &85.84\std{0.09} &84.02\std{0.10} &83.00\std{0.07} &86.53\std{0.07} &\redd{+0.87} \\
         
         \midrule
         
         \multirow{2}{*}{Caltech256} & Rob. &9.69 &7.24 &42.63 &44.55 &39.58 &13.11\std{0.05} &\textbf{59.81}\std{3.97} &27.17\std{0.07} &24.54\std{0.09} &58.79\std{0.07} &\redd{+49.10} \\
         & Acc.& 81.72 & 78.53 & 61.14 & 62.24 & 73.32 &81.25\std{0.03} &\textbf{82.48}\std{0.08} &79.38\std{0.12} &79.38\std{0.05} &79.66\std{0.04} & \blue{-2.06}\\
         
         \midrule
         
         \multirow{2}{*}{OxfordPets} & Rob. &1.64 &1.14 &37.91 &39.28 &33.85 &3.11\std{0.04} &51.12\std{6.98} &22.99\std{0.52} &13.84\std{0.27} &\textbf{57.15}\std{0.61} &\redd{+55.51} \\
         & Acc.& 87.44 & 84.14 & 62.12 & 65.88 & 79.37 &87.41\std{0.12} &\textbf{88.13}\std{0.13} &80.62\std{0.35} &80.64\std{0.15} &83.35\std{0.21} &\blue{-4.09} \\

         \midrule
         
         \multirow{2}{*}{Flowers102} & Rob. &1.35 &0.80 &21.13 &21.34 &17.25 &2.13\std{0.06} &34.97\std{4.25} &8.06\std{0.07} &8.51\std{0.04} &\textbf{36.84}\std{0.13} &\redd{+35.49} \\
         & Acc.& 65.46 & 53.37 & 36.80 & 37.00 & 47.98 &64.62\std{0.19} &\textbf{65.20}\std{0.23} &62.66\std{0.14} &57.79\std{0.08} &64.16\std{0.19} &\blue{-1.30} \\
         
         \midrule
         
         \multirow{2}{*}{FGVCAircraft} & Rob. &0.00 &0.00 &2.25 &1.86 &1.35 &0.00\std{0.00} &5.15\std{1.25} &0.83\std{0.11} &0.97\std{0.06} &\textbf{12.41}\std{0.32} &\redd{+12.41} \\
         & Acc.& 20.10 & 14.04 & 5.31 & 5.55 & 10.86 &19.25\std{0.18} &\textbf{20.18}\std{0.35} &15.88\std{0.23} &16.18\std{0.21} &18.00\std{0.16} &\blue{-2.10} \\
         
         \midrule
         
         \multirow{2}{*}{StanfordCars} & Rob. &2.38 &2.04 &8.74 &10.53 &9.14 &2.44\std{0.02} &21.19\std{3.41} &4.76\std{0.18} &5.11\std{0.05} &\textbf{30.38}\std{0.12} &\redd{+28.00} \\
         & Acc.& 52.02 & 42.11 & 20.91 & 25.44 & 38.68 &52.14\std{0.09} &\textbf{52.73}\std{0.31} &36.21\std{0.27} &43.60\std{0.05} &48.16\std{0.16} &\blue{-3.86} \\
         
         \midrule
         
         \multirow{2}{*}{SUN397} & Rob. &1.75 &1.48 &18.36 &20.39 &15.73 &2.48\std{0.03} &29.37\std{4.05} &8.85\std{0.01} &7.90\std{0.03} &\textbf{39.44}\std{0.07} &\redd{+37.69} \\
         & Acc.& 58.50 & 55.73 & 36.69 & 37.98 & 52.42 &\textbf{59.69}\std{0.06} &59.12\std{0.08} &56.00\std{0.04} &54.07\std{0.01} &55.13\std{0.06} & \blue{-3.37}\\
         
         \midrule
         
         \multirow{2}{*}{Country211} & Rob. &0.08 &0.05 &1.46 &1.74 &0.92 &0.15\std{0.02} &3.00\std{0.74} &0.72\std{0.05} &0.75\std{0.02} &\textbf{6.17}\std{0.11} &\redd{+6.09}\\
         & Acc.& 15.25 & 12.07 & 4.75 & 4.64 & 9.26 &\textbf{14.80}\std{0.02} &14.66\std{0.14} &11.58\std{0.12} &11.98\std{0.02} &13.08\std{0.05} &\blue{-2.17} \\
         
         \midrule
         
         \multirow{2}{*}{Food101} & Rob. &1.09 &0.55 &12.87 &16.57 &12.93 &1.92\std{0.04} &44.61\std{6.42} &15.03\std{0.11} &9.77\std{0.06} &\textbf{54.65}\std{0.13} &\redd{+53.56} \\
         & Acc.& 83.88 & 64.86 & 29.98 & 36.61 & 55.31 &83.44\std{0.04} &\textbf{83.96}\std{0.01} &75.81\std{0.22} &81.02\std{0.05} &82.18\std{0.02} &\blue{-1.70}\\
         
         \midrule
         
         \multirow{2}{*}{EuroSAT} & Rob. &0.03 &0.03 &11.66 &11.94 &10.66 &0.16\std{0.00} &6.44\std{1.74} &2.57\std{0.08} &3.47\std{0.17} &\textbf{12.69}\std{0.07} &\redd{+12.66} \\
         & Acc.& 42.59 & 27.64 & 16.58 & 18.53 & 21.88 &\textbf{53.24}\std{0.09} &44.38\std{1.62} &36.78\std{0.18} &40.12\std{0.13} &\textbf{53.24}\std{0.09} &\redd{+10.65} \\
         
         \midrule
         
         \multirow{2}{*}{DTD} & Rob. &2.87 &2.77 &16.28 &13.72 &14.36 &3.46\std{0.04} &22.62\std{2.06} &6.06\std{0.04} &10.11\std{0.16} &\textbf{27.39}\std{1.07} &\redd{+24.52} \\
         & Acc.& 40.64 & 36.49 & 25.16 & 21.76 & 32.07 &37.96\std{0.13} &\textbf{41.35}\std{0.29} &38.92\std{0.22} &35.25\std{0.22} &36.98\std{0.21} &\blue{-3.66} \\
         
         \midrule
         
         \multirow{2}{*}{PCAM} & Rob. &0.10 &1.10 &48.29 &46.36 &16.41 &0.44\std{0.02} &10.70\std{3.25} &5.07\std{0.02} &46.92\std{0.10} &\textbf{52.86}\std{0.06} &\redd{+52.76} \\
         & Acc.& 52.02 & 47.21 & 49.96 & 50.03 & 52.54 &\textbf{52.73}\std{0.07} &50.92\std{0.04} &52.49\std{0.02} &50.35\std{0.01} &\textbf{52.73}\std{0.07} &\redd{+0.71} \\
         
         \midrule
         \midrule
         
         \multirow{2}{*}{\textbf{Avg.}} & Rob. &3.54 &2.86 &26.09 &27.62 &20.86 &4.84\std{0.01} & 32.85\std{3.70} &13.17\std{0.04} &14.45\std{0.03} &\textbf{38.17}\std{0.09} &\redd{+34.63} \\
         & Acc. & 61.51 & 55.80 & 40.25 & 42.30 & 51.02 &61.61\std{0.03} & \textbf{61.79}\std{0.13} &57.35\std{0.03} &56.88\std{0.02} &59.75\std{0.06} &\blue{-1.76} \\
         \bottomrule
    \end{tabular}}
    \caption{\label{tab:white_box_eps1_CW}Classification accuracy (\%) on both adversarial images (Rob.) under 10-step CW attack \cite{carlini2017towards} at $\epsilon_a=1/255$ and clean images (Acc.) across 16 datasets. 
    Weights and gradients of the deployed model are assumed to be known to the threat model. Comparison is made among our paradigm and test-time defences adapted from existing adversarial studies, with finetuning-based models implemented as a reference. We report the mean and standard deviation for test-time methods over 3 runs. The last column reports the gains w.r.t. original CLIP without any finetuning or test-time operations.}
\end{table*}

\subsection{Robustness under $\epsilon_a=4/255$}
\cref{tab:white_box_eps4_full} reports the full table of robustness under 10-step PGD attack with the attack budget being $\epsilon_a=4/255$. 
It can be seen that TTC achieves consistent and stable robustness gains across 16 datasets. \textit{Anti-adversary} \cite{alfarra2022combating} and \textit{HD} \cite{wu2021attacking} bring little to no robustness under a high attack strength at $\epsilon_{a}=4/255$.
RN and TTE \cite{perez2021enhancing} perform best in terms of accuracy on clean images, which is understandable because they do not optimise any perturbation to counter the adversary.
RN does not provide any robustness, showing that an additive random noise in the pixel space as large as the attack budget is not able to counteract the false stability of adversarial images.
TTE \cite{perez2021enhancing} improves robustness of CLIP against PGD attacks with a high strength $\epsilon_a=4/255$ to some extent. However, the robustness gain is unstable, as indicated by the high standard deviation of robust accuracy across different runs. 
For TTC, the number of steps $N$ is increased to 5 for more effective counterattacks in this setting, which reduces the average clean accuracy by $5.52$, compared to the original CLIP model. This trade-off is still reasonable given the consistent robustness gains.

\begin{table*}[t]
    \centering
    \resizebox{\textwidth}{!}{
    \begin{tabular}{cc|c|ccccccc||ccccc|c}
         \toprule
         \multicolumn{2}{c|}{\multirow{2}{*}{(\%)}} & \multirow{2}{*}{CLIP} & \multicolumn{7}{c||}{Adversarial Finetuning} & \multicolumn{5}{c|}{Test-time Defence} & \multirow{2}{*}{$\Delta$} \\
         \cline{4-10}\cline{8-15}
         & & &  CLIP-FT & $\textrm{TeCoA}^1$ & $\textrm{TeCoA}^4$ & $\textrm{PMG-AFT}^1$ & $\textrm{PMG-AFT}^4$ & $\textrm{FARE}^1$ & $\textrm{FARE}^4$ & RN & TTE & Anti-adv  & HD & TTC (ours) & \\
         \midrule
         \color{lightgray}\multirow{2}{*}{TinyImageNet} & \color{lightgray}Rob. & \color{lightgray}0.00 & \color{lightgray}2.19 & \color{lightgray}4.87 & \color{lightgray}10.12 & \color{lightgray}4.39 & \color{lightgray}9.59 & \color{lightgray}0.29 & \color{lightgray}1.24 &\color{lightgray}0.00$\pm0.00$ &\color{lightgray}1.77$\pm1.28$ &\color{lightgray}0.09$\pm0.01$ &\color{lightgray}0.01$\pm0.00$ &\color{lightgray}6.75$\pm0.21$ &\color{lightgray}+6.75 \\
         
         & \color{lightgray}Acc.& \color{lightgray}57.64 & \color{lightgray}77.06 & \color{lightgray}70.86 & \color{lightgray}63.84 &\color{lightgray}66.85 &\color{lightgray}59.77 &\color{lightgray}73.63 &\color{lightgray}70.69 &\color{lightgray}
         51.85$\pm0.04$ 
         &\color{lightgray}
         56.73$\pm0.22$ 
         &\color{lightgray}
         52.62$\pm0.20$ 
         &\color{lightgray}
         51.07$\pm0.09$ 
         &\color{lightgray}
         51.85$\pm0.04$ 
         &\color{lightgray}
         -5.79
         \\
         
         \midrule
         \multirow{2}{*}{CIFAR10} & Rob. &0.43 &2.75 &7.69 &11.7 &10.20 &15.59  &1.94 &5.42 &0.00\std{0.00} &3.47\std{2.77} &0.32\std{0.02} &1.67\std{0.08} &\textbf{28.51}\std{0.36} &\redd{+28.08} \\
         
         & Acc.& 85.12 & 84.90 & 64.61 &65.15 & 70.69 &71.45  & 74.44 &78.46&81.18\std{0.07} &\textbf{84.74}\std{0.40} &83.44\std{0.07} &78.23\std{0.16} &81.18\std{0.07} &\blue{-3.94} \\
         
         \midrule
         \multirow{2}{*}{CIFAR100} & Rob. &0.05 &0.67 &6.54 &9.25 &7.60 &10.80 &2.64 &4.54&0.00\std{0.00} &1.37\std{0.96} &0.22\std{0.03} &0.00\std{0.00} &\textbf{9.06}\std{0.11} &\redd{+9.01} \\
         & Acc.& 57.14 & 59.51 & 35.96 &36.30 & 40.32 &41.51 & 46.67 &47.38&56.34\std{0.20} &\textbf{58.61}\std{0.25} &53.96\std{0.17} &52.86\std{0.16} &56.34\std{0.20} &\blue{-0.80} \\
         
         \midrule
         \multirow{2}{*}{STL10} & Rob. &0.16 &3.75 &24.80 &31.83&28.49 &35.40 &9.99 &17.59& 0.06\std{0.01}&32.56\std{11.76} &2.25\std{0.10} &3.39\std{0.12} &\textbf{52.40}\std{0.34} &\redd{+52.24} \\
         & Acc.& 96.40 & 94.49 & 87.40 &81.69& 88.56 &84.35 & 91.72 &89.11& 95.85\std{0.04} &\textbf{96.26}\std{0.04} &95.47\std{0.06} &89.50\std{0.07} &95.83\std{0.03} &\blue{-0.57} \\
         
         \midrule
         \multirow{2}{*}{ImageNet} & Rob. &0.00 &0.07 &1.65 &3.00  &2.07 &3.34  &0.16 &0.65& 0.00\std{0.00} &6.31\std{3.32} &0.15\std{0.00} &0.01\std{0.00} &\textbf{12.68}\std{0.03} &\redd{+12.68} \\
         & Acc.& 59.69 & 54.24 & 34.89 &27.76  & 36.12 &28.51  & 48.79 &40.48& 59.34\std{0.06} &\textbf{60.02}\std{0.13} &54.29\std{0.07} &54.54\std{0.05} &34.00\std{0.06} &\blue{-25.69} \\
         
         \midrule
         \multirow{2}{*}{Caltech101} & Rob. &0.59 &4.81 &15.75 &21.00  &19.48 &25.03&5.15 &10.13& 0.68\std{0.02} &30.19\std{7.92} &3.14\std{0.07} &1.27\std{0.03} &\textbf{36.66}\std{0.25} &\redd{+36.07} \\
         & Acc.& 85.66 & 83.63 & 71.68 &64.41  & 75.45 &69.06  & 80.95 &76.58& \textbf{86.61}\std{0.10} &85.84\std{0.09} &83.99\std{0.07} &82.33\std{0.04} &86.15\std{0.08} & \redd{+0.49}\\
         
         \midrule
         \multirow{2}{*}{Caltech256} & Rob. &0.12 &1.41 &8.29 &11.76  &10.65 & 13.68 &2.18 &5.09& 0.16\std{0.00} &23.23\std{7.77} &1.44\std{0.03} &0.34\std{0.02} &\textbf{27.25}\std{0.08} &\redd{+27.13}\\
         & Acc.& 81.72 & 78.53 & 61.14 &52.05  & 62.24 &53.32  & 73.32 &67.22& 81.25\std{0.03} &\textbf{82.48}\std{0.08} &79.40\std{0.07} &79.12\std{0.01} &76.59\std{0.12} &\blue{-5.13} \\
         
         \midrule
         \multirow{2}{*}{OxfordPets} & Rob. &0.00 &1.66 &0.90 &3.71  &1.74 & 5.10 &0.19 &0.30& 0.00\std{0.00} &3.18\std{2.94} &0.10\std{0.04} &0.00\std{0.00} &\textbf{24.64}\std{0.53} &\redd{+24.64} \\
         & Acc.& 87.44 & 84.14 & 62.12 &53.94  & 65.88 & 56.66 & 79.37 &70.10& 87.41\std{0.12} &\textbf{88.13}\std{0.13} &80.53\std{0.17} &80.91\std{0.05} &64.70\std{0.33} &\blue{-22.74} \\
         
         \midrule
         \multirow{2}{*}{Flowers102} & Rob. &0.00 &0.13 &1.87 &3.81  &2.57 &4.26&0.03 &0.62& 0.00\std{0.00}&3.52\std{2.51} &0.05\std{0.02} &0.00\std{0.00} &\textbf{13.60}\std{0.33} &\redd{+13.60} \\
         & Acc.& 65.46 & 53.37 & 36.80 &27.78 & 37.00 & 28.88 & 47.98 &41.01& 64.62\std{0.19} &\textbf{65.20}\std{0.23} &62.80\std{0.02} &58.22\std{0.12} &63.24\std{0.21} &\blue{-2.22} \\
         
         \midrule
         \multirow{2}{*}{FGVCAircraft} & Rob. &0.00 &0.00 &0.03 &0.12  &0.03 &0.06&0.00 &0.03& 0.00\std{0.00}&0.43\std{0.43} &0.00\std{0.00} &0.00\std{0.00} &\textbf{6.40}\std{0.38} &\redd{+6.40} \\
         & Acc.& 20.10 & 14.04 & 5.31 &3.51 & 5.55 &3.24& 10.86 &7.77& 19.25\std{0.18} &\textbf{20.18}\std{0.35} &15.64\std{0.17} &16.36\std{0.03} &15.99\std{0.04} & \blue{-4.11}\\
         
         \midrule
         \multirow{2}{*}{StanfordCars} & Rob. &0.00 &0.00 &0.15 &0.41  &0.15 &0.40&0.01 &0.04& 0.00\std{0.00} &1.46\std{1.21} &0.00\std{0.00} &0.00\std{0.00} &\textbf{12.84}\std{0.20} &\redd{+12.84} \\
         & Acc.& 52.02 & 42.11 & 20.91 &15.18  & 25.44 &16.79& 38.68 &32.09& 52.14\std{0.09} &\textbf{52.73}\std{0.31} &36.14\std{0.30} &44.28\std{0.02} &41.52\std{0.15} &\blue{-10.50} \\
         
         \midrule
         \multirow{2}{*}{SUN397} & Rob. &0.00 &0.02 &1.30 &2.31 &1.90 &3.24&0.13 &0.57  &0.00\std{0.00} &5.95\std{3.39} &0.11\std{0.00} &0.00\std{0.00} &\textbf{13.43}\std{0.08} &\redd{+13.43} \\
         & Acc.& 58.50 & 55.73 & 36.69 &28.16  & 37.98 &29.93& 52.42 &43.57& \textbf{59.69}\std{0.06} &59.12\std{0.08} &55.99\std{0.04} &53.17\std{0.02} &46.68\std{0.02} &\blue{-11.82} \\
         
         \midrule
         \multirow{2}{*}{Country211} & Rob. &0.00 &0.00 &0.05 &0.19  &0.12 &0.24&0.00 &0.02& 0.00\std{0.00} &0.24\std{0.15} &0.00\std{0.00} &0.00\std{0.00} &\textbf{2.44}\std{0.15} &\redd{+2.44} \\
         & Acc.& 15.25 & 12.07 & 4.75 &3.66& 4.64 &3.34& 9.26 &6.58& \textbf{14.80}\std{0.02} &14.66\std{0.14} &11.60\std{0.08} &11.72\std{0.07} &11.99\std{0.01} &\blue{-3.26} \\
         
         \midrule
         \multirow{2}{*}{Food101} & Rob. &0.00 &0.04 &0.56 &1.35  &1.03 &2.12&0.06 &0.24& 0.00\std{0.00} &5.31\std{4.09} &0.07\std{0.02} &0.01\std{0.00} &\textbf{17.89}\std{0.13} &\redd{+17.89} \\
         & Acc.& 83.88 & 64.86 & 29.98 &21.90 & 36.61 &27.97& 55.31 &41.98& 83.44\std{0.04} &\textbf{83.96}\std{0.01} &75.95\std{0.17} &80.30\std{0.05} &80.00\std{0.07} &\blue{-3.88} \\
         
         \midrule
         \multirow{2}{*}{EuroSAT} & Rob. &0.00 &0.00 &9.77 &10.71  &9.61 &10.36&0.00 &7.34& 0.00\std{0.00} &0.11\std{0.09} &0.03\std{0.02} &0.20\std{0.02} &\textbf{13.57}\std{0.12} &\redd{+13.57} \\
         & Acc.& 42.59 & 27.64 & 16.58 &17.53  & 18.53 &19.19& 21.88 &18.22& \textbf{53.24}\std{0.09} &44.38\std{1.62} &36.81\std{0.12} &39.08\std{0.06} &\textbf{53.24}\std{0.09} &\redd{+10.65} \\
         
         \midrule
         \multirow{2}{*}{DTD} & Rob. &0.11 &0.00 &4.20 &5.16  &4.31 &5.21&0.90 &2.50& 0.11\std{0.00} &7.16\std{2.32} &0.37\std{0.04} &0.16\std{0.04} &\textbf{11.40}\std{0.28} &\redd{+11.29} \\
         & Acc.& 40.64 & 36.49 & 25.16 &20.11  & 21.76 &17.29& 32.07 &28.03& 37.96\std{0.13} &\textbf{41.35}\std{0.29} &38.55\std{0.12} &34.89\std{0.35} &35.69\std{0.08} &\blue{-4.95} \\
         
         \midrule
         \multirow{2}{*}{PCAM} & Rob. &0.00 &0.00 &20.54 &44.13  &12.59 &36.38&0.64 &3.74& 0.00\std{0.00} &0.22\std{0.23} &0.25\std{0.03} &12.04\std{0.11} &\textbf{47.39}\std{0.20} &\redd{+47.39} \\
         & Acc.& 52.02 & 47.21 & 49.96 &49.98  & 50.03 &49.80& 52.54 &50.17& \textbf{52.73}\std{0.07} &50.92\std{0.04} &52.61\std{0.07} &50.38\std{0.04} &\textbf{52.73}\std{0.07} &\redd{+0.71} \\
         
         \midrule
         \midrule
         \multirow{2}{*}{\textbf{Avg.}} & Rob. &0.09 &0.96 &6.51 &10.03  &7.03 &10.70&1.50 &3.67& 0.06\std{0.00} &7.79\std{3.23} &0.53\std{0.00} &1.19\std{0.01} &\textbf{20.63}\std{0.05} &\redd{+20.54} \\
         & Acc. & 61.51 & 55.80 & 40.25 &35.57 & 42.30 &37.58& 51.02 &46.17&61.61\std{0.03} &\textbf{61.79}\std{0.13} &57.32\std{0.03} &56.62\std{0.02} &55.99\std{0.06} &\blue{-5.52} \\
         \bottomrule
    \end{tabular}}
    \caption{\label{tab:white_box_eps4_full}Classification accuracy (\%) on both adversarial images (Rob.) under 10-step PGD attack at $\epsilon_a=4/255$ and clean images (Acc.) across 16 datasets.
    Weights and gradients of the deployed model are assumed to be known to the threat model. Comparison is made among our paradigm and test-time defences adapted from existing adversarial studies, with finetuning-based models implemented as a reference.
    The superscripts of the model names indicate the attack budget used for generating adversarial images in the phase of adversarial finetuning.
    We report the mean and standard deviation for test-time methods over 3 runs. 
    The last column reports the gains w.r.t. original CLIP without any finetuning or test-time operations.}
\end{table*}

\begin{figure}[h]
    \centering
    \includegraphics[width=0.8\linewidth]{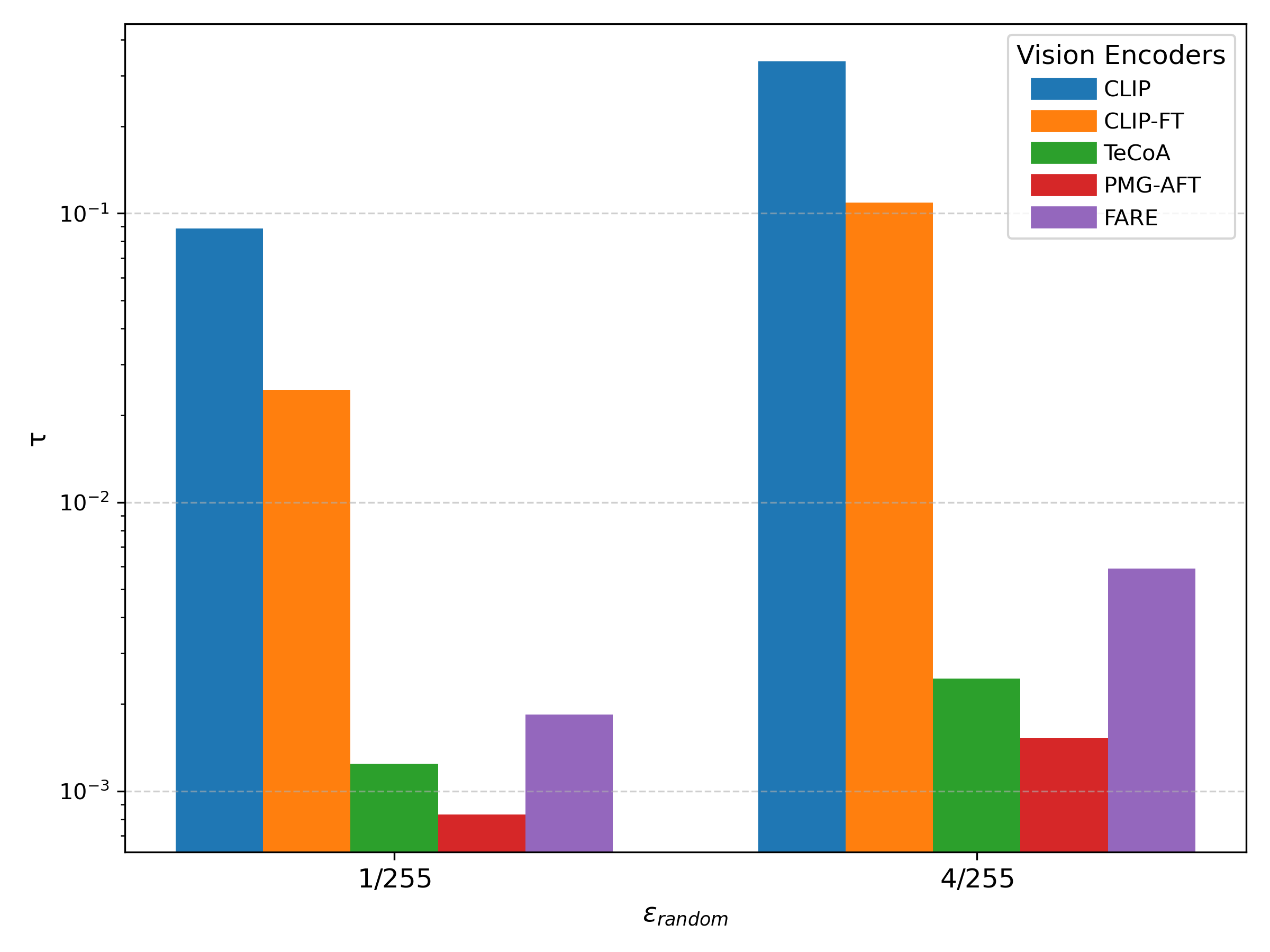}
    \caption{Average $\tau$ of different CLIP vision encoders on randomly sampled clean images across 16 datasets.}
    \label{fig:tau_AFT_pitfall}
    \vspace{-0.5cm}
\end{figure}

\section{Pitfalls of Adversarial Finetuning}\label{sec:pitfalls_aft}
In the main paper, we find that although TTC can further improve robustness of adversarially finetuned CLIP models at test time (\cref{tab:ttc_robust_models}), the robustness gains are less obvious compared to the original CLIP. 
We also find that employing TTC on unsupervised adversarial finetuning method FARE \cite{pmlr-v235-schlarmann24a} achieves greater gains compared to when employing TTC on TeCoA \cite{mao2023understanding} and PMG-AFT \cite{wang2024pre}, which are supervised adversarially finetuned CLIP models.
Since our TTC paradigm is based on the expressiveness of the pre-trained vision encoder $f_\theta$, we investigate this behaviour from the perspective of $f_\theta$.
Through analysis of randomly sampled images, we find that adversarial finetuning significantly reduces the sensitivity of $f_\theta$ to nuanced variations in the pixel space.
We 
study the values of $\tau$ of different adversarially finetuned vision encoders when a random noise is imposed on clean images and report the results in \cref{fig:tau_AFT_pitfall}. 
As can be seen from the figure, adversarial finetuning reduces 
the sensitivity of $f_\theta$ to pixel-level variations
by orders of magnitude,
which we believe is the key mechanism through which the adversarially finetuned models of CLIP achieve robustness against adversaries.
Regular finetuning of CLIP (CLIP-FT), i.e., finetuning the vision encoder with clean images on TinyImageNet, also reduces the perception sensitivity to some extent.
Among adversarially finetuned models, FARE shows greater preservation of sensitivity compared to its supervised counterparts TeCoA and PMG-AFT, which explains the 
lower levels of adversarial robustness of FARE (\cref{tab:white_box_eps1}) and 
better robustness gains when employing TTC on FARE at test time (\cref{tab:ttc_robust_models}).
Although resilience to pixel-level variations translates to robustness of CLIP to imperceptible malicious perturbations, 
it causes the vision encoder to be less expressive. 
We argue that a fundamental difference between CLIP and non-foundational models is that CLIP has learned massive amounts of real-world knowledge, which should be taken into account in attempts aiming to enhance its robustness. 
We also recommend cautious use of adversarial finetuning as the only robustifying approach for CLIP and other large pre-trained models in general.

\section{Effects of Other Hyperparameters}\label{sec:other_hyperparams}
In the main paper, we find that the number of counterattack steps $N$ is the crucial hyperparameter that greatly impacts robustness.
In this section, we investigate the impact of the other two hyperparameters $\tau_{thres}$ (\cref{eq:tau}) and $\beta$ (\cref{eq:weight}), which control the threshold of $L_2$ drift ratio and the ascending rate of weights across counterattack steps, respectively (\cref{alg:ttc}).
We vary one hyperparameter at a time w.r.t. the default setting $\tau_{thres}=0.2$ and $\beta=2.0$.
The counterattack budget $\epsilon_{ttc}$ and steps $N$ are fixed to $\epsilon_{ttc}=4/255$ and $N=2$, respectively.
We report the results in \cref{tab:tau_beta_impact}. It can be seen that both hyperparameters control the trade-off between the accuracy on clean images and adversarial images. 
When the threshold $\tau_{thres}$ is relatively small, the accuracy on clean images can be better retained, while the robustness gains are limited, since the values of $\tau$ for most clean and adversarial images are above the set threshold, which halts necessary counterattacks.
Robustness increases as $\tau_{thres}$ is set higher, and reaches a plateau after $\tau_{thres}=0.2$. Further increasing the threshold compromises accuracy on clean images.
The impact of $\beta$ is less obvious. In general, a larger $\beta$ assigns higher weights to counterattack perturbations at later steps, thereby favouring robustness.

\begin{table*}[t!]
    \centering
    \resizebox{\textwidth}{!}{
    \begin{tabular}{cc|cccccccccccccccc|cc}
    \toprule
    $\tau_{thres}$ & $\beta$ & \rotatebox{60}{CIFAR10} & \rotatebox{60}{CIFAR100} & \rotatebox{60}{STL10} & \rotatebox{60}{ImageNet} & \rotatebox{60}{Caltech101} & \rotatebox{60}{Caltech256} & \rotatebox{60}{OxfordPets} & \rotatebox{60}{Flower102} & \rotatebox{60}{FGVCAircraft} & \rotatebox{60}{StanfordCars} & \rotatebox{60}{SUN397} & \rotatebox{60}{Country211} & \rotatebox{60}{Food101} & \rotatebox{60}{EuroSAT} & \rotatebox{60}{DTD} & \rotatebox{60}{PCAM} & \rotatebox{60}{Avg. Rob.} & \rotatebox{60}{Avg. Acc.}\\
    \midrule
    \rowcolor{blue!15}
    0.2 & 2.0 & 28.75 & 14.31 & 76.70 & 38.41 & 65.78 & 60.11 & 57.87 & 39.14 & 13.77 & 33.01 & 41.52& 7.09 &57.84  &12.19 &27.32 &52.85 &39.17 &59.75 \\
    \midrule
    0.05 & 2.0 &2.07 &0.69 &16.35 &2.77 &19.14 &11.83 &3.35 &2.75 &0.00 &0.37 &2.35 &0.12 &1.51 &0.14 &5.74 &3.72 &4.56 &61.63 \\
    0.1 & 2.0 &2.15 &0.88 &26.21 &18.95 &32.96 &28.28 &32.35 &25.13 &2.19 &16.78 &17.06 &2.35 &29.31 &0.67 &17.66 &37.25 &18.14 &61.46 \\
    0.15 & 2.0 &7.62 &4.96 &56.45 &33.06 &55.43 &50.48 &52.77 &36.82 &9.33 &30.33 &34.07 &5.48 &53.22 &5.66 &25.21 &48.50 &31.84 &61.00 \\
    0.25 & 2.0 &44.20 &22.29 &83.09 &40.18 &69.32 &63.45 &58.93 &40.01 &14.97 &33.42 &43.64 &7.73 &58.63 &18.16 &28.46 &55.49 &42.62 &57.31 \\
    0.3 & 2.0 &50.10 &25.94 &84.86 &40.66 &70.43 &64.48 &59.12 &40.14 &15.30 &33.44 &44.30 &7.89 &58.87 &21.19 &28.88 &56.65 &43.89 &54.18 \\
    0.35 & 2.0 &51.97 &27.07 &85.49 &40.83 &70.76 &64.94 &59.23 &40.15 &15.45 &33.48 &44.49 &7.96 &58.95 &22.64 &29.10 &57.17 &44.35 &50.67 \\
    0.4 & 2.0 &52.36 &27.51 &85.56 &40.91 &70.89 &65.10 &59.25 &40.15 &15.51 &33.48 &44.56 &7.99 &58.99 &23.44 &29.15 &57.40 &44.52 &47.75 \\
    \midrule
    0.2 & 0.5 &27.08 &13.25 &74.58 &33.53 &63.50 &57.44 &48.24 &32.90 &10.98 &27.75 &36.45 &5.71 &51.96 &12.11 &24.95 &41.20 &35.10 &60.24 \\
    0.2 & 1.0 &28.01 &13.84 &75.97 &36.39 &64.94 &59.12 &53.80 &36.27 &12.72 &30.95 &39.35 &6.48 &55.60 &12.44 &26.65 &48.03 &37.54 &60.00 \\
    0.2 & 1.5 &28.42 &14.02 &76.46 &37.73 &65.54 &59.81 &56.55 &38.12 &13.50 &32.28 &40.80 &6.90 &57.18 &12.49 &27.45 &51.25 &38.66 &59.85 \\
    0.2 & 2.5 &28.82 &14.13 &76.81 &38.77 &65.89 &60.25 &58.54 &39.76 &13.71 &33.44 &41.83 &7.28 &58.10 &12.55 &27.77 &53.98 &39.48 &59.72 \\
    0.2 & 3.0 &28.95 &14.15 &76.91 &38.95 &66.03 &60.34 &58.90 &40.06 &13.92 &33.62 &42.07 &7.36 &58.25 &12.54 &27.87 &54.50 &39.65 &59.70 \\
    \bottomrule
    \end{tabular}}
    \caption{The Effects of hyperparameters $\tau_{thres}$ and $\beta$ under 10-step PGD attack with $\epsilon_a=1/255$. The counterattack budget and steps are fixed at $\epsilon_{ttc}=4/255$ and $N=2$, respectively.
    We report the robust accuracy for each dataset. The last column reports the average accuracy on clean images across 16 datasets.
    }
    \label{tab:tau_beta_impact}
\end{table*}

\section{Adaptive Attacks}\label{sec:adaptive_attack}
In the paper, we demonstrate that CLIP possesses the ability to defend itself from adversarial attacks that aim to maximise the classification loss of CLIP, assuming that such counterattacks by the end user are not known to the attacker.
Here we provide a gradient-based method tailored to break our TTC.
Our TTC paradigm can be written as $\varphi(x)=x+\delta^{\ast}(x)$, where $x$ is a test image and $\delta^{\ast}$ is a function of $x$ that induces the maximum $L_2$ drift of $x$ in the embedding space of CLIP:
\begin{equation}\label{eq:delta_ast}
    \delta^{\ast}(x) = \arg\max_{\delta} \|f_\theta(x+\delta)-f_\theta(x)\|,\;s.t.\,\|\delta\|\leq\epsilon_{ttc}
\end{equation}
Therefore, the attacker may incorporate $\varphi(x)$ into the objective when crafting an adversarial image aiming to maximise the classification loss:
\begin{equation}\label{eq:adaptive_attack}
    x' = \arg\max_{x_s}L(f_\theta(\varphi(x_s)),t_c),\;s.t.\,\|x_s-x\|\leq\epsilon_a
\end{equation}
When employing gradient-based attack methods such as PGD to solve \cref{eq:adaptive_attack}, 
the inner optimization of \cref{eq:delta_ast} can be approximated by a one-step update:
\begin{equation}\label{eq:one_step_appro}
\begin{split}
    \varphi(x) &= x+\delta^{\ast}(x)\\
     &\approx x + \delta^0+\eta\nabla_{\delta}\|f_\theta(x+\delta^0)-f_\theta(x)\|
\end{split}    
\end{equation}
where $\eta$ is the step size for the counterattack and $\delta^0\thicksim U(-\epsilon_{ttc},\epsilon_{ttc})$ is a randomly initialised noise $\delta^0$.
Thus, the objective for generating the adversarial attack can be written as $L\left(f_\theta(x+\delta^0+\eta\nabla_{\delta}\|f_\theta(x+\delta^0)-f_\theta(x)\|),\:t_c\right)$.
By employing PGD to craft an adversary that maximises this objective, the attacker may break the counterattacks performed by the end user.


\end{document}